\documentclass{article}

\usepackage{PRIMEarxiv}

\usepackage[utf8]{inputenc} % allow utf-8 input
\usepackage[T1]{fontenc}    % use 8-bit T1 fonts
\usepackage{hyperref}       % hyperlinks
\usepackage{url}            % simple URL typesetting
\usepackage{booktabs}       % professional-quality tables
\usepackage{amsfonts}       % blackboard math symbols
\usepackage{nicefrac}       % compact symbols for 1/2, etc.
\usepackage{microtype}      % microtypography
\usepackage{lipsum}
\usepackage{fancyhdr}       % header
\usepackage{graphicx}       % graphics
\usepackage{subfigure}
\usepackage{xcolor}
\usepackage{amsmath}
\usepackage{cleveref}
\usepackage{bm}
\graphicspath{{media/}}     % organize your images and other figures under media/ folder
\usepackage{textcomp}  % Required for encoding \textbigcircle
\usepackage{scalerel}  % Required for emoji \scalerel
\usepackage{tabularx}
\usepackage{algorithm}
\usepackage{algpseudocode}
\usepackage{makecell}
\usepackage{framed} 
\usepackage{enumitem}
\usepackage{array,multirow,graphicx}
\usepackage{float}
\usepackage{mdframed}
\usepackage{listings} % Code formatting
\usepackage{xcolor} % Colors for code
\usepackage[most]{tcolorbox/tcolorbox}
\usepackage{threeparttable}
\def\thellm{\textsc{PLM}}

\title{\thellm{}: Efficient Peripheral Language Models\\ Hardware-Co-Designed for Ubiquitous Computing}
\author{
Cheng Deng$^{1,*,\dagger}$, Luoyang Sun$^{2,*}$, Jiwen Jiang$^{2,*}$, Yongcheng Zeng$^2$, \\ \textbf{Xinjian Wu$^3$, Wenxin Zhao$^1$, Qingfa Xiao$^1$, Jiachuan Wang$^4$, Haoyang Li$^5$,}\\ \textbf{Lei Chen$^{1,4}$,  Lionel M. Ni$^1$, Haifeng Zhang$^2$, Jun Wang$^{3,6,\dagger}$} \\
$^1$The Hong Kong University of Science and Technology (Guangzhou) \\
$^2$Institution of Automation, Chinese Academy of Sciences \\
$^3$University College London \\
$^4$The Hong Kong University of Science and Technology \\
$^5$The Hong Kong Polytechnic University \\
$^6$UCL Centre for Artificial Intelligence \\
\texttt{davendw49@gmail.com, leichen@cse.ust.hk, jun.wang@cs.ucl.ac.uk} \\
 \\
\textbf{\href{https://www.project-plm.com/}{\thellm{} Team}} \\
\texttt{\url{https://github.com/plm-team/PLM}}
}

\begin{document}
\begin{sloppypar}
\maketitle

\renewcommand{\thefootnote}{\fnsymbol{footnote}}
    \footnotetext[1]{Equal contribution.}
    \footnotetext[2]{Correspondence to \href{mailto:davendw49@gmail.com}{Cheng Deng}, \href{leichen@cse.ust.hk}{Lei Chen}, \href{jun.wang@cs.ucl.ac.uk}{Jun Wang}.}
    \footnotetext[3]{Version: v1 (major update on 15$^{th}$ March, 2025).}
\renewcommand{\thefootnote}{\arabic{footnote}}

\begin{abstract}

While scaling laws have been continuously validated in large language models (LLMs) with increasing model parameters, the inherent tension between the inference demands of LLMs and the limited resources of edge devices poses a critical challenge to the development of edge intelligence. Recently, numerous small language models have emerged, aiming to distill the capabilities of LLMs into smaller footprints. However, these models often retain the fundamental architectural principles of their larger counterparts, still imposing considerable strain on the storage and bandwidth capacities of edge devices.
In this paper, we introduce the \thellm{}, a \textbf{P}eripheral \textbf{L}anguage \textbf{M}odel, developed through a co-design process that jointly optimizes model architecture and edge system constraints. The \thellm{} utilizes a Multi-head Latent Attention mechanism and employs the squared ReLU activation function to encourage sparsity, thereby reducing peak memory footprint during inference.
During training, we collect and reorganize open-source datasets, implement a multi-phase training strategy, and empirically investigate the Warmup-Stable-Decay-Constant (WSDC) learning rate scheduler. Additionally, we incorporate Reinforcement Learning from Human Feedback (RLHF) by adopting the ARIES preference learning approach. Following a two-phase SFT process, this method yields performance gains of 2\% in general tasks, 9\% in the GSM8K task, and 11\% in coding tasks.
In addition to its novel architecture, evaluation results demonstrate that \thellm{} outperforms existing small language models trained on publicly available data while maintaining the lowest number of activated parameters. Furthermore, deployment across various edge devices, including consumer-grade GPUs, mobile phones, and Raspberry Pis, validates \thellm{}'s suitability for peripheral applications. The \thellm{} series models, including \thellm{}-1.8B-Base and \thellm{}-1.8B-Instruct, are publicly available at \url{https://github.com/plm-team/PLM}.
\end{abstract}

% \newpage

% \tableofcontents

% \begin{figure}[H]
%     \centering
%     \includegraphics[width=\linewidth]{imgs/devices.png}
%     \caption{\thellm{} deployment cases.\dc{Real cases images.}}
%     \label{fig:deploy}
% \end{figure}

% \newpage

% keywords can be removed
% \keywords{First keyword \and Second keyword \and More}

\section{Introduction}

% General starting
% Why edge side needs LLM
Large Language Models (LLMs) have revolutionized various AI applications. On-device LLMs are gaining prominence, as running these models locally on edge devices can avoid the drawbacks brought by the cloud-based LLMs, like privacy risks from uploading sensitive data~\cite{sensitive}, increased carbon footprint due to data transfers and reliance on cloud GPUs~\cite{faiz2023llmcarbon}, and inaccessibility during network outages, limiting use in critical applications~\cite{li2024chatterbox}. In many specialized fields, applying AI is challenging due to the trade-off between model size~\cite{lin2023geogalactica} and available computing resources.

To address this concern, Small Language Models (SLMs) have emerged as a viable solution~\cite{lu2024small}. The 1.5B models in the Qwen series~\cite{yang2024qwen2,yang2024qwen2_5} exemplify out-of-the-box LLMs optimized for small-scale deployment. Similarly, the MiniCPM series~\cite{hu2024minicpm} demonstrates the potential of end-side LLMs, while the SmolLM~\cite{allal2024SmolLM2} and Yulan-mini~\cite{hu2024yulan} series provide compelling evidence that high-quality training data significantly enhances the performance of SLMs. Moreover, Deepseek has put forward a reasoning model based on Qwen2-1.5B distilled from Deepseek R1~\cite{guo2025deepseek}, making it a direction for making SLM have the abilities of reasoning and doing decision.

Beyond training models at a reduced scale, model compression techniques such as AWQ~\cite{lin2024awq} and GPTQ~\cite{frantar2022gptq} offer alternative approaches to facilitate LLM deployment in resource-constrained environments. Moreover, advancements in inference methods have been closely aligned with system constraints, enabling efficient deployment. For instance, MLC-LLM~\cite{mlcllm}, leveraging the TVM backend~\cite{chen2018tvm}, along with frameworks such as llama.cpp~\cite{llamacpp} and PowerInfer~\cite{song2024powerinfer,xue2024powerinfer}, enable optimized performance on various edge CPU and GPU platforms. Additionally, systems like vLLM~\cite{kwon2023efficient} and SGLang~\cite{zheng2024sglang} provide significant acceleration on devices with limited computational resources.

\begin{figure}[t]
    \centering
    \includegraphics[width=\linewidth]{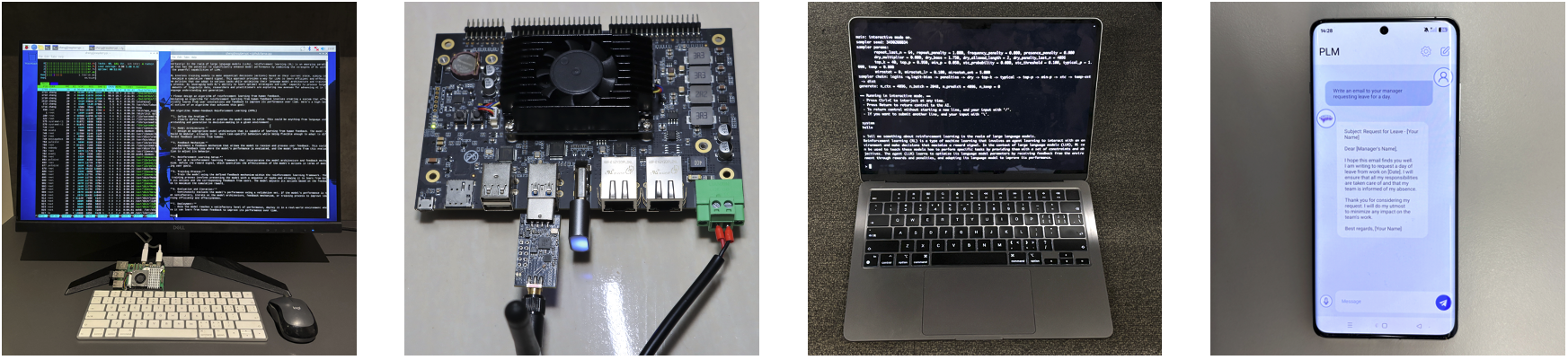}
    \caption{System hardwares used in our experiments: (L-R) Broadcom (Raspberry Pi), NVIDIA Jetson Orin NX, Apple M3 (MacBook Air 2024), Qualcomm Snapdragon 8 Gen 3 (OnePlus 12 Pro), and the NVIDIA A10 experimental device is installed in the server.}
    \label{fig:devices}
    \vspace{-2em}
\end{figure}

The research and development efforts in this domain consistently highlight the constraints of computing resources on edge devices, underscoring the critical challenge of peripheral computing. In practical scenarios, such as on mobile phones and personal computers, available memory cannot be exclusively allocated to LLMs. During real-world deployment, computational resources must be shared with other applications. This challenge becomes even more pronounced on specialized System-on-Chip (SoC) platforms, such as personal computer, mobile phone, and Raspberry Pi devices. These scenarios necessitate dynamic resource management, where the model is loaded into memory only when required and its occupied memory is released when no longer in use.

PowerInfer2~\cite{xue2024powerinfer} and LLM in the Flash~\cite{alizadeh2023llm} offer promising solutions to address these challenges. However, the inference process of LLMs introduces additional complexity, as the generation of KV caches increases both computational demands and memory usage. Offloading KV caches to external storage mitigates memory constraints but simultaneously incurs higher I/O overhead, thereby imposing additional limitations on many deployment scenarios.

% In summary, designing models for edge devices should go beyond merely reducing model size or optimizing hardware-specific inference frameworks. A more holistic approach is required—one that emphasizes co-design of models and system architectures, allowing for tailored solutions that address specific deployment environments and requirements.

What constitutes the most effective model design for edge devices? There is no universally accepted answer to this question. Based on extensive research and experimental validation, in this paper, we shares the way we think, design, train and deploy the LLM on edge devices. 

Current advancements in LLM design highlight two critical components for optimization: attention mechanisms and MLP layers. Attention mechanisms often represent a significant computational bottleneck during LLM inference, particularly as sequence lengths increase. Innovations in attention design provide valuable solutions to this challenge. 
% For example, the evolution from Google's Multi-Head Attention (MHA)~\cite{vaswani2017attention} and Group Query Attention (GQA)~\cite{ainslie2023gqa} to more advanced techniques like Deepseek's Multi-Head Latent Attention (MLA)~\cite{liu2024deepseek} and state-space models such as Mamba~\cite{gu2023mamba} has demonstrated substantial improvements in efficiency. These methods reduce computational overhead while maintaining or enhancing the model's ability to capture complex relationships in data.
MLP layers play a critical role in processing each token independently to refine its representation through dense matrix multiplications and activation functions. However, these operations contribute significantly to the computational load in LLMs. 
% To improve the efficiency of MLP layers, various techniques have been employed, including low-rank approximations~\cite{hu2021lora} to reduce the dimensionality of weight matrices, sparsity to minimize unnecessary computations~\cite{song2024turbo}, quantization, and weight sharing to reduce model size and computation.

Our proposed model, \thellm{} (\textbf{P}eripheral \textbf{L}anguage \textbf{M}odel), builds upon the strengths of recent advancements by achieving a balance between performance and efficiency. Guided by the principle of modeling and system co-design, \thellm{} incorporates innovative features such as squared ReLU, an activation function inspired by recent research of activation sparsity in LLMs~\cite{song2024turbo,mirzadeh2023relu,so2021searching}, designed to enhance computational sparsity. Additionally, \thellm{} leverages Multi-Head Latent Attention (MLA)~\cite{liu2024deepseek}, incorporating insights from Deepseek-v2 from the angle of low-rank to achieve superior efficiency in attention mechanisms.

In this report, we introduce \thellm{}, a series of edge-oriented LLMs that address key research challenges, including sketching model architecture through language modeling and system co-design, multi-phases training using Warm-up-Stable-Decay-Constant (WSDC) learning rate scheduler, multi-phases supervised fine-tuning from shallow to deep, preference learning, and fast inference on peripheral devices. To support further research and application development, we publicly release \thellm{} model weights. The distinctiveness and contributions of \thellm{} can be summarized as follows:

\begin{itemize}[leftmargin=1.2em]
\item[1.] Introducing \thellm{}-1.8B, a novel LLM tailored for peripheral devices. This 1.8 billion-parameter model leverages multi-head latent attention and a squared ReLU sparsification technique. It demonstrates competitive performance against similarly sized LLMs trained on open-source data, while also providing efficient decoding capabilities. 
\item[2.] In this report, we detail the whole construction progress of \thellm{}, including modeling, data curation, sandbox experiments, training, model evaluation, and deployment on edge devices.
\item[3.] We discuss the observation and benefit of \thellm{}. Utilizing the "low-rank and sparsity" modeling and system co-design, we implement the examples on edge devices, such as consume level GPUs, Jetson Orin, mobile phone, personal laptop, and Raspberry Pis.
\item[4.] We discuss the key factors in the LLM design for peripheral computing, from the perspective of peripheral systems and language modeling co-design. Based on the methods survey and observation from the benchmarking experiments, we illustrate the future development of the edge LLM for peripheral computing and beyond.
\end{itemize}

All the models, data used, benchmarks, and related introduction are shown on \href{https://github.com/plm-team/PLM}{Github} and \url{https://sites.google.com/view/project-plm}.

\begin{figure}[t]
    \centering
    \includegraphics[width=0.6\linewidth]{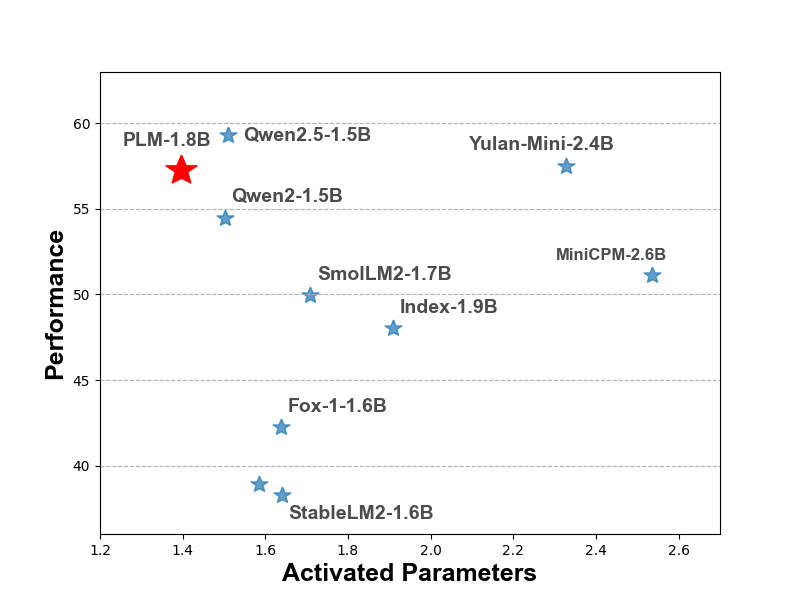}
    \caption{Evaluations on LLM benchmarks demonstrate that our sparsified MLA models, \thellm{}, maintain competitive performance. The performance is evidenced by the averaged benchmark scores in~\autoref{expall}, in terms of general knowledge comprehension (MMLU, CMMLU, C-Eval, ARC), Math problem solving (GSM8K and MathQA), coding proficiency (HumanEval and MBPP), and commonsense and logical reasoning (HellaSwag, BooQ, LogiQA, and PIQA). Meanwhile, the activated parameters are determined by the minimal computation required to preserve modeling performance. Detailed results are presented in~\autoref{obsall}.}
    \label{fig:sparse_compare}
\end{figure}
\section{Architecture}

% \usepackage{booktabs}
\begin{table*}[h]
\resizebox{\textwidth}{!}{%
\begin{tabular}{@{}cccccccccc@{}}
\toprule
\multicolumn{1}{c}{\textbf{\begin{tabular}[c]{@{}c@{}}Number of \\ layers\end{tabular}}} & \multicolumn{1}{c}
{\textbf{\begin{tabular}[c]{@{}c@{}}Hidden \\ Dimension\end{tabular}}} & \multicolumn{1}{c}
{\textbf{\begin{tabular}[c]{@{}c@{}}Number of \\ Attention Heads\end{tabular}}} & \multicolumn{1}{c}
{\textbf{\begin{tabular}[c]{@{}c@{}}Number of \\ KV Heads\end{tabular}}} & \multicolumn{1}{c}
{\textbf{\begin{tabular}[c]{@{}c@{}}KV LoRA \\ Rank\end{tabular}}} & \multicolumn{1}{c}
{\textbf{\begin{tabular}[c]{@{}c@{}}Sequence \\ Length\end{tabular}}} & \multicolumn{1}{c}
{\textbf{\begin{tabular}[c]{@{}c@{}}Intermediate \\ Size\end{tabular}}} & \multicolumn{1}{c}
{\textbf{\begin{tabular}[c]{@{}c@{}}Vocabulary \\ Size\end{tabular}}} & \multicolumn{1}{c}
{\textbf{\begin{tabular}[c]{@{}c@{}}Attention \\ Type\end{tabular}}} & \multicolumn{1}{c}
{\textbf{\begin{tabular}[c]{@{}c@{}}Activation \\ Function\end{tabular}}}\\ \midrule
 32 & 2,048 & 16 & 16 & 512 & 4,096 & 8,192 & 151,936 & MLA & ReLU$^2$ \\ \bottomrule
\end{tabular}
}
\caption{Core settings decide the size of \thellm{}-1.8B.}
\label{elm_arch}
\end{table*}

\thellm{} is based on a standard decoder-only Transformer architecture. Specific hyperparameters related to model size are provided in~\autoref{elm_arch}. 
The model consists of \emph{0.31B} embedding parameters and \emph{1.51B} non-embedding parameters. Key architectural features include Multi-head Latent Attention (MLA)~\cite{liu2024deepseek}, Decoupled Rotary Position Embeddings (RoPE)~\cite{su2024roformer}, a BPE tokenizer inheriting the Qwen2 tokenizer~\cite{yang2024qwen2}, squared ReLU activations~\cite{so2021searching} within the MLP layers, and tied input-output embeddings. 

\begin{figure}
    \centering
    \includegraphics[width=0.9\linewidth]{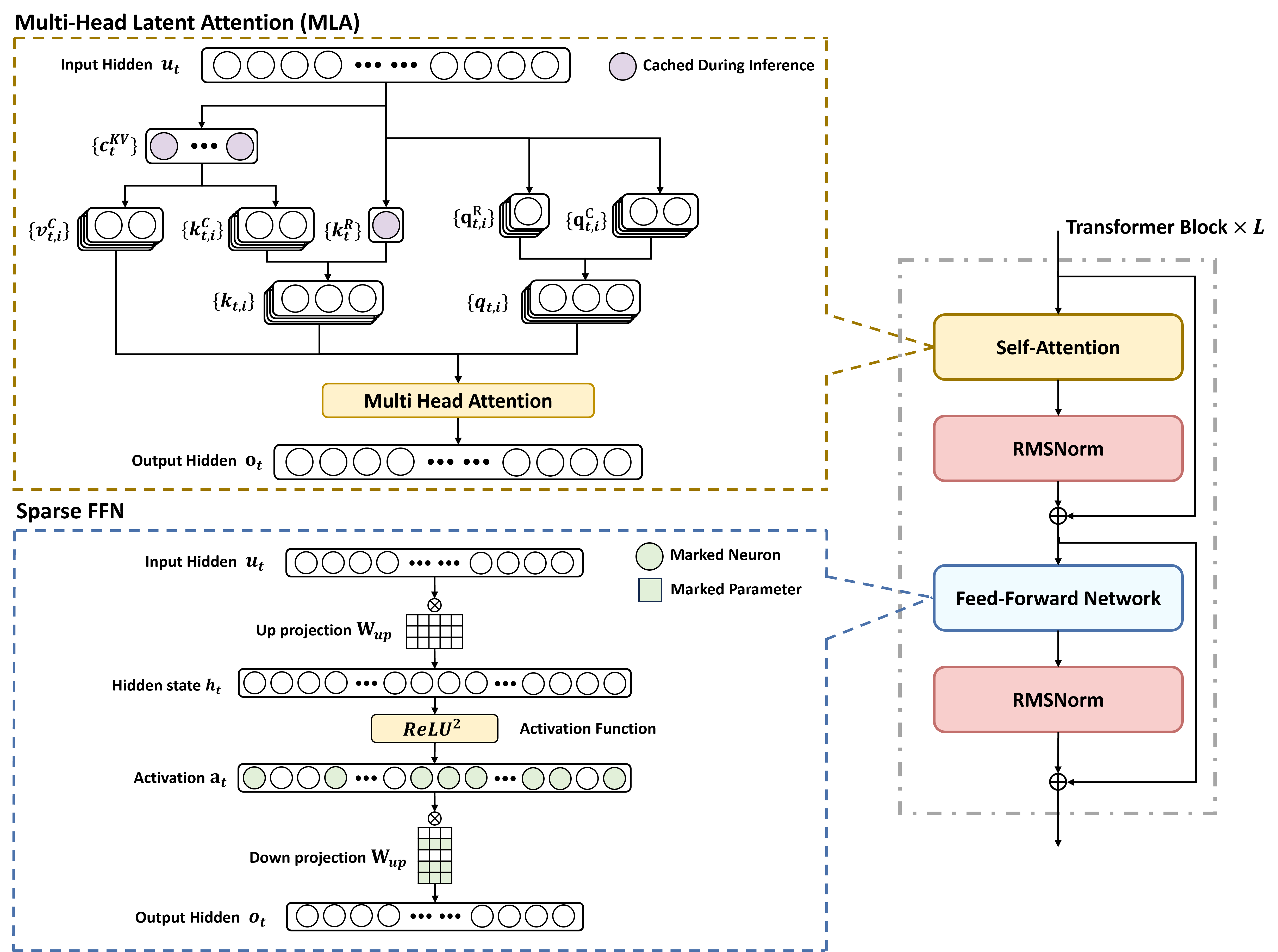}
    \caption{\thellm{} architecture.}
    \label{fig:arch}
\end{figure}

\textbf{Multi-head Latent Attention (MLA)}~\cite{liu2024deepseek} We use MLA in \thellm{}, which is an advanced architectural enhancement designed to improve the efficiency of LLMs by significantly reducing the memory footprint associated with key-value (KV) caching. Conventional multi-head attention mechanisms maintain independent key and value vectors for each attention head, leading to substantial memory consumption. In contrast, MLA employs low-rank matrix decomposition to compress these projections into a shared latent space of reduced dimensionality. This approach enables the model to store a compact representation while reconstructing the necessary vectors dynamically during computation, thereby preserving model performance while substantially lowering memory overhead.

By alleviating memory bottlenecks and improving scalability, MLA enhances the feasibility of deploying LLMs in resource-constrained environments, particularly on edge devices where lightweight, scalable, and computationally efficient models are required to deliver real-time performance. To further optimize \thellm{}, we adapt the original MLA framework by omitting compression on the query projections within the attention mechanism.

In the context of processing the attention input for the \( t \)-th token within an attention layer, MLA introduces \emph{Low-Rank Key-Value Joint Compression} by applying a down-projection to the token's hidden state \( h_t \), formulated as:

\begin{equation}
    \bm{c}^{KV}_t = W_{DKV} \bm{h}_t,
\end{equation}

where \( \bm{c}^{KV}_t \) represents the latent embedding of the keys and values, and \( W_{DKV} \) denotes the down-projection matrix. The key and value representations for the \( i \)-th attention head are subsequently reconstructed using the up-projection matrices \( W_{UK,i} \) and \( W_{UV,i} \), while the query is computed separately using the matrix \( W_{Q,i} \):

\begin{equation}
\bm{k}^C_{t,i} = W_{UK,i} \bm{c}^{KV}_{t,i}, \quad \bm{v}^C_{t,i} = W_{UV,i} \bm{c}^{KV}_{t,i}, \quad \bm{q}^C_{t,i} = W_{Q,i} \bm{h}_t.
\end{equation}

Notably, we deliberately refrains from applying low-rank compression to the queries. While query compression can enhance training efficiency, it introduces additional computational overhead during inference, which is undesirable in latency-sensitive applications. Finally, we integrate the Decoupled Rotary Position Embedding (RoPE) mechanism to further refine the attention dynamics. In summary, \thellm{} leverages MLA with preserving the full-rank structure of the query projections to balance training efficiency and inference speed effectively.

\textbf{Sparse FFN} \;
We adopt a streamlined Feed-Forward Network (FFN) architecture that omits gating mechanisms, thereby reducing computational complexity and memory usage. To further enhance efficiency, 
\emph{ReLU$^2$ activations}~\cite{so2021searching} is an activation function that achieves an optimal trade-off between performance and sparsity, making it highly advantageous for sparse computation in practice. Defined as follows:
\begin{equation}
    \text{ReLU}^{2}(x) = \left(\max(0, x)\right)^2 = \begin{cases}
    x^2, & \text{if } x > 0,\\[1mm]0, & \text{if } x \le 0.\end{cases},
\end{equation}
ReLU$^2$ introduces quadratic growth in activation values, enabling the network to capture richer feature representations. ReLU$^2$ demonstrates exceptional performance by balancing sparsity and computational efficiency, allowing models to dynamically skip inactive neurons while maintaining high accuracy. Furthermore, ReLU$^2$ enhances hardware affinity by consistently activating similar neurons across consecutive tokens, effectively reducing the I/O overhead in feedforward networks.

With the combination of a low-rank attention mechanism and sparse activation function integration, we balance KV cache reduction and computational load across the model, achieving a trade-off between memory cost and latency. Briefly, \thellm{} used squared ReLU without hidden state gating.

In our investigation, the use of MLA in MiniCPM3~\cite{hu2024minicpm} and the adoption of ReLU$^2$ activations in Nemotron-4~\cite{parmar2024nemotron}, and related empirical study~\cite{zhang2024relu} provided compelling evidence supporting this combination strategy. Furthermore, we conducted a series of architecture search experiments, with a particular focus on the relationship between the number of attention heads and the hidden dimension, driven by our implementation of MLA. These insights were instrumental in finalizing the architecture of the \thellm{} model. 

\section{Data}

Compared to the well-established data engineering and accumulation approaches in industry, we employ high-quality open-source datasets alongside research-oriented strategies for data collection and generation. Specifically, we construct chain-of-thought reasoning data, augment data using Retrieval-Augmented Generation (RAG), and distill data from large-parameter language models. We then train \thellm{} in three distinct phases: (1) stable learning of general knowledge, (2) integration of high-quality data, and (3) learning on internal, knowledge-intensive datasets. This three-phase approach yields a comprehensive training corpus for \thellm{}.

\subsection{Data Collection}

\paragraph{Open-source Dataset Collection} 
Following the recommendations of recent surveys~\cite{zhao2023survey,liu2024datasets}, we select open-source datasets based on three key principles: high quality, large volume, and coverage of various domains. A detailed list of the pre-training datasets is provided in~\autoref{pretrain-dataset}. To filter these datasets, we employ the data classifier offered by CCI3-HQ. Specifically, we utilize texts annotated with scores from DCLM and CCI3-HQ, perform score normalization, and train a scoring model. This scoring model is then applied to datasets lacking intrinsic scoring or quality metrics (e.g., MAP-cc), enabling us to sample data according to the assigned scores. Furthermore, we use the MinHash algorithm, as recommended by the Data-Prep-Kit~\cite{wood2024data}, to remove duplicate entries from the Chinese training corpus. Due to the limited time and data processing efficiency, we pose a room for the improvement in the next version of \thellm{}.

\paragraph{Synthetic Data Construction} 
Synthetic data generation has become an increasingly important tool for LLM research and development~\cite{liu2024best}. It mitigates data shortages, helps reduce biases, ensures compliance with privacy regulations, and steers models toward more specialized or robust capabilities. Inspired by Cosmopedia~\cite{benallal2024cosmopedia} and the use of extended Chain-of-Thought reasoning~\cite{qin2024o1,wang2024openr}, we generate an internal dataset for pre-training \thellm{}.

\subsection{Data for Pre-Training}

In the first phase, we train \thellm{} on a pre-training dataset comprising \emph{1.65} trillion tokens in total. The dataset spans four main categories—English, Chinese, Code, and Mathematics. The English dataset (DCLM) is split evenly into five subsets, each containing 280B tokens. The Chinese dataset includes CCI3-HQ (180B), MAP-CC (high-quality; 120B), and MAP-CC (medium-quality; 400B). The Code dataset is derived entirely from StarCoder (230B). Finally, the Mathematics dataset combines Proof-Pile (58B) and MathPile (12B), resulting in a comprehensive training corpus. Detailed configurations of these datasets for Phase 1 are shown in~\autoref{tab:data_pretrain_1}.

\begin{table}[h]
\caption{Summary of training data distribution and allocation of phase 1.}
\label{tab:data_pretrain_1}
\centering
\resizebox{0.7\textwidth}{!}{%
\begin{tabular}{@{}ccccc@{}}
\toprule
\textbf{Category}     & \textbf{Data Source}   & \textbf{Selected Size (B)} & \textbf{Training Allocation (B)} & \textbf{Normalized Ratio} \\ \midrule
\textbf{English} & DCLM                       & 1485    & 966   & 58.3\%                    \\
\textbf{Chinese} & CCI3-HQ, MAP-CC            & 648     & 483   & 29.2\%                    \\
\textbf{Code}         & StarCoder             & 226     & 158   & 9.6\%                     \\
\textbf{Math}  & Proof-Pile, MathPile  & 59      & 48    & 2.9\%                     \\ \bottomrule
\end{tabular}%
}
\end{table}

In the second phase, we train \thellm{} on a higher-quality dataset totaling \emph{550B} tokens, encompassing Chinese general data, English general data, and additional content (e.g., code and math) at a 5:2:2 ratio. The Chinese dataset is downsampled such that the Chinese Fineweb Edu component is reduced to 120B tokens, while the remaining datasets are weighted according to their respective token counts.

During this phase, we also introduce high-quality instruction data, including extended Chain-of-Thought (CoT) data produced by the OpenR~\cite{wang2024openr} framework (0.009B), QA data (0.12B), and Zh\_Baike (0.019B) generated using GPT4o. The primary sources in this phase comprise Chinese Fineweb Edu v2 (120B), the SmolLM Corpus (222B), and the YuLan-Mini-Datasets (65B). Within the SmolLM Corpus, Cosmopedia-v2 (27.2B), FineWeb-Edu (191.4B), and Python-Edu (3.4B) are included. The YuLan-Mini-Datasets encompass deduplicated Python data (36B) and synthetic code, math, and QA data (29B).

Furthermore, we incorporate the Opc-annealing-corpus (64B) and Dolmino-mix-1124 (74B), the latter containing Flan data (17B) and the Pes2o academic paper dataset (60B). Detailed dataset compositions and specifications for Phase 2 are provided in~\autoref{tab:data_pretrain_2}.

% \usepackage{booktabs}
\begin{table}[h]
\caption{Summary of training data distribution and allocation of phase 2.}
\label{tab:data_pretrain_2}
\centering
\resizebox{\textwidth}{!}{%
\begin{tabular}{@{}ccccc@{}}
\toprule
\textbf{Category} & \textbf{Data Source}  & \textbf{Selected Size (B)} & \textbf{Training Allocation (B)} & \textbf{Normalized Ratio} \\ \midrule
\textbf{English} & SmolLM-Corpus, dolmino-mix-1124-Flan   & 292.6                     & 300     & 54.5\%  \\
\textbf{Chinese}       & Chinese-Fineweb-Edu-V2                    & 221                       & 120     & 21.8\%  \\
\textbf{Math, Code, Paper}  & \makecell{opc-annealing-corpus, dolmino-mix-1124-pes2o,\\ YuLan-Mini-Datasets}  & 132.4   & 130     & 23.6\%   \\ \bottomrule
\end{tabular}%
}
\end{table}

% 40 finewebedu-v2
% 11 dolmino-mix
% 10 yulan_infiimmwebmath
% 21 yulancosmppedia
% 100 smollm-fineweb-edu
% 47 scot_internal
% 29 yulan syntheticdata
% 36 yulan the stack
% During Phase 3, we curated a smaller but more diverse dataset to further refine the model's capabilities. This phase incorporates the IndustryCorpus2-HQ dataset, comprising 174.6 billion tokens (136.6 billion in English and 38 billion in Chinese), to enhance general proficiency. Additionally, we utilized an Internal Instruction-Restructured Dataset (44.7 billion tokens, detailed in ~\autoref{internal-pretrain-dataset}) to improve instruction-following performance and alignment with downstream tasks. To preserve balanced bilingual capacity in Chinese and English, the Aquila Dataset (1.6 billion tokens) was included. For advanced reasoning, we leveraged the high-value synthetic YuLan Dataset (32 billion tokens). We also incorporated Dolmino-Mix-1124 (11 billion tokens), a high-quality math-related dataset.

% To maintain general knowledge, we selectively down-sampled datasets such as MathPile, Proof-Pile, CCI3-HQ, and FineWeb-Edu from earlier phases. This comprehensive and strategically refined dataset supports the development of a large language model with broad versatility and strong domain-specific expertise.

% This version enhances readability, emphasizes key datasets, and maintains a professional tone.

% The distribution are summarized as 
% We use 280B for phase 3
During Phase 3, we curated a smaller but more diverse dataset to further refine the model's capabilities. In this phase, we utilized an Internal Instruction-Restructured Dataset, detailed in~\autoref{internal-pretrain-dataset}) to improve instruction-following performance and alignment with downstream tasks, with over \emph{60M} samples. For advanced reasoning, we leveraged the high-value YuLan-Mini-Dataset and selected the corrsponding part. We also incorporated Dolmino-Mix-1124-Math, a high-quality math-related dataset.

To maintain general knowledge, we selectively down-sampled datasets such as FineWeb-Edu from SmolLM-Corpus and Chinese-FineWeb-Edu-V2, an upgraded version. This comprehensive and strategically refined dataset supports the development of an LLM with broad versatility and strong domain-specific expertise. The distribution are summarized in~\autoref{tab:data_pretrain_3}

% This version enhances readability, emphasizes key datasets, and maintains a professional tone.
\begin{table}[h]
\caption{Summary of training data distribution and allocation of phase 3.}
\label{tab:data_pretrain_3}
\centering
\resizebox{\textwidth}{!}{%
\begin{tabular}{@{}ccccc@{}}
\toprule
\textbf{Category} & \textbf{Data Source} & \textbf{Selected Size (B)} & \textbf{Training Allocation (B)} & \textbf{Normalized Ratio} \\ \midrule
\textbf{English} & SmolLM-Corpus & 191 &  95.2 & 34.0\% \\
\textbf{Chinese} &  Chinese-Fineweb-Edu-V2 & 221 & 38.1 & 13.6\%                    \\
\textbf{Math, Code, Paper}& YuLan-Mini-Datasets, dolmino-mix-1124-math & 108 & 101.9 & 36.4\%\\ 
\textbf{Instruction-Restructured} & Internal Dataset & 47 & 44.8 & 16.0\%  \\ \bottomrule
\end{tabular}%
}
\end{table}

\subsection{Data for Post-Training}

During the post-training phase, we employ open-source \textit{instruction–input–output} triplet datasets to perform supervised fine-tuning on \thellm{}, enhancing its ability to follow instructions and generate high-quality responses. Additionally, we incorporate both offline and online preference data to conduct reinforcement learning with human feedback, further refining \thellm{}'s performance.

\paragraph{Supervised Fine-tuning Data} Following the systematic data preparation approach used in K2~\cite{deng2024k2}, we structure the supervised fine-tuning process in a progressive manner. The training of \thellm{} is conducted in two distinct phases:

\begin{itemize}[leftmargin=1.2em] 

\item \textbf{Phase 1 (Basic Instruction Tuning)}: In the initial phase, we utilize \emph{17} general-domain instruction tuning datasets, comprising a total of \emph{407M} samples. These datasets are sourced from publicly available internet corpora or distilled from larger, more capable models. Notably, some datasets contain coding- and math-related tasks, which we retain as they contribute to enhancing the model’s reasoning capabilities.

\item \textbf{Phase 2 (Advanced Instruction Tuning)}: In the second phase, we incorporate \emph{30} datasets, a total of \emph{131M} samples, including three trust-and-safety datasets, ten mathematical datasets, six coding datasets, and ten high-quality, knowledge-intensive instruction tuning datasets. A subset of datasets from Phase 1 is retained to maintain distributional balance within the training data. The inclusion of specialized datasets in this phase aims to further improve the model’s competency in complex reasoning and domain-specific tasks. 

\end{itemize}

\paragraph{RLHF Data}  
During the reinforcement learning stages, our objective is to align \thellm{} with human expression preferences while preserving its performance across various benchmarks. To achieve this,
we first conduct offline fine-tuning of \thellm{} utilizing the  llama-3.1-tulu-3-70b-preference-mixture general preference dataset~\footnote{\url{https://huggingface.co/datasets/allenai/llama-3.1-tulu-3-70b-preference-mixture}}~\cite{lambert2024tulu3}. Following this, we employ the prompt dataset from Ultrafeedback~\footnote{\url{https://huggingface.co/datasets/HuggingFaceH4/ultrafeedback_binarized}}~\cite{cui2023ultrafeedback} to gather on policy data. Combining this with the CodeUltraFeedback dataset~\footnote{\url{https://huggingface.co/datasets/coseal/CodeUltraFeedback_binarized}}~\cite{weyssow2024codeultrafeedback}, we then perform a second round of preference training on \thellm{}. This approach ensures that the model effectively adapts to human-like response preferences without compromising its overall task performance.

% we focus exclusively on alignment using the general preference dataset 
% \href{https://huggingface.co/datasets/allenai/llama-3.1-tulu-3-70b-preference-mixture}{llama-3.1-tulu-3-70b-preference-mixture}~\cite{lambert2024tulu3} and online feedback alignment data~\cite{cui2023ultrafeedback,weyssow2024codeultrafeedback}. This approach ensures that the model effectively adapts to human-like response preferences without compromising its overall task performance.

The details of the supervised fine-tuning data are shown in~\autoref{sft-data}. Moreover, it is important to note that \thellm{} is exclusively trained on Chinese (Simplified) and English instruction corpora and alignment corpora, filtered using \textit{Langid}~\footnote{\url{https://github.com/saffsd/langid.py}} to ensure linguistic consistency.

\newcommand{\methodabb}{ARIES}

\section{Pre-Training Details}

\begin{figure}[t]
    \centering
    \includegraphics[width=\linewidth]{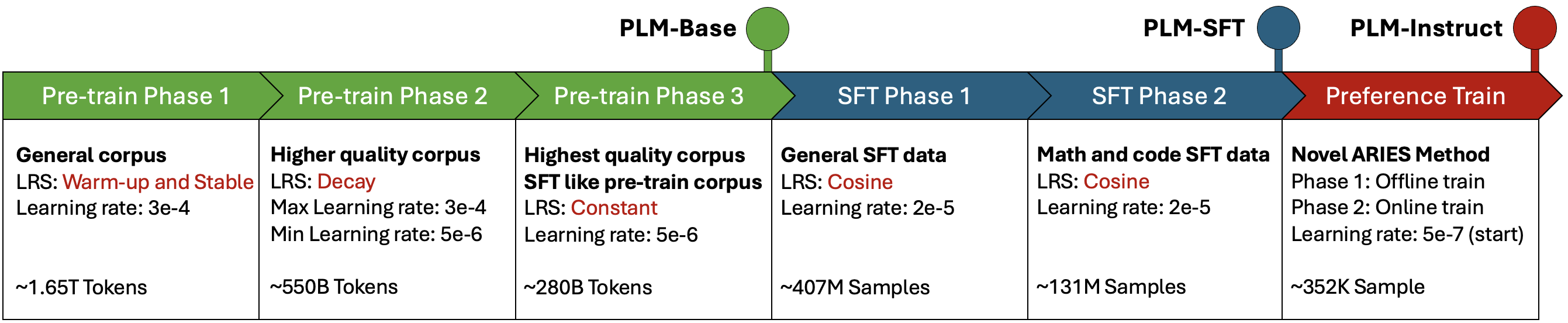}
    \vspace{-2em}
    \caption{Training pipeline adapted for \thellm{}.}
    \label{fig:pipe}
\end{figure}

The \thellm{} pre-training process consists of three core elements: search and selection of an appropriate model architecture, the stage of pre-training, and the stage of post-training. This section provides a detailed illustration of each sessions.

\subsection{Model Architecture Search}

Since \thellm{} uses MLA as the attention mechanism, and the ReLU$^{2}$ as the activation function at the same time, and \thellm{} is designed for peripheral scenarios, we do sandbox experiments for search the best model architecture. Since we target the edge devices, we observe the theoretical amount of floating-point operations (FLOPs), multiply-add operations (MACs), and the empirical performance on 100B specially-selected dataset sampled from the final pre-train corpus. The selected experimented architectures and the results are shown as follows.

% \usepackage{booktabs}
% \begin{table}[h]
% \caption{Experiments for Architecture Search, ~\autoref{arch-search} records more experimental results and detail.~\ly{should double check}}
% \label{tab:arch-search}
% \resizebox{\textwidth}{!}{%
% \begin{tabular}{@{}ccccccccccc@{}}
% \toprule
% \textbf{Params. (w/o vocab)} & \textbf{Params. (w/ vocab)} & $n_{layers}$ & $d_{model}$ & $d_{ffn}$ & $kv_{rank}$ &\textbf{MACs/Param} & \textbf{FLOPS/Param} & \textbf{Time/Layers} & \textbf{Peak Memory} & \textbf{Loss} \\ \midrule
% 1.54B  & 2.01 & 28 & 2816   & 7040  & 256 & 0.5243 & 0.0001261 & 7.3188 & 0.00007031 & 2.639 \\
% 1.21B  & 1.60 & 32 & 2304   & 5760  & 256 & 0.2023 & 0.0005209 & 3.2364 & 0.00007578 & 2.653\\
% 1.47B  & 1.90 & 32 & 2560   & 6400  & 256 & 0.1986 & 0.0003660 & 3.1728 & 0.00006595 & 2.659\\
% 1.36B  & 1.76 & 36 & 2304~\ly{change to: 2048}  & 5760 ~\ly{change to: 6400}  & 256 & 0.1806 & 0.0004142 & 3.2660 & 0.00006802 & 2.652 \\
% 1.51B  & 1.91 & 40 & 2304   & 5760  & 256 & 0.1645 & 0.0003365 & 3.2853 & 0.00007121 & 2.665 \\
% 1.55B  & 1.90 & 36 & 2048   & 8192  & 256 & 0.1785 & 0.0003512 & 3.2225 & 0.00007184 & 2.632\\
% 1.54B  & 1.88 & 32 & 2048   & 8192  & 512 & 0.1722 & 0.0004214 & 2.7499 & 0.00006687 & 2.634\\ \bottomrule
% \end{tabular}
% }
% \end{table}
\begin{table}[h]
\caption{Experiments for architecture search,~\autoref{arch-search} records more experimental results and details.}
\label{tab:arch-search}
\resizebox{\textwidth}{!}{%
\begin{tabular}{@{}c|ccccccccc@{}}
\toprule
\textbf{\#Params \footnotemark} & $n_{layer}$ & $d_{model}$ & $d_{ffn}$ & $kv_{rank}$ &\textbf{MACs/Param} & \textbf{FLOPs/Param} & \textbf{Time/Layer} & \textbf{Peak Memory} & \textbf{Loss} \\ \midrule
1.54B  &  28 & 2816   & 7040  & 256 & 0.5243 & 1.261e-4 & 7.3188 & 7.031e-5 & 2.639 \\
1.21B  &  32 & 2304   & 5760  & 256 & 0.2023 & 5.209e-4 & 3.2364 & 7.578e-5 & 2.653\\
1.47B  &  32 & 2560   & 6400  & 256 & 0.1986 & 3.660e-4 & 3.1728 & 6.595e-5 & 2.659\\
1.36B  &  36 & 2304  & 5760   & 256 & 0.1806 & 4.142e-4 & 3.2660 & 6.802e-5 & 2.653 \\
1.51B  &  40 & 2304   & 5760  & 256 & 0.1645 & 3.365e-4 & 3.2853 & 7.121e-5 & 2.665 \\
1.55B  &  36 & 2048   & 8192  & 256 & 0.1785 & 3.512e-4 & 3.2225 & 7.184e-5 & 2.632\\
1.54B  &  32 & 2048   & 8192  & 512 & 0.1722 & 4.214e-4 & 2.7499 & 6.687e-5 & 2.634 \\ \bottomrule
\end{tabular}
}

\end{table}
\footnotetext{It is refers to non-embedding parameters here.}

\subsection{Pre-training}

We train \thellm{} on a total of \textbf{2.48 trillion tokens}. Refer to~\cite{hu2024minicpm}, during the training, we deploy a WSDC learning rate scheduler, adding one more phase with constant learning rate after the stage of decay, which proceeds in three phases:

\begin{figure}[h]
  \centering
  \subfigure[Entire loss curve.]{
      \label{val_loss_train}
      \includegraphics[width=0.47\linewidth]{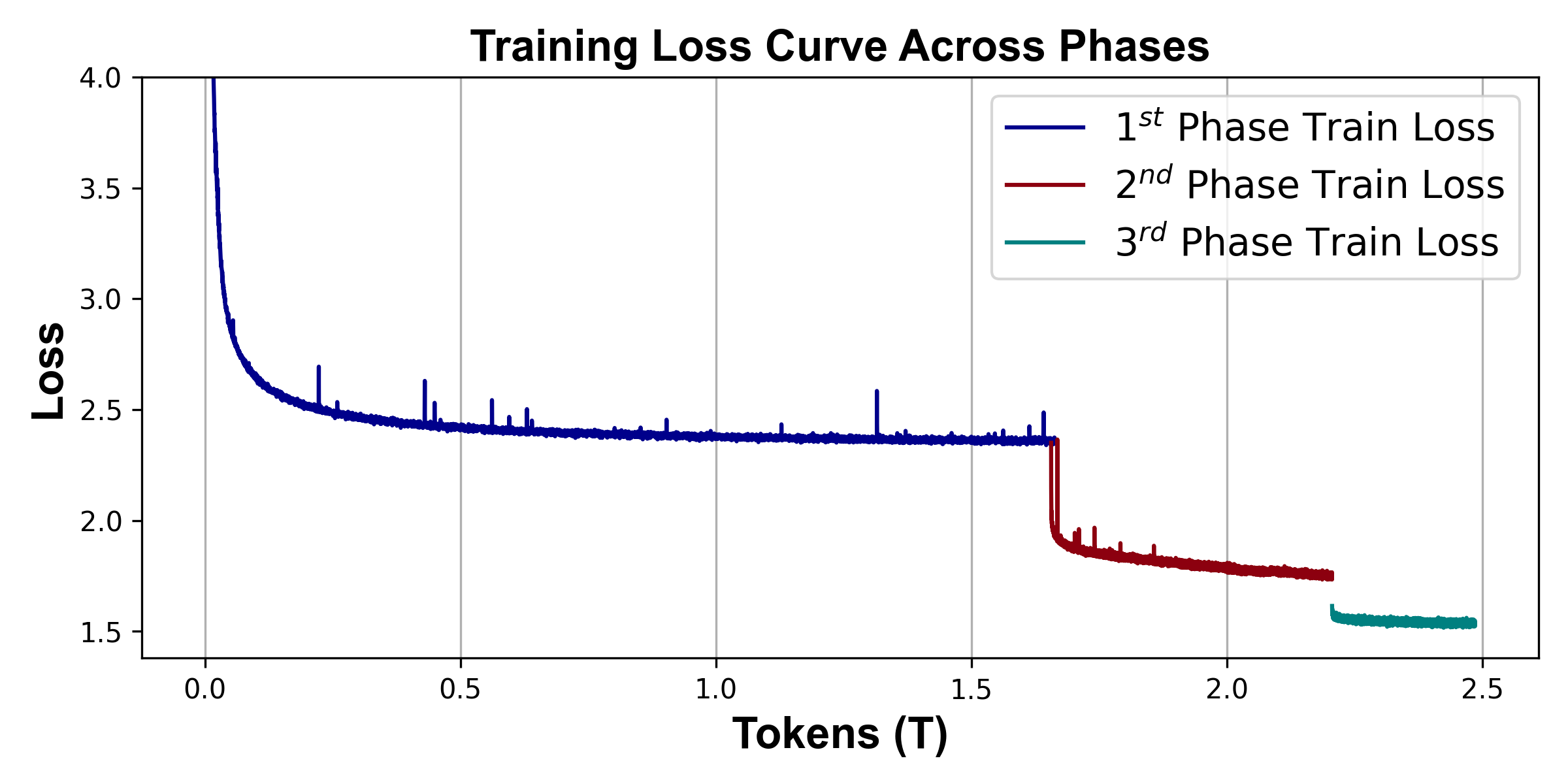}
  }
  \subfigure[Gradient norm curve.]{
      \label{grad_norm}
      \includegraphics[width=0.47\linewidth]{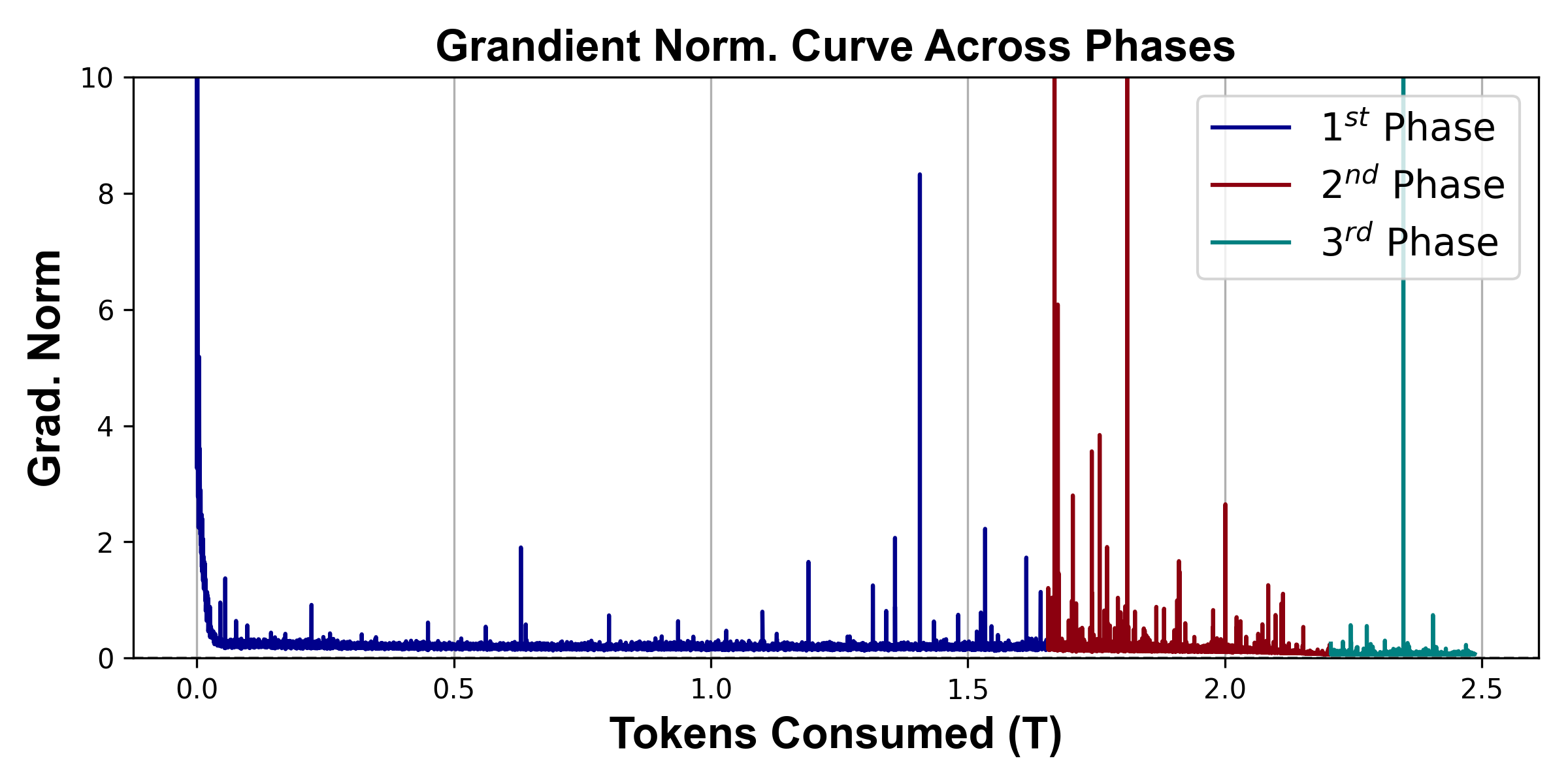}
  }
  \caption{Pre-training loss curve for \thellm{}.}
  \label{curve2}
\end{figure}

\begin{itemize}[leftmargin=1.2em]
\item \textbf{Phase 1 (Warm-up and Stable)}: For 1\% of the total steps, the learning rate gradually increases until it reaches a peak of \emph{3e-4}, which is then maintained.
\item \textbf{Phase 2 (Decay)}: The learning rate is reduced to \emph{3e-5} while the model shifts to smaller, higher-quality datasets.
\item \textbf{Phase 3 (Constant)}: During the final phase of pre-training, spanning the last 280 million tokens, we utilize a Constant learning rate, maintaining a context length of 4,096 tokens. In this phase, we introduced instruction-like data and significantly up-sampled the high-quality data from phase 2.
\end{itemize}

\begin{figure}[h]
  \centering
  \subfigure[Loss curve for phase 1.]{
      \label{phase1_image}
      \includegraphics[width=0.3\linewidth]{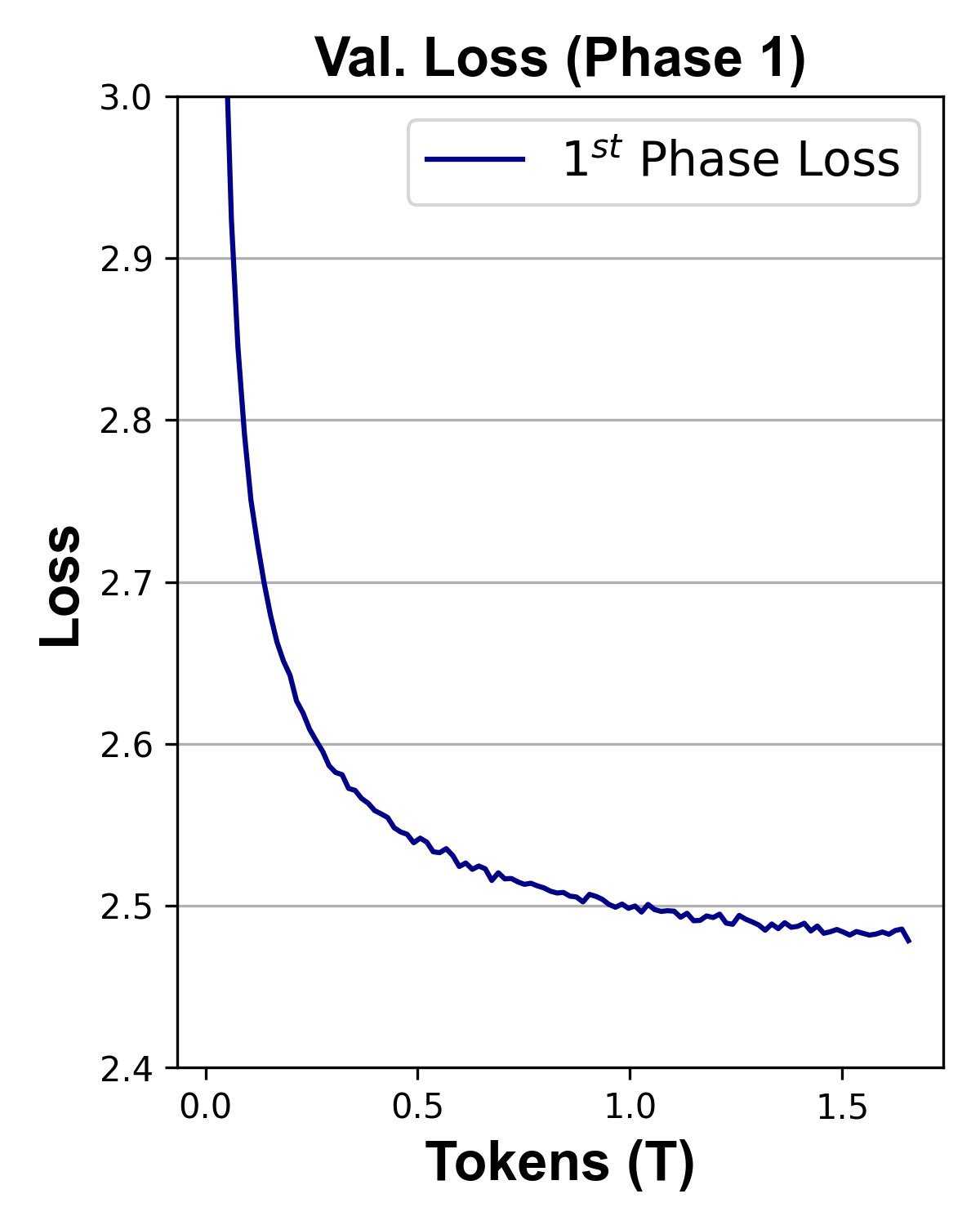}
  }
  \subfigure[Loss curve for phase 2 \& 3.]{
      \label{phase2_image}
      \includegraphics[width=0.3\linewidth]{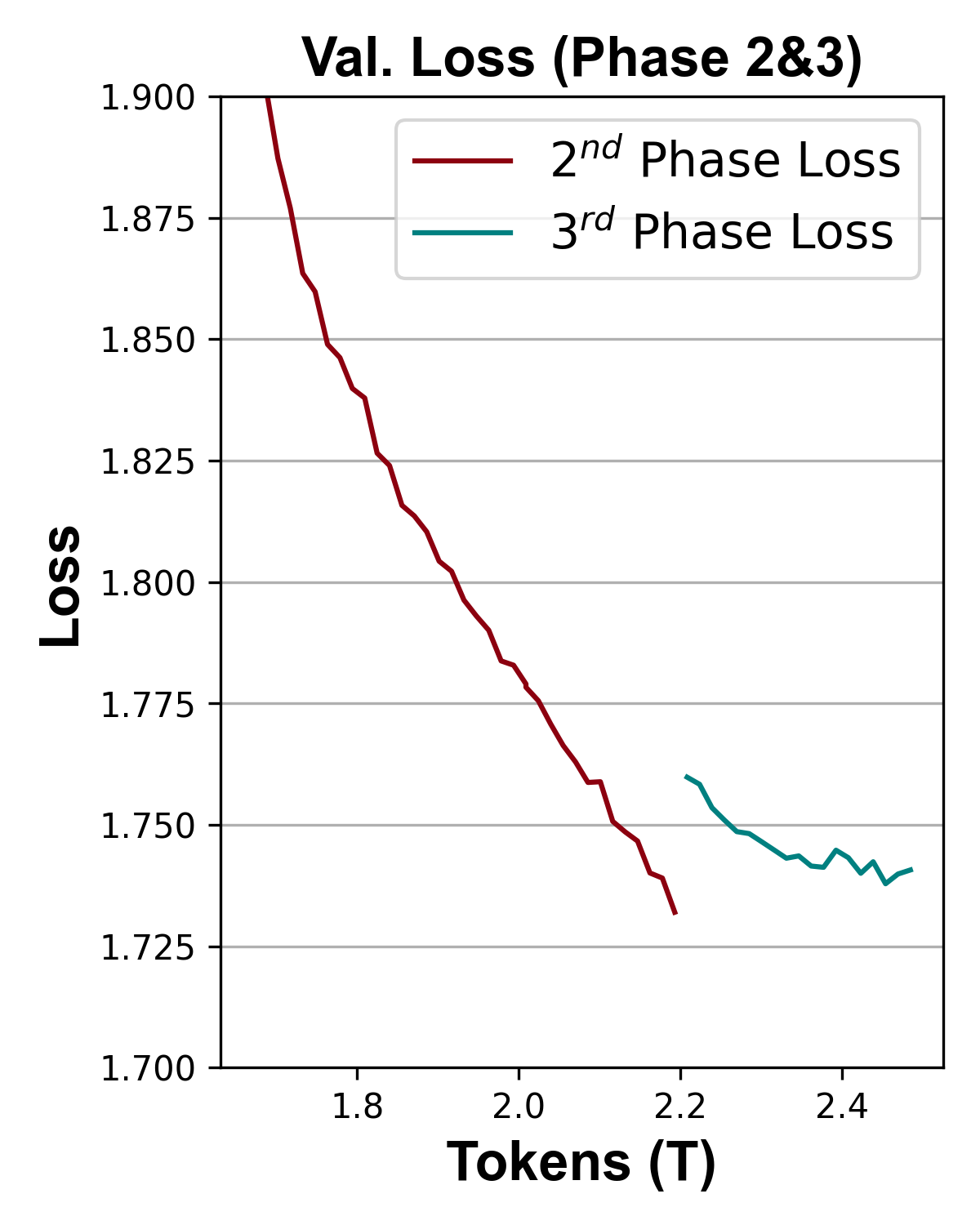}
  }
  \subfigure[Learning rate during training.]{
      \label{phase3_image}
      \includegraphics[width=0.3\linewidth]{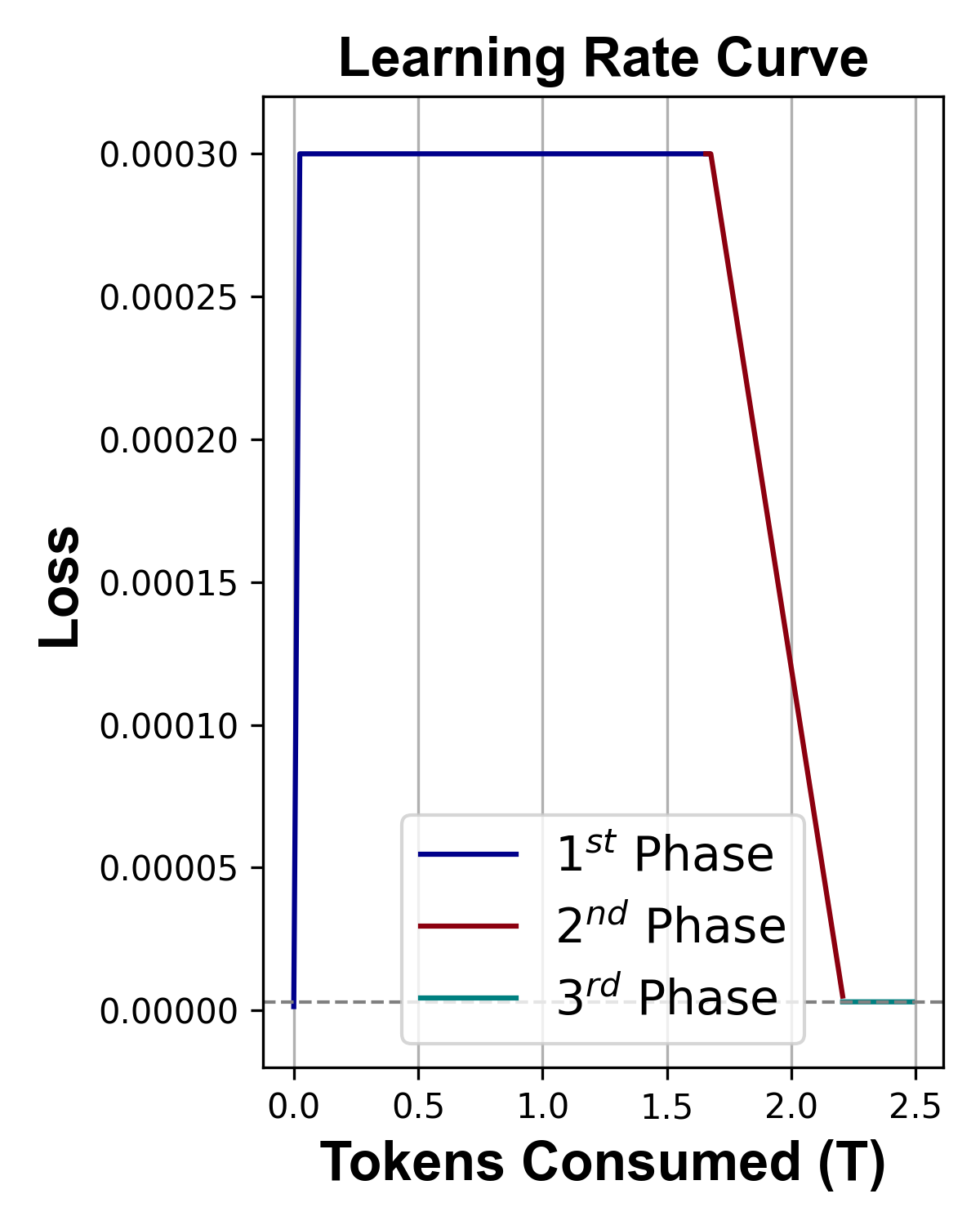}
  }
  \caption{Phase level pre-training loss curve for \thellm{}.}
  \label{curve}
\end{figure}

Throughout training, we maintained a fixed sequence length of 4K tokens and utilized a batch size of 15 million tokens, distributed across 208 accelerators. The training curves for \thellm{}-1.8B-Base are presented in~\autoref{curve}. The details of the pre-training are introduced in~\autoref{pre_args}.
% \newpage

During phase 1 of training, we continuously monitored the validation loss to evaluate the model's performance. Following the methodology outlined in~\cite{du2024understanding}, we estimated \thellm{}'s performance throughout training and reported its improvements alongside the total number of tokens processed. 

As illustrated in~\autoref{phase1_image}, the model converged to a validation loss exceeding \emph{2.3} (The majority of emergent abilities manifest when the training loss is below \emph{2.2}~\cite{du2024understanding}). Given the constraints of the initial dataset, we opted to incorporate a higher-quality dataset for subsequent training phases, as suggested in~\cite{dong2024hymba}. 

Building on these WSD stages, we extended pre-training to enhance the model's expertise in mathematics, coding, and encyclopedic knowledge, thereby boosting its performance on knowledge-intensive tasks.

During the phase 3, inspired by the training strategies of ResNet~\cite{he2016deep}, we adjusted the learning rate to \emph{3e-5} and initiated a cosine decay learning rate schedule. This adjustment guided the training process through an additional 280 billion tokens, culminating in the final stage of model optimization. 
\subsection{Supervised Fine-tuning}

In the the stage of supervised fine-tuning, we refer to both data-centric model training~\cite{deng2024k2} and empirical learning on SFT and RL~\cite{chu2025sft}, we deploy a two-phase training in a progressive manner. For each phase we utilize LLaMA-Factory~\cite{zheng2024llamafactory} to do full-parameter SFT over prepared datasets. The detail training arguments are shown in~\autoref{sft_args}.

During the training, in Phase 1, we utilize a diverse range of general-purpose instruction-following datasets, including those focused on chat, web questions, and general instruction. In Phase 2, we expand upon this with a significant emphasis on specialized domains, particularly mathematics and coding, incorporating datasets designed for complex reasoning, code generation, and domain-specific instruction following in areas like education and medicine. This progression suggests a strategy of initially building a strong foundation in general capabilities before refining the model's expertise in specialized tasks. 

\begin{figure}[h]
  \centering
  \subfigure[Training loss curve of 1$^{st}$ Phase SFT.]{
      \label{sftcurve1}
      \includegraphics[width=0.47\linewidth]{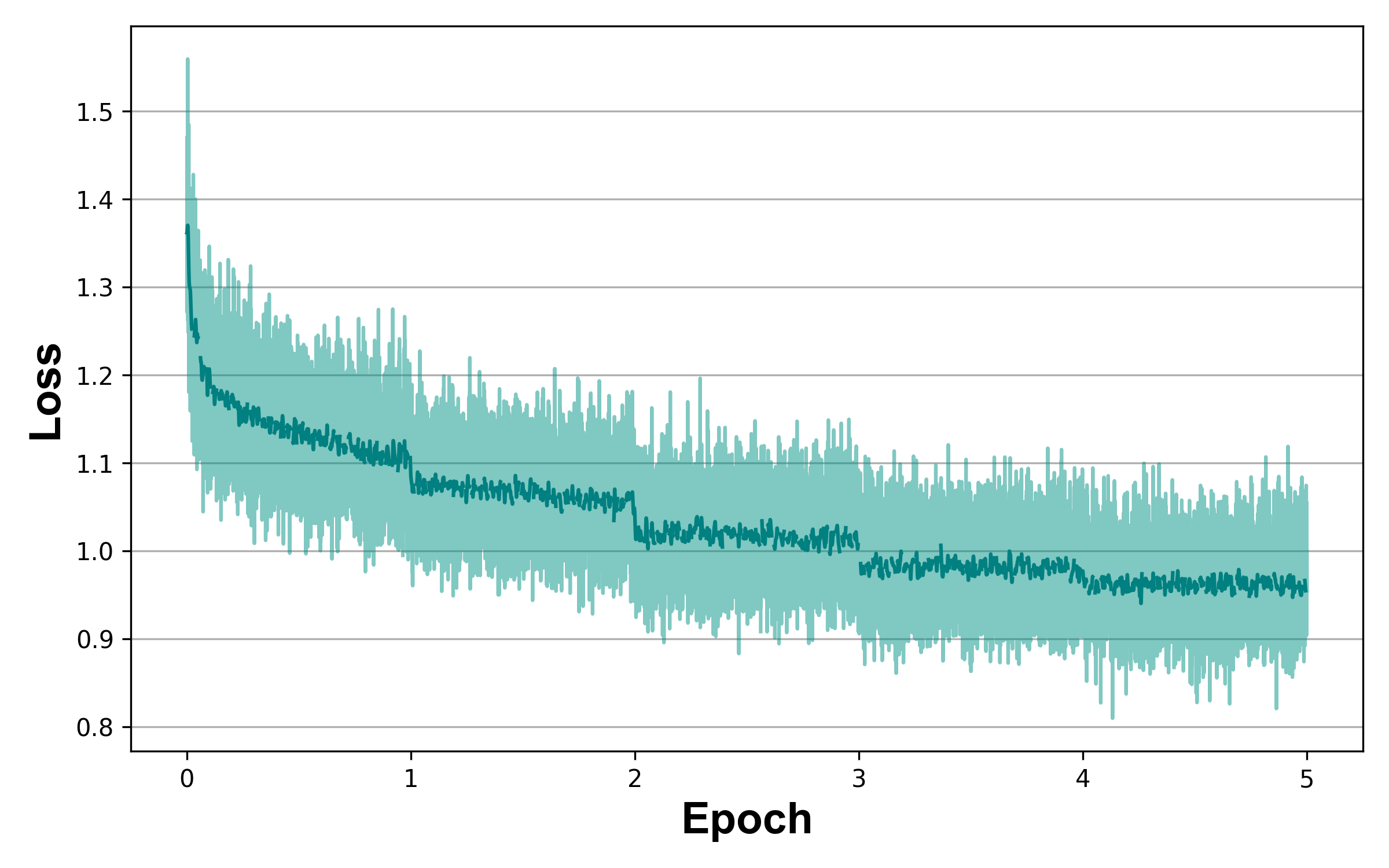}
  }
  \subfigure[Training loss curve of 2$^{nd}$ Phase SFT.]{
      \label{sftcurve2}
      \includegraphics[width=0.47\linewidth]{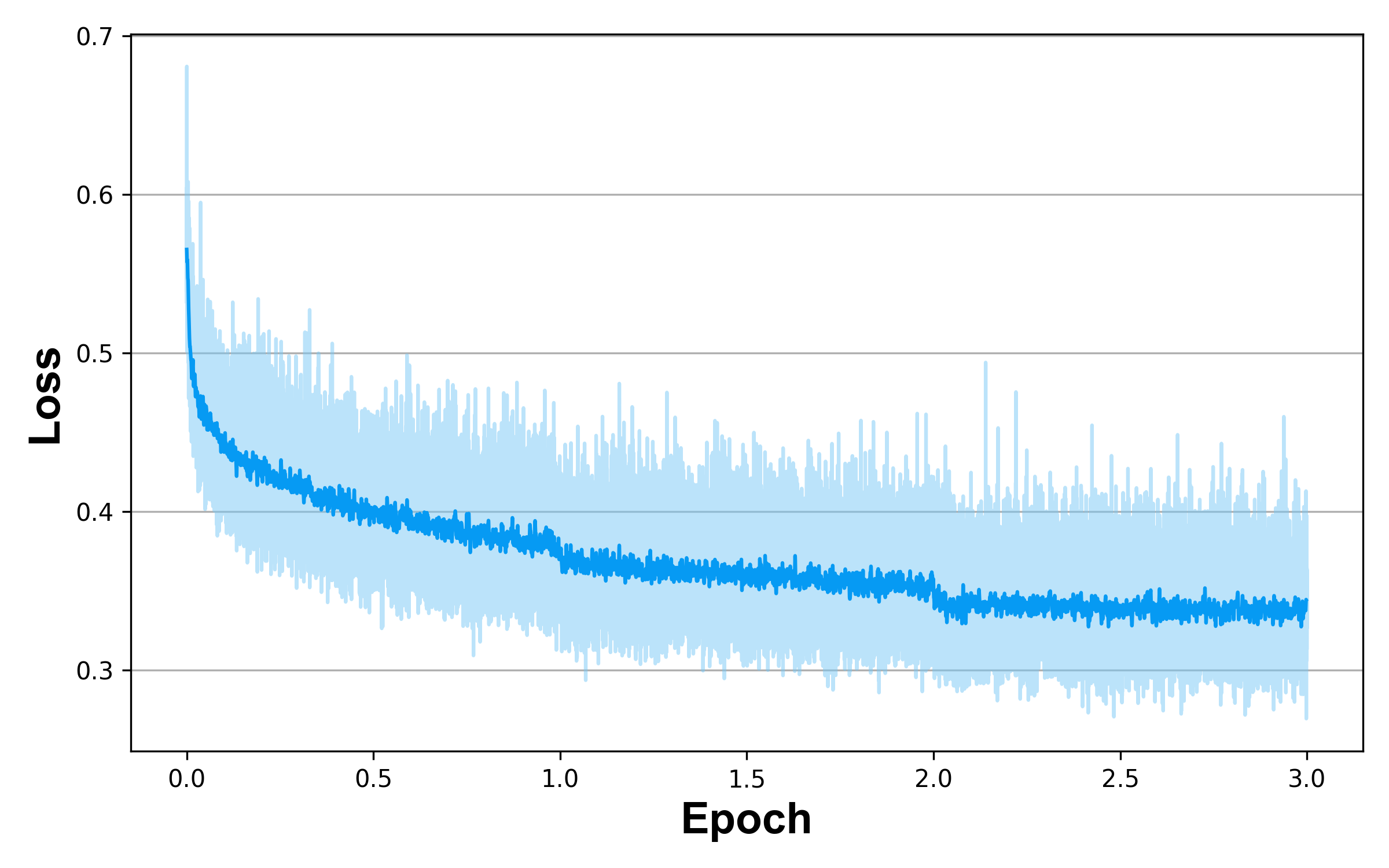}
  }
  \caption{SFT training curve for \thellm{}.}
  \label{sftloss}
\end{figure}

Finally, we obtain the \thellm{}-Instruct, and the~\autoref{sftloss} shows the curves during this two phases SFT training.

\subsection{Preference Training}

Previous open-source LLMs have exhibited a significant limitation: their performance often degrades markedly after multiple rounds of self-refinement. This behavior is inconsistent with the expected characteristics of highly intelligent language models. To address this issue, during the preference training phase, we adopt a novel training approach akin to our method, ARIES~\cite{zeng2025ariesstimulatingselfrefinementlarge}. We leverage ARIES' self-refinement loss to conduct preference training on \thellm{}:
\begin{equation}
\begin{aligned}
\mathcal{L}_{\mathrm{\methodabb{}}}(\pi_{\theta};\pi_{\mathrm{ref}}) =(1- \alpha) \mathcal{L}_{\mathrm{DPO}}(\pi_{\theta};\pi_{\mathrm{ref}}) + \alpha\mathcal{L}_{\mathrm{refine}}(\pi_{\theta};\pi_{\mathrm{ref}}), 
\end{aligned}
\label{loss}
\end{equation}
where the refinement loss is employed to stimulate and strengthen the model's self-refinement capability:
\begin{equation}
\small
\begin{aligned}
\mathcal{L}_{\mathrm{refine}}(\pi_{\theta};\pi_{\mathrm{ref}})& =\underset{(x,y_{w},y_{l})\sim\mathcal{D}}{\mathbb{E}}\left[\frac{1}{2}-\beta\left[\log\left(\frac{\pi_{\theta}(y_{w}|x, y_{l},z)}{\pi_{\mathrm{ref}}(y_{w}|x, y_{l},z))}\right)-\log\left(\frac{\pi_{\theta}(y_{l}|x, y_{l},z)}{\pi_{\mathrm{ref}}(y_{l}|x, y_{l},z)}\right)\right]\right]^{2}  \\
&+\underset{(x,y_{w},y_{l})\sim\mathcal{D}}{\mathbb{E}}\left[\frac{1}{2}-\beta\left[\log\left(\frac{\pi_{\theta}(y_{w}| x, y_{w},z)}{\pi_{\mathrm{ref}}(y_{w}|x, y_{w},z))}\right)-\log\left(\frac{\pi_{\theta}(y_{l}|x, y_{w},z)}{\pi_{\mathrm{ref}}(y_{l}|x, y_{w},z)}\right)\right]\right]^{2} .
\end{aligned}
\label{loss:refine}
\end{equation}
In the above loss function, \(x\) denotes the input prompt, \(y_w\) is the chosen response, \(y_l\) is the rejected response, and \(\mathcal{D} = \{(x^{(i)}, y_w^{(i)}, y_l^{(i)})\}_{i=1}^N\) represents the preference dataset, \(z\) denotes the refinement template, while \(\pi_{\mathrm{ref}}\) and \(\pi_{\theta}\) represent the reference model and the learnable policy model, respectively. The parameter \(\beta\) controls the strength of the regularization term, which penalizes deviations from the reference model \(\pi_{\mathrm{ref}}\).
% where $\mathcal{D} = \{x^{(i)}, y_w^{(i)}, y_l^{(i)}\}_{i=1}^N$ is the given preference dataset, $x$ is the prompt, $y_l$ is the rejected response and $y_w$ is the chosen response, $z$ represents the refinement template, $\pi_{\mathrm{ref}}$ denotes the reference model and  $\pi_{\theta}$ is the policy model. $\beta$ is a parameter controlling the deviation from the reference model $\pi_{\mathrm{ref}}$.
This loss function also demonstrates an auxiliary effect in improving the model's direct answering ability in experiments.

% During the data collection phase, since we did not perform Supervised Fine-tuning similar to ARIES-SFT, the dataset collected through multiple parallel sampling demonstrated superior quality compared to that obtained via self-refinement. Consequently, we employed multiple parallel sampling to construct the preference dataset for preference training at this stage.

The overall training pipeline for preference training follows a similar approach to ARIES. We first conduct an offline preference training round on the llama-3.1-tulu-3-70b-preference-mixture dataset~\footnote{\url{https://huggingface.co/datasets/allenai/llama-3.1-tulu-3-70b-preference-mixture}}~\cite{lambert2024tulu3} (\textbf{Phase 1}), and then sample 5K prompt dataset from the UltraFeedback dataset~\footnote{\url{https://huggingface.co/datasets/HuggingFaceH4/ultrafeedback_binarized}}~\cite{cui2023ultrafeedback} to gather on policy dataset. Combining this with the CodeUltraFeedback preference dataset~\footnote{\url{https://huggingface.co/datasets/coseal/CodeUltraFeedback_binarized}}~\cite{weyssow2024codeultrafeedback}, we perform a second round of preference training on \thellm{} (\textbf{Phase 2}). During the data collection process, we employ the RLHFlow/ArmoRM-Llama3-8B-v0.1~\footnote{\url{https://huggingface.co/RLHFlow/ArmoRM-Llama3-8B-v0.1}}~\cite{ArmoRM} as the reward model to construct the preference dataset. The basic argument settings and training curves are shown in~\autoref{rl_args}.

% Similar to DPO~\cite{rafailov2023direct}, we define the implicit reward as $\hat{r}(x, y) = \beta \log \frac{\pi_{\theta}(y|x)}{\pi_{\mathrm{ref}}(y|x)}$, and the implicit reward accuracy is defined as the probability that $\beta \log \frac{\pi_{\theta}(y_w|x)}{\pi_{\mathrm{ref}}(y_w|x)} > \beta \log \frac{\pi_{\theta}(y_l|x)}{\pi_{\mathrm{ref}}(y_l|x)}$ holds for the preferred response $y_w$ over the dispreferred response $y_l$.
% The implicit reward accuracy curves and basic training arguments and training loss curves are shown in~\autoref{rl_args}.

\section{Evaluation: Training Results}
\label{expall}

As an LLM tailored for edge-based peripheral computing, we first show the results of the training results of \thellm{}, then we perform a comprehensive evaluation from three distinct perspectives: fundamental model performance, architectural efficiency, and real-world operational efficiency. For each aspect, we select appropriate baseline models for comparison. Specifically, for performance evaluation, we rely on quantitative results obtained directly from system feedback.

\subsection{Evaluation and Experiment Setup}

To evaluate the model trained during the process, assess the effectiveness of the proposed methods, and compare them with baseline models, we conduct a comprehensive evaluation of the LLM's performance. To ensure a thorough assessment, we select benchmarks based on the following aspects.

\begin{itemize}[leftmargin=1.2em]
\item Language Modeling: We utilize the subset of DCLM, MAP-cc, StarCoder, and ProofPile, to evaluate the improvement along with the \thellm{} training.
\item General Knowledge Comprehension: We utilize widely-adopted benchmarks such as the Massive Multitask Language Understanding (MMLU), Chinese Massive Multitask Language Understanding (CMMLU), Chinese Evaluation (C-Eval), and the AI2 Reasoning Challenge (ARC), which includes both the ARC-Easy (ARC-E) and ARC-Challenge (ARC-C) sets.
\item Mathematical Problem Solving: To assess mathematical reasoning, we employ benchmarks like the Grade School Math 8K (GSM8K) and MathQA datasets.
\item Coding Proficiency: We evaluate coding capabilities using the HumanEval and Mostly Basic Python Problems (MBPP) benchmarks.
\item Commonsense and Logical Reasoning: The model's ability to understand and apply everyday commonsense knowledge is tested using datasets such as HellaSwag, BoolQ, LogiQA, and the Physical Interaction: Question Answering (PIQA) benchmark.
\end{itemize}

These benchmarks provide a diverse and rigorous framework to assess the model's performance across multiple dimensions. For fair evaluation and comparison in equal, we set \textit{zero-shot} for HumanEval, MBPP, BoolQ, LogiQA, and PIQA, set MMLU, CMMLU, C-Eval, MathQA, and WinoGrande as \textit{5-shots} tasks, HellaSwag as \textit{10-shots} task, and ARC-c and ARC-e as \textit{25-shots} tasks. For evaluation consistency, we use EvalPlus~\cite{evalplus} for the evaluation of Coding Proficiency, and utilize the LM-Evaluation Harness~\cite{evalharness} to assess the performance of \thellm{} and all the models across the rest of aforementioned tasks.

\subsection{Evaluating Model Performance during Training}

In this section, we summarize the key findings from the phased training and evaluation of \thellm{}, integrating insights from the provided perplexity measurements and benchmark results across pre-training, supervised fine-tuning (SFT), and reinforcement learning (RL) stages.

\paragraph{Phased Pre-Training Analysis}

As shown in Table~\ref{tab:ppl-observation}, perplexity consistently decreased across all training corpora (DCLM, MAP-cc, StarCoder, ProofPile) through the three-phase curriculum pre-training. Phase 3 achieved the lowest perplexity (e.g., 6.75 for ProofPile), indicating progressive improvements in the model’s ability to predict token sequences across diverse domains. This reduction reflects effective knowledge consolidation, particularly for code and mathematical reasoning tasks, where StarCoder and ProofPile saw substantial gains.

% \usepackage{booktabs}
\begin{table}[h]
\caption{Perplexity observation during the three-phases pre-training.}
\label{tab:ppl-observation}
\centering
\resizebox{0.5\textwidth}{!}{%
\begin{tabular}{@{}cccccccc@{}}
\toprule
\textbf{Phase}   & \textbf{DCLM} & \textbf{MAP-cc} & \textbf{StarCoder} & \textbf{ProofPile}  \\ \midrule
\textbf{Phase 1} &   15.59   &    11.47     &  3.13    &   7.60  \\
\textbf{Phase 2} &   14.71   &    10.54     &  2.91    &   7.16  \\
\textbf{Phase 3} &   \textbf{14.61}   &    \textbf{10.47}     &  \textbf{2.89}    &   \textbf{6.75}  \\ \bottomrule
\end{tabular}
}
\end{table}

\paragraph{Supervised Fine-Tuning Results}

During the SFT stage, we mainly focus on the performance on MMLU, CMMLU, BoolQ, GSM8K, and HumanEval. Table~\ref{tab:res_sft} highlights performance gains during SFT. While the base model (\thellm{}-Base) achieved an average score of 51.29 across benchmarks, phased SFT yielded incremental improvements. Notably:

\begin{itemize}[leftmargin=1.2em]
\item Phase 2 SFT achieved the highest average score (57.00), with significant gains in reasoning tasks: GSM8K improved by 26 points (34.80 $\rightarrow$ 60.80) compared to earlier phases, and HumanEval (code generation) rose from 43.30 to 57.90.

\item Mathematical reasoning (MathQA) and language understanding (BoolQ, ARC-c) also improved, though MBPP (code) saw a temporary dip in Phase 1, recovering in later phases. This suggests task-specific sensitivity to curriculum ordering.
\end{itemize}

% \usepackage{booktabs}
\begin{table}[h]
\centering
\caption{SFT results of each phase.}
\label{tab:res_sft}
\resizebox{\textwidth}{!}{%
\begin{tabular}{@{}c|cccccccccc@{}}
\toprule
\textbf{Model} & \textbf{Avg.}  & \textbf{MMLU} & \textbf{CMMLU} & \textbf{ARC-c} & \textbf{ARC-e} & \textbf{BoolQ} & \textbf{GSM8K} & \textbf{MathQA} & \textbf{HumanEval} & \textbf{MBPP}  \\ \midrule
\textbf{\thellm{}-Base}  & 51.29 & 48.02   & 45.40  & 48.12     & 77.31 & 73.67 & 41.70 & 28.81     & 43.30      & 55.30  \\
\textbf{\thellm{}-SFT-Tulu}  & 51.49 & 47.67   & 45.16    & 48.63     & 78.24 & 73.79 & 34.80    & 28.71     & 54.90     &  54.50 \\
\textbf{\thellm{}-SFT-Phase-1} & 52.22 & 48.99   & 46.45    & \textbf{51.54}     & \textbf{79.00} & 76.70 & 35.86    & 29.55     & 52.40     & 49.50 \\
\textbf{\thellm{}-SFT-Phase-2} & \textbf{57.00} & \textbf{49.79}   & \textbf{46.86}    & 49.57     & 77.98 & \textbf{76.97} & \textbf{60.80}     & \textbf{32.53}     & \textbf{57.90}      & \textbf{60.60}  \\ \bottomrule
\end{tabular}
}
\end{table}

\paragraph{Reinforcement Learning Enhancements}

% During the Reinforcement learning after training the model to follow the instruction, deploy the curriculum training using ARIES~\cite{zeng2025ariesstimulatingselfrefinementlarge}. We can see the result from the , the performance of \thellm{} becomes better and better in majority tasks.

After SFT, Our novel RL methods, ARIES~\cite{zeng2025ariesstimulatingselfrefinementlarge}, further refined performance (Table~\ref{tab:rlhf_eval}):

\begin{itemize}[leftmargin=1.2em]
\item Online RLHF and \thellm{}-Instruct achieved the best overall averages (55.99 and 56.94, respectively), with GSM8K reaching 63.50 (up 5.4 points from SFT) and HumanEval stabilizing at 64.63.
\item Identity RL underperformed compared to Online/Offline RLHF, emphasizing the importance of preference-based alignment. Notably, while most tasks improved, MathQA showed minimal gains, indicating limitations in arithmetic generalization.
\end{itemize}

% \usepackage{booktabs}
% \usepackage{booktabs}
\begin{table}[h]
\caption{Reinforcement learning result of each iteration.}
\label{tab:rlhf_eval}
\resizebox{\textwidth}{!}{%
\begin{tabular}{@{}c|cccccccccc@{}}
\toprule
& \textbf{Avg} & \textbf{MMLU} & \textbf{C-Eval} & \textbf{ARC-c} & \textbf{ARC-e} & \textbf{BoolQ} & \textbf{GSM8K} & \textbf{MathQA} & \textbf{HumanEval} & \textbf{MBPP} \\ \midrule
\textbf{\thellm{}-1.8B-SFT}                 & 56.66 & 49.79 & 43.76 & 49.57 & 77.98 & 76.97 & 60.80 & 32.53 & 57.90 & \textbf{60.60} \\
\textbf{\thellm{}$_{\text{Identity RL}}$}   & 56.78 & 49.80 & 44.50 & 50.43 & 77.99 & 74.04 & 57.16 & 32.86 & 64.00 & 60.30 \\ 
\textbf{\thellm{}$_{\text{Offline RLHF}}$}  & 57.51 & 50.08 & 43.83 & 50.94 & 77.78 & 77.16 & 60.73 & 32.63 & 64.60 & 59.80 \\
\textbf{\thellm{}$_{\text{Online RLHF}}$}   & \textbf{58.42} & \textbf{50.47} & \textbf{44.76} & \textbf{52.65} & \textbf{78.68} & \textbf{77.67} & \textbf{63.50} & \textbf{33.11} & \textbf{64.63} & 60.32 \\
\bottomrule
\end{tabular}
}
\end{table}

In general, the phased training curriculum—combining progressive pre-training, task-focused SFT, and preference-aware RLHF enable \thellm{} to achieve robust performance across reasoning, coding, and language understanding benchmarks. Key gains in GSM8K and HumanEval highlight strengths in mathematical and code generation tasks, while fluctuations in MBPP and MathQA suggest avenues for future refinement. These results underscore the efficacy of curriculum-driven training and strategic alignment for specialized capabilities.

\subsection{Evaluating \thellm{} and the baselines}

To compare the performance comprehensively, we introduce several small size LLMs with comparable scales to \thellm{} (Under 2.7B, including embedding sizes) as baselines for comparison, we choose three data-close-source models and four data-open-source models, which are listing as follows.

% \usepackage{booktabs}
% \usepackage{multirow}
\begin{table}[h]
\centering
\caption{Basic information of the selected baselines.}
\label{tab:comp}
\resizebox{0.9\textwidth}{!}{%
\begin{tabular}{@{}c|cccccccc@{}}
\toprule
\textbf{Model} & \textbf{Data}  & \textbf{Size}  & \textbf{$n_{layer}$} & \textbf{$d_{model}$} & \textbf{$n_{head}$} & \textbf{Attention} &  \textbf{Activation}  & \textbf{Institute} \\ \midrule
\textbf{MiniCPM-2.7B}      & Close (1.3T)  & 2.74B & 40 & 2304 & 36 & GQA & SwiGLU & ModelBest \\
\textbf{Yulan-Mini-2.4B}   & Open (1.1T)   & 2.42B & 56 & 1920 & 30 & GQA & SwiGLU & RUC\\
\textbf{Index-1.9B}      & Open (2.8T)   & 1.91B & 36 & 2048 & 16 & GQA & SwiGLU & Bilibili \\
\textbf{SmolLM2-1.7B}      & Open (11T)  & 1.71B & 24 & 2048 & 32 & GQA & SwiGLU & Hugging face \\
\textbf{StableLM2-1.6B}    & Open (2T)  & 1.64B & 24 & 2048 & 32 & MHA & SwiGLU & StableAI  \\ 
\textbf{Fox-1-1.6B} & Open (3T)  & 1.61B & 32 & 2048 & 32 & GQA & SwiGLU & TensorOpera \\
\textbf{GLM-Edge-1.5B}     & Close (-) & 1.59B & 28 & 2048 & 16 & GQA & SwiGLU & Zhipu   \\
\textbf{Qwen2.5-1.5B}      & Close (18T)  & 1.54B & 28 & 1536 & 12 & GQA & SwiGLU & Alibaba   \\
\textbf{Qwen2-1.5B} & Close (7T)  & 1.54B & 28 & 1536 & 12 & GQA & SwiGLU & Alibaba   \\ \midrule
\textbf{\thellm{}-1.8B}   & Open (2.5T)   & 1.83B & 32 & 2048 & 16 & MLA & ReLU$^2$ & \thellm{}-Team  \\ \bottomrule
\end{tabular}
}
\end{table}

\begin{itemize}[leftmargin=1.2em]
\item \textbf{Qwen2-1.5B \& Qwen2.5-1.5B~\cite{yang2024qwen2,yang2024qwen2_5}:} We introduce the Qwen series models, particularly Qwen2-1.5B, pre-trained on 7T tokens, and Qwen2.5-1.5B, pre-trained on 18T tokens. These models exhibit strong performance in both general-purpose and domain-specific tasks, although detailed training methodologies remain undisclosed.
\item \textbf{MiniCPM-2.4B~\cite{hu2024minicpm}:} MiniCPM-2.4B, pre-trained on 1.06T tokens, employs an annealing training strategy. Despite its relatively modest parameter count, it delivers impressive performance across general tasks and is optimized for deployment on resource-constrained hardware.
\item \textbf{GLM-Edge-1.5B~\cite{glmedge}:} Released by GLM team, GLM-Edge-1.5B is designed to address the challenges of deploying real-world applications on edge devices. It offers several models in different sizes tailored for natural language multimodal understanding. However, the training data and details are not open-sourced.
\item \textbf{Fox-1-1.6B~\cite{hu2024fox}:} Developed by TensorOpera, Fox-1-1.6B is pre-trained on 3T tokens of web-scraped document data using an innovative three-stage data curriculum training strategy.
\item \textbf{Index-1.9B~\cite{Indexbilibili}:} Developed by Bilibili, Index-1.9B is a lightweight version of the Index models, featuring the base model—which leads benchmarks with 1.9 billion non-embedding parameters and a 2.8T Chinese-English corpus, Index-1.9B has two version, one is trained on the open-sourced corpus named with ``pure'', and the other is trained over an close-sourced dataset, named ``base''. We choose Index-1.9B-base as the baseline.
\item \textbf{StableLM2-1.6B~\cite{bellagente2024stable}:} Proposed by StabilityAI, StableLM2-1.6B leverages a blend of open-source datasets for pre-training. Its data selection process is informed by insights from smaller language models to optimize dataset composition.
\item \textbf{SmolLM2-1.7B~\cite{allal2024SmolLM2}:} Developed by Hugging Face TB Research, SmolLM2-1.7B is pre-trained on an 11T-token high-quality corpus. It achieves a fine balance between computational efficiency and task performance.
\item \textbf{Yulan-Mini-2.4B~\cite{hu2024yulan}:} Yulan-Mini, our concurrent work, is an exceptional SLM trained on 1.08T tokens sourced entirely from open data. Yulan-Mini stands out for providing complete transparency regarding its development details and training corpus.
\end{itemize}

As shown in~\autoref{tab:main}, \thellm{} performs competitively, securing the third-highest average score (57.29), just behind Qwen2.5-1.5B (59.25) and Yulan-Mini-2.4B (57.51). Notably, \thellm{}-1.8B achieves the highest score in HumanEval among all models and ranks second in ARC-C, ARC-E, MBPP, and BoolQ, slightly trailing the industry-leading Qwen2.5-1.5B. This highlights its strong general knowledge comprehension and coding capabilities.

Compared to Qwen2.5-1.5B, \thellm{}-1.8B falls slightly behind in general knowledge benchmarks such as MMLU, CMMLU, and C-Eval, though its strong performance in ARC-E demonstrates its competency in reasoning tasks. However, it is worth noting that the Qwen series leverages an 18T closed-source corpus. 
Against Yulan-Mini-2.4B, \thellm{}-1.8B performs slightly worse in coding and logical reasoning tasks but remains a competitive option overall. Furthermore, with only 1.8B parameters and fewer layers (32 (\thellm{}) vs 56 (Yulan-Mini)), \thellm{} would offer significantly lower inference latency.

In conclusion, \thellm{}-1.8B is a strong and reliable model, particularly in basic knowledge understanding, coding and simple reasoning tasks.~\autoref{cases} shows the cases generated by \thellm{}.

% \usepackage{booktabs}
% \usepackage{multirow}
\begin{table}[h]
\caption{Comparison between and other representative models.}
\label{tab:main}
\resizebox{1\textwidth}{!}{
\begin{tabular}{@{}cccccccccccc@{}}
\hline
\toprule
\multirow{2}{*}{Benchmark} & MiniCPM & Yulan-Mini & Index & SmolLM2 & Fox & StableLM2 & Qwen2.5 & Qwen2 & GLM-Edge & \thellm{}$_{Base}$ & \thellm{}$_{Inst}$        \\ 
         & 2.7B    & 2.4B       & 1.9B  & 1.7B    & 1.6B  & 1.6B      & 1.5B    & 1.5B  & 1.5B     & 1.8B & 1.8B. \\ \midrule
ARC-C    & 43.86   & 50.51      & 42.23 & 50.29   & 41.21 & 40.78     & \textbf{53.41}   & 43.90 & 24.15    & 48.12     &  \underline{51.14}      \\
ARC-E    & 55.51   & 69.87      & 67.05 & 77.78   & 65.00 & 54.00     & \textbf{79.13}   & 62.21 & 36.83    & 77.31     & \underline{78.18}      \\
MMLU     & 51.13   & 49.10      & 50.98 & 51.91   & 43.05 & 40.37     & \textbf{59.79}   & \underline{56.50} & 54.84    & 48.02     & 51.18      \\
CMMLU    & 48.97   & 48.35      & 53.79 & 33.46   & 28.28 & 29.29     & \underline{67.82}   & \textbf{70.30} & 54.23    & 45.40     & 48.18      \\
C-Eval   & 48.24   & 51.47      & 54.82 & 35.10   & 28.75 & 26.99     & \underline{69.05}   & \textbf{70.60} & 55.05    & 44.50     & 44.93      \\ \midrule
GSM8K    & 53.83   & \underline{66.65}      & 28.66 & 47.68   & 36.39 & 20.62     & \textbf{68.50}   & 46.90 & 54.89    & 41.70     & 60.73      \\ 
MathQA   & 30.59   & \underline{34.84}      & 31.79 & 34.30   & 27.30 & 21.56     & \textbf{35.14}   & 31.66 & 33.94    & 28.81     & 33.23      \\ \midrule
HumanEval& 50.00   & \underline{61.60}      & 1.20  & 23.35   & 24.57 & 8.50      & 37.20   & 34.80 & 1.21     & 43.30     & \textbf{64.60}      \\
MBPP     & 47.31   & \textbf{66.70}      & 33.07 & 45.00   & 16.54 & 17.50     & 60.20   & 46.90 & 3.44     & 55.30     & \underline{60.40}      \\ \midrule
BoolQ    & 73.55   & 70.89      & \textbf{83.24} & 72.26   & 71.93 & 69.08     & 72.91   & 72.69 & 60.95    & 73.67     & \underline{77.86}      \\
Hellaswag& 53.06   & \underline{71.47}      & 70.91 & \textbf{71.48}   & 62.32 & 67.79     & 67.73   & 65.41 & 29.39    & 66.27     & 68.17      \\
LogiQA   & \textbf{31.64}   & 29.65      & 30.41 & 29.65   & 28.57 & 27.19     & \underline{31.03}   & 31.02 & 22.73    & 29.19     & 30.12      \\
PIQA     & 77.04   & 76.50      & 76.11 & 77.04   & 75.35 & 74.10     & \textbf{76.01}   & \underline{75.35} & 74.32    & 75.35     & 76.01      \\ \midrule
Average  & 51.13   & \textbf{57.51 (2$^{nd}$)} & 48.02 & 49.95   & 42.25 & 38.29     & \textbf{59.84 (1$^{st}$)}  & 54.48 & 38.92    & 52.07     & \textbf{57.29 (3$^{rd}$)} \\ \bottomrule \hline
\end{tabular}
}
\end{table}

\section{Evaluation: In the Context of Modeling-Hardware Co-Design}

In this section, we introduce the baseline models from the perspectives of modeling-hardware co-design and architectural comparisons during inference. This analysis provides insights into the trade-offs between efficiency and performance across different model designs.

\begin{itemize}[leftmargin=1.2em]
    \item \textbf{MiniCPM Series}~\cite{hu2024minicpm}, we choose 2.7B model and 4B model developed by ModelBest, models designed for edge-devices. Moreover, the 4B parameter model also incorporates the MLA mechanism.
    \item \textbf{GLM-Edge (1.5B and 4B)~\cite{glmedge} :} Released by GLM team, GLM-Edge-1.5B is designed to address the challenges of deploying real-world applications on edge devices. It offers several models in different sizes tailored for natural language multimodal understanding. However, the training data and details are not open-sourced.
    \item \textbf{Nemotron}~\cite{muralidharan2024compact}, a 4B parameter model released by NVIDIA, employing the squared ReLU (ReLU$^2$) activation function to enhance expressivity and computational efficiency.
    \item \textbf{Fox}~\cite{hu2024fox}, a 1.6B parameter model developed by TensorOpera, which leverages Grouped-Query Attention (GQA) and the SwiGLU activation function. Given that Fox is a 32-layer dense model, we consider it a direct baseline for efficiency comparisons.
    \item \textbf{Index}~\cite{Indexbilibili}, a 1.9B parameter model developed by Bilibili, utilizing Multi-Query Attention (MQA) alongside the SwiGLU activation function.
\end{itemize}

These models employ diverse attention mechanisms and activation functions, each optimized for different computational constraints and inference efficiency requirements. In our evaluation of model efficiency, on one hand, Nemotron-4B and GLM-Edge-4B are treated as a comparative group. A comparison between MiniCPM-4B and GLM-Edge-4B illustrates the inference speed impact of MLA, while a comparison involving Nemotron-4B and GLM-Edge-4B demonstrates the speed benefits of sparsity. In a separate comparison, PLM-1.8B, Fox-1.6B, Index-1.9B, and GLM-Edge-1.5B are analyzed as a group. A direct speed comparison between Fox-1.6B and PLM-1.8B highlights the differences between our model design and prevalent architectures at equivalent layer counts. The impact of layer count on inference speed is demonstrated by comparing Fox-1.6B and GLM-Edge-1.5B. Lastly, the difference in model size between Index-1.9B and PLM-1.8B highlights the impact of size on small language model performance.

By analyzing their performance characteristics and design rationales, we aim to systematically assess the impact of modeling and system co-design on efficiency and scalability.

% \usepackage{booktabs}
\begin{table}[h]
\caption{Baseline models compared in edge-side practice evaluation.}
\label{tab:hardware-comp-baselines}
\centering
\resizebox{0.55\textwidth}{!}{%
\begin{tabular}{@{}c|ccc@{}}
\toprule
\textbf{Model}    & \textbf{Attention} & \textbf{Activation} & \textbf{Parameters} \\ \midrule
\textbf{MiniCPM3-4B} & MLA                & SwiGLU              &   4.21B  \\
\textbf{GLM-Edge-4B} & GQA             & SwiGLU              &   4.21B  \\
\textbf{Nemotron-4B} & GQA                & ReLU$^2$            &   4.13B  \\ \midrule
\textbf{MiniCPM2}  & MLA                 & SwiGLU              &  2.74B  \\
\textbf{GLM-Edge-1.5B} & GQA           & SwiGLU              &   1.59B  \\
\textbf{Fox-1.6B}      & GQA                & SwiGLU              &   1.67B  \\
\textbf{Index-1.9B}    & MQA                & SwiGLU              &   1.91B  \\
\textbf{\thellm{}-1.8B}& MLA                & ReLU$^2$            &   1.83B  \\ \bottomrule
\end{tabular}
}
\end{table}

% \usepackage{booktabs}
% \begin{table}[h]
% \caption{Efficiency Comparison among Model Architectures} 
% \label{tab:efficient-arch}
% \resizebox{\textwidth}{!}{%
% \begin{tabular}{@{}ccc|cccccccc@{}}
% \toprule
% & $n_{layer}$ & $d_{model}/d_{ffn}$ & \textbf{Attention} & \textbf{Activation} & \textbf{MACs} & \textbf{FLOPs} & \textbf{MACs/Param} & \textbf{FLOPs/Param} & \textbf{Token Rate} & \textbf{Peak Memory} \\ \midrule
% \textbf{\thellm{}-1.8B} & $32$ & $2048/8192$ & MLA & ReLU$^2$ & 202.94 & 405.92 & 107.9468 & 225.5111 & 8.2566 & 6.687e-5 \\
% \textbf{Qwen2.5-1.5B}& $28$ & $1536/8960$ & GQA & SwiGLU  & 167.71 & 335.44 & 111.8067 & 223.6267 & 9.2378 & 1.57e-4   \\
% \textbf{MiniCPM3-4B} & $62$ & $2560/6400$ & MLA & SwiGLU & 497.34 & 994.73 & 124.335 & 248.6825 & 2.0143 & 2.8e-4  \\
% \textbf{Nemotron-4B} & $32$ & $3072/9216$ & GQA & ReLU$^2$ &        &          &               &                &               &            \\ \bottomrule
% \end{tabular}
% }
% \end{table}

Furthermore, we evaluate the performance of deploying models on different kinds of hardware, including consumer level GPU (NVIDIA A10), edge-side GPU (Jetson), CPU (Intel CPU), Hybird process unit (Apple M3 and Qualcomm Snapdragon 8 Gen 3) and entry-level ARM SoC process unit (Broadcom BCM2712). Particularly, we evaluate the selected architectures with the most representative models, shown in~\autoref{tab:hardware-comp-baselines}.

\begin{itemize}[leftmargin=1.2em]
    % \item \textbf{NVIDIA H100}, built on the Hopper architecture, offers 80 GB of HBM3 memory, a 700W TDP, and advanced capabilities like the Transformer Engine and FP8 precision for large-scale AI, generative models, and HPC tasks. We use single H100 chip to present extraordinary edge-side setting.
    \item \textbf{NVIDIA A10}, based on the Ampere architecture, features 24 GB of GDDR6 memory, a 150W TDP, and is optimized for AI inferencing, training, VDI, and visualization in enterprise workloads. We use single A10 chip to present general edge-side GPU setting.
    \item \textbf{Jetson Orin NX 16GB}, based on NVIDIA's Ampere architecture, features an 8-core Cortex-A78AE CPU, a 1024-core Ampere GPU with 32 Tensor Cores, 16 GB of LPDDR5 memory (102.4 GB/s), and delivers up to 100 TOPS for edge AI and robotics applications. We use single A10 chip to present edge-side high-performance computing unit setting.
    % \item \textbf{Intel Core i7-1360P}, a 13th-gen processor, features 12 cores (4P + 8E), 16 threads, up to 5.0 GHz turbo, and Intel Iris Xe graphics, offering power-efficient performance for laptops. We select the laptop with this CPU to perform the experiments.
    \item \textbf{Apple M3}, a 3nm SoC, features an 8-core CPU (4P + 4E), up to 10-core GPU with ray tracing, a 16-core Neural Engine (18 TOPS), and supports up to 24 GB LPDDR5 memory (100 GB/s), powering MacBooks and iMacs. We use MacBook Air M3 in the experiments.
    \item \textbf{Qualcomm Snapdragon 8 Gen 3}, built on a 4nm process, features an 8-core CPU (1 Cortex-X4, 3 Cortex-A720 @ 3.15 GHz, 2 Cortex-A720 @ 2.96 GHz, 2 Cortex-A520), an Adreno 750 GPU, LPDDR5X support (24 GB, 76.8 GB/s), and advanced AI and imaging capabilities. We use the One Plus 12 as the experiment environment.
    \item \textbf{Broadcom BCM2712} is an entry-level ARM SoC featuring a quad-core Cortex-A76 CPU, VideoCore V3D VII GPU (12 cores, 800 MHz), LPDDR4X memory (32-bit, 17 GB/s), and H.264 1080p60 decode/1080p30 encode. Used in the Raspberry Pi 5, it is built on a 16nm process.
\end{itemize}

\autoref{tab:syslatency} presents a comprehensive evaluation of inference latency for various edge-size LLMs across different hardware platforms and quantization levels. The models assessed include GLM-Edge-4B, Nemotron, MiniCPM3, MiniCPM2, GLM-Edge-15B, \thellm{} (\textbf{bold}), Fox, and Index, with activation functions such as SwiGLU and ReLU² under varying attention mechanisms. Latency measurements are provided for both prefill (512 tokens) and generation (128 tokens) scenarios, comparing full-precision (FP16) and lower-bit quantizations (Q8, Q4). The results span multiple computing platforms, including high-performance GPUs (NVIDIA A10, Orin NX), Apple’s M3 chip, Qualcomm’s Snapdragon 8 Gen 3, and embedded systems like BCM2712.  

The data reveal significant trade-offs between precision and efficiency, with quantized models (Q8, Q4) generally offering lower memory cost compared to their FP16 counterparts, particularly suitable for edge devices. However, the extent of latency reduction varies by model architecture and hardware, with Snapdragon 8 Gen 3 and M3 achieving competitive performance in specific settings. This analysis highlights the importance of selecting appropriate quantization strategies and hardware acceleration for balancing model performance and computational efficiency in real-world deployment scenarios.

\begin{figure}[h]
    \centering
    \includegraphics[width=\linewidth]{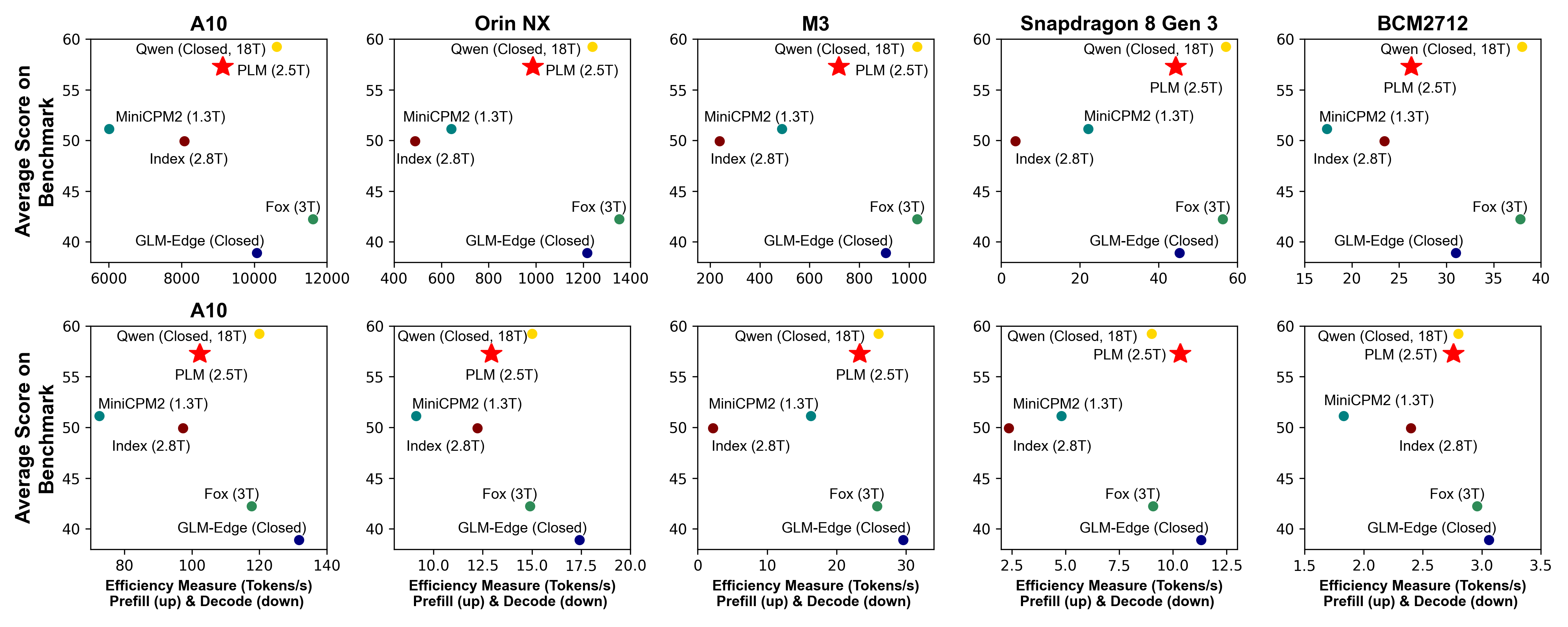}
    \caption{Efficiency comparison of LLMs across different hardware configurations and quantization levels. Optimal hardware-quantization combinations are highlighted, illustrating that PLM achieves an effective balance between robust performance and high inference speed.}
    \label{fig:latencyall}
\end{figure}

In general, \thellm{} demonstrates strong optimization for both FP16 and quantized configurations (8-bit and 4-bit), positioning it as a highly efficient model for inference across diverse hardware platforms. Notably, \thellm{} surpasses Fox on the Snapdragon 8 Gen 3 in both the FP16 generation case and the 8-bit quantization prefill case (\underline{underline}), despite having the same number of layers and leveraging a widely deployed architecture. This performance advantage suggests that \thellm{} is particularly well-suited for latency-sensitive applications, offering significant benefits for real-time inference and on-device LLM deployments where computational efficiency is critical.Moreover,~\autoref{fig:latencyall} visualize the performance and effieciency trade-off of the models.

% \usepackage{booktabs}
% \usepackage{multirow}
\begin{table}[]
\caption{Inference latency of LLMs on various hardware and quantization Levels.}
\label{tab:syslatency}
\resizebox{\linewidth}{!}{
\begin{tabular}{@{}cccccccc@{}}
\toprule
& & & A10 & NVIDIA Orin NX  & M3 & Snapdragon 8 Gen 3 & BCM2712  \\ 
& & & \begin{minipage}[b]{0.15\columnwidth}
		\centering
		\raisebox{-.5\height}{\includegraphics[width=\linewidth]{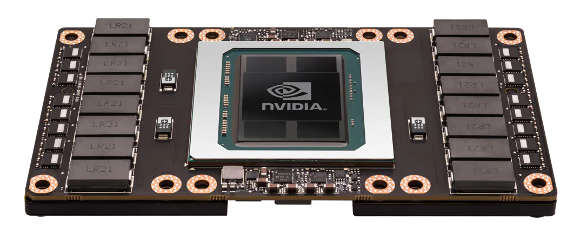}}\end{minipage} &
      \begin{minipage}[b]{0.15\columnwidth}
		\centering
		\raisebox{-.5\height}{\includegraphics[width=\linewidth]{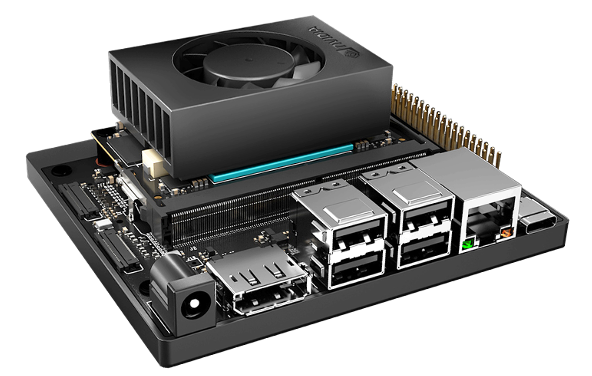}}\end{minipage} &
      \begin{minipage}[b]{0.15\columnwidth}
		\centering
		\raisebox{-.5\height}{\includegraphics[width=0.8\linewidth]{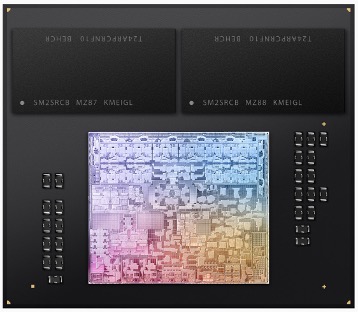}}\end{minipage} &
      \begin{minipage}[b]{0.15\columnwidth}
		\centering
		\raisebox{-.5\height}{\includegraphics[width=0.6\linewidth]{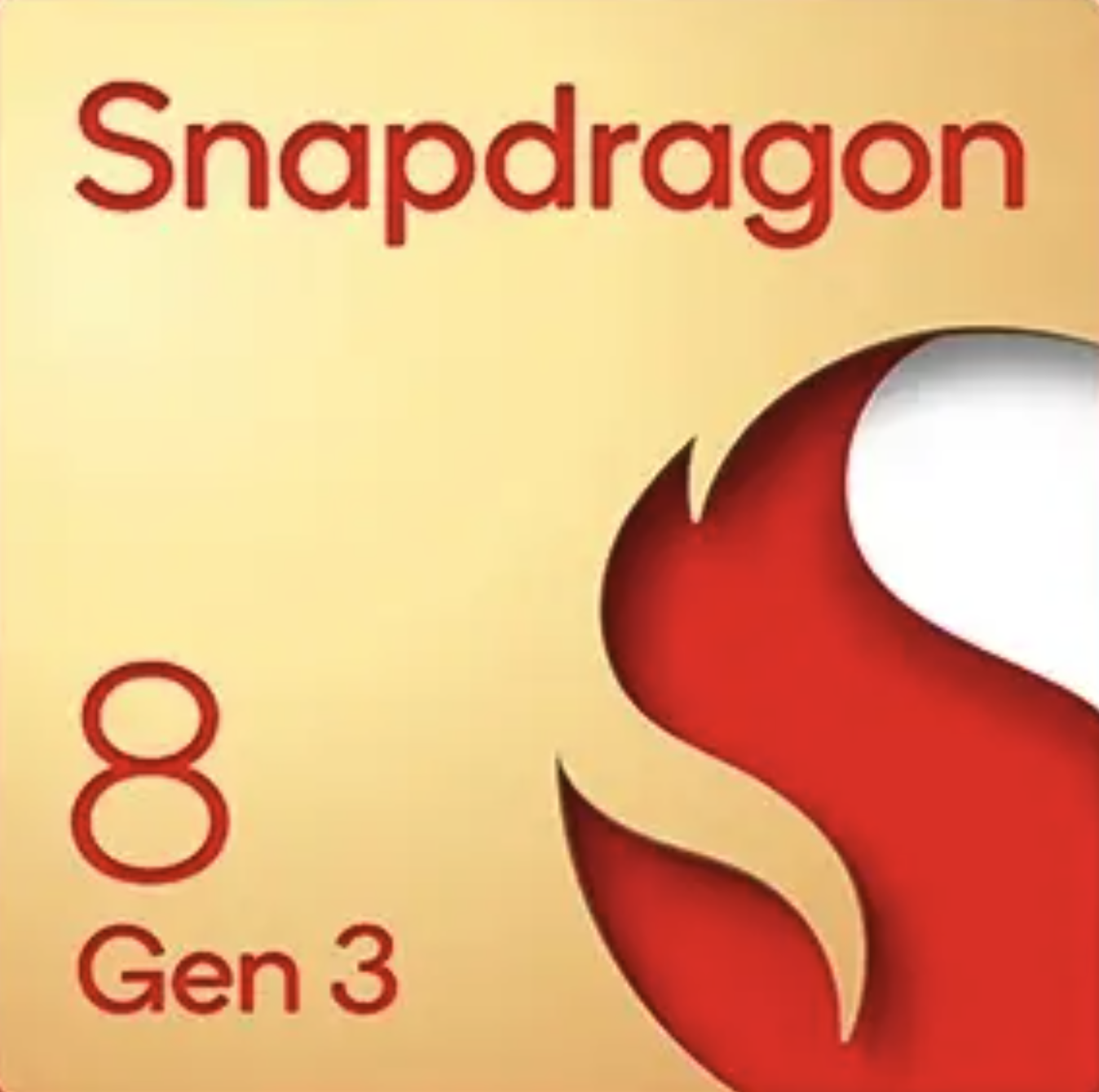}}\end{minipage} &
      \begin{minipage}[b]{0.15\columnwidth}
		\centering
		\raisebox{-.5\height}{\includegraphics[width=0.9\linewidth]{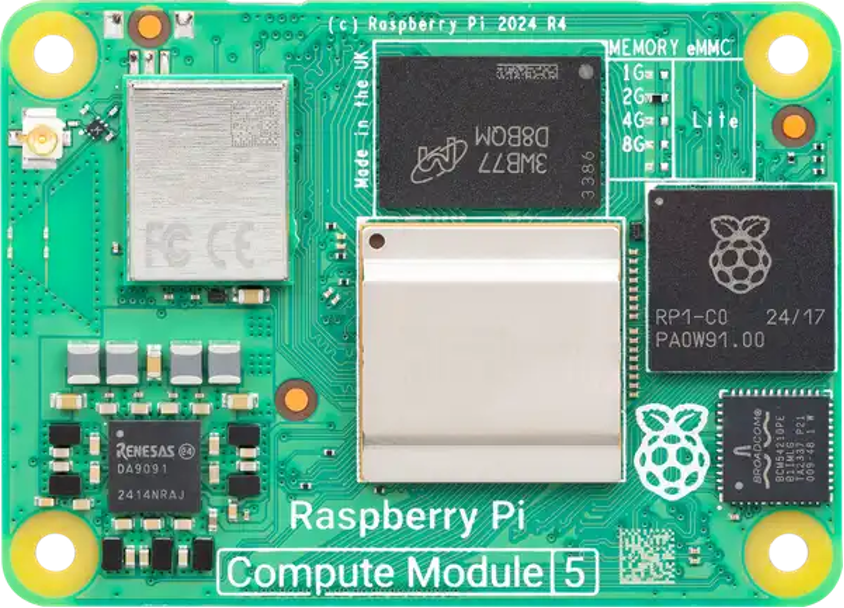}}\end{minipage} 
      \\ \midrule
\multicolumn{1}{c|}{\multirow{16}{*}{FP16}} & \multicolumn{1}{c|}{\multirow{8}{*}{Prefill 512}}  & \multicolumn{1}{c|}{GLM-Edge-4B}   & 4335.31 ± 15.52   & 421.43 ± 15.03  & 292.84 ± 12.02 & 12.15 ± 0.15       & 10.12 ± 0.13 \\
\multicolumn{1}{c|}{}                       & \multicolumn{1}{c|}{}                              & \multicolumn{1}{c|}{Nemotron-4B}      & 6543.59 ± 14.38   & 605.89 ± 12.51  & 365.06 ± 81.42 & 20.06 ± 2.29       & 16.82 ± 0.09 \\
\multicolumn{1}{c|}{}                       & \multicolumn{1}{c|}{}                              & \multicolumn{1}{c|}{MiniCPM3-4B}      & 3139.28 ± 73.67   & 331.36 ± 12.91  & 129.27 ± 21.09 & 12.87 ± 1.67       & 4.73 ± 0.45  \\
\multicolumn{1}{c|}{}                       & \multicolumn{1}{c|}{}                              & \multicolumn{1}{c|}{MiniCPM-2.7B}      & 6003.20 ± 33.99   & 641.36 ±10.80   & 488.45 ± 0.61  & 22.09 ± 3.47       & 17.36 ± 0.08 \\
\multicolumn{1}{c|}{}                       & \multicolumn{1}{c|}{}                              & \multicolumn{1}{c|}{GLM-Edge-1.5B} & 10074.07 ± 53.99  & 1217.45 ± 9.74  & 905.82 ± 0.43  & 45.33 ± 5.78       & 27.01 ± 0.11 \\
\multicolumn{1}{c|}{}                       & \multicolumn{1}{c|}{}                              & \multicolumn{1}{c|}{\textbf{PLM-1.8B}}           & \textbf{9129.39 ± 82.80}   & \textbf{987.22 ± 19.13}  & \textbf{716.94 ± 13.62} & \textbf{44.35 ± 2.11}       & \textbf{26.29 ± 1.01} \\
\multicolumn{1}{c|}{}                       & \multicolumn{1}{c|}{}                              & \multicolumn{1}{c|}{Fox-1.6B}           & 11616.08 ± 134.66 & 1351.74 ± 22.89 & 1032.83 ± 1.50 & 56.29 ± 2.09       & 37.80 ± 0.09 \\
\multicolumn{1}{c|}{}                       & \multicolumn{1}{c|}{}                              & \multicolumn{1}{c|}{Index-1.9B}         & 8074.48 ± 55.85   & 488.89 ± 20.53  & 237.80 ± 3.17  & 3.62 ± 0.38        & 23.44 ± 0.02 \\ \cmidrule(l){2-8} 
\multicolumn{1}{c|}{}                       & \multicolumn{1}{c|}{\multirow{8}{*}{Generate 128}} & \multicolumn{1}{c|}{GLM-Edge-4B}   & 52.25 ± 0.07      & 6.75 ± 0.04     & 10.44 ± 0.11   & 3.34 ± 0.05        & 0.74 ± 0.01  \\
\multicolumn{1}{c|}{}                       & \multicolumn{1}{c|}{}                              & \multicolumn{1}{c|}{Nemotron-4B}      & 63.96 ± 0.08      & 8.19 ± 0.05     & 9.01 ± 0.53    & 4.02 ± 0.08        & 1.49 ± 0.00  \\
\multicolumn{1}{c|}{}                       & \multicolumn{1}{c|}{}                              & \multicolumn{1}{c|}{MiniCPM3-4B}      & 44.17 ± 0.11      & 5.79 ± 0.01     & 6.37 ± 1.14    & 3.16 ± 0.22        & 0.02 ± 0.00  \\
\multicolumn{1}{c|}{}                       & \multicolumn{1}{c|}{}                              & \multicolumn{1}{c|}{MiniCPM-2.7B}      & 72.56 ± 0.17      & 9.11 ± 0.07     & 16.31 ± 0.11   & 4.81 ± 0.19        & 1.83 ± 0.04  \\
\multicolumn{1}{c|}{}                       & \multicolumn{1}{c|}{}                              & \multicolumn{1}{c|}{GLM-Edge-1.5B} & 131.71 ± 0.26     & 17.41 ± 0.02    & 29.57 ± 0.08   & 11.31 ± 0.62       & 3.12 ± 0.02  \\
\multicolumn{1}{c|}{}                       & \multicolumn{1}{c|}{}                              & \multicolumn{1}{c|}{\textbf{PLM-1.8B}}           & \textbf{102.30 ± 0.25}     & \textbf{12.94 ± 0.10}    & \textbf{23.31 ± 0.05}   & \textbf{\underline{10.34 ± 0.47}}       & \textbf{2.76 ± 0.00}  \\
\multicolumn{1}{c|}{}                       & \multicolumn{1}{c|}{}                              & \multicolumn{1}{c|}{Fox-1.6B}           & 117.69 ± 0.26     & 14.90 ± 0.07    & 25.81 ± 0.33   & 9.07 ± 1.23        & 2.96 ± 0.08  \\
\multicolumn{1}{c|}{}                       & \multicolumn{1}{c|}{}                              & \multicolumn{1}{c|}{Index-1.9B}         & 97.36 ± 0.21      & 12.23 ± 0.06    & 2.22 ± 0.11    & 2.35 ± 0.09        & 2.40 ± 0.00  \\ \midrule
\multicolumn{1}{c|}{\multirow{16}{*}{Q8}}   & \multicolumn{1}{c|}{\multirow{8}{*}{Prefill 512}}  & \multicolumn{1}{c|}{GLM-Edge-4B}   & 3770.60 ± 7.56    & 458.50 ± 10.29  & 281.71 ± 8.48  & 18.77 ± 0.13       & 11.37 ± 0.01 \\
\multicolumn{1}{c|}{}                       & \multicolumn{1}{c|}{}                              & \multicolumn{1}{c|}{Nemotron-4B}      & 6430.92 ± 29.40   & 728.69 ± 2.00   & 403.62 ± 15.96 & 28.25 ± 0.48       & 17.20 ± 0.11 \\
\multicolumn{1}{c|}{}                       & \multicolumn{1}{c|}{}                              & \multicolumn{1}{c|}{MiniCPM3-4B}      & 3107.81 ± 62.28   & 349.60 ±4.89    & 268.59 ± 0.94  & 17.51 ± 0.16       & 10.89 ± 0.01 \\
\multicolumn{1}{c|}{}                       & \multicolumn{1}{c|}{}                              & \multicolumn{1}{c|}{MiniCPM-2.7B}      & 5279.70 ± 12.02   & 654.74 ±4.96    & 463.40 ± 0.52  & 29.40 ± 0.99       & 17.96 ± 0.09 \\
\multicolumn{1}{c|}{}                       & \multicolumn{1}{c|}{}                              & \multicolumn{1}{c|}{GLM-Edge-1.5B} & 9030.11 ± 73.89   & 1156.42 ± 1.56  & 861.27 ± 0.58  & 68.36 ± 1.87       & 30.82 ± 0.03 \\
\multicolumn{1}{c|}{}                       & \multicolumn{1}{c|}{}                              & \multicolumn{1}{c|}{\textbf{PLM-1.8B}}           & \textbf{8833.31 ± 243.13}  & \textbf{989.01 ±3.96}    & \textbf{707.76 ± 0.38}  & \textbf{\underline{64.18 ± 2.51}}       & \textbf{29.04 ± 0.21} \\
\multicolumn{1}{c|}{}                       & \multicolumn{1}{c|}{}                              & \multicolumn{1}{c|}{Fox-1.6B}           & 11250.68 ± 109.62 & 1385.81 ±17.66  & 828.30 ± 20.52 & 58.33 ± 4.20       & 35.34 ± 4.22 \\
\multicolumn{1}{c|}{}                       & \multicolumn{1}{c|}{}                              & \multicolumn{1}{c|}{Index-1.9B}         & 7869.27 ± 37.52   & 956.54 ± 1.27   & 581.14 ± 5.25  & 43.90 ± 0.26       & 19.93 ± 0.13 \\ \cmidrule(l){2-8} 
\multicolumn{1}{c|}{}                       & \multicolumn{1}{c|}{\multirow{8}{*}{Generate 128}} & \multicolumn{1}{c|}{GLM-Edge-4B}   & 83.11 ± 0.16      & 11.79 ± 0.09    & 19.45 ± 0.03   & 5.85 ± 0.57        & 1.18 ± 0.09  \\
\multicolumn{1}{c|}{}                       & \multicolumn{1}{c|}{}                              & \multicolumn{1}{c|}{Nemotron-4B}      & 110.29 ± 0.25     & 14.35 ± 0.16    & 22.93 ± 0.34   & 7.79 ± 1.24        & 2.97 ± 0.00  \\
\multicolumn{1}{c|}{}                       & \multicolumn{1}{c|}{}                              & \multicolumn{1}{c|}{MiniCPM3-4B}      & 67.19 ± 0.27      & 8.61 ± 0.03     & 17.76 ± 0.10   & 5.41 ± 0.48        & 2.38 ± 0.00  \\
\multicolumn{1}{c|}{}                       & \multicolumn{1}{c|}{}                              & \multicolumn{1}{c|}{MiniCPM-2.7B}      & 108.06 ± 0.28     & 14.09 ± 0.05    & 27.05 ± 0.21   & 9.06 ± 1.53        & 3.47 ± 0.01  \\
\multicolumn{1}{c|}{}                       & \multicolumn{1}{c|}{}                              & \multicolumn{1}{c|}{GLM-Edge-1.5B}     & 191.80 ± 0.40     & 27.43 ± 0.25    & 50.40 ± 0.06   & 17.08 ± 0.35       & 6.19 ± 0.12  \\
\multicolumn{1}{c|}{}                       & \multicolumn{1}{c|}{}                              & \multicolumn{1}{c|}{\textbf{PLM-1.8B}}           & \textbf{157.35 ± 0.50}     & \textbf{20.94 ± 0.19}    & \textbf{38.89 ± 0.83}   & \textbf{13.78 ± 0.75}       & \textbf{5.31 ± 0.01}  \\
\multicolumn{1}{c|}{}                       & \multicolumn{1}{c|}{}                              & \multicolumn{1}{c|}{Fox-1.6B}           & 180.39 ± 0.73     & 23.50 t 0.31    & 62.00 ± 3.17   & 15.80 ± 1.67       & 6.00 ± 0.00  \\
\multicolumn{1}{c|}{}                       & \multicolumn{1}{c|}{}                              & \multicolumn{1}{c|}{Index-1.9B}         & 152.79 ± 0.38     & 20.23 ± 0.03    & 34.46 ± 0.71   & 13.30 ± 0.55       & 3.90 ± 0.01  \\ \midrule
\multicolumn{1}{c|}{\multirow{16}{*}{Q4}}   & \multicolumn{1}{c|}{\multirow{8}{*}{Prefill 512}}  & \multicolumn{1}{c|}{GLM-Edge-4B}   & 3748.14 ± 11.02   & 435.25 ± 0.52   & 293.26 ± 1.39  & 16.33 ± 0.06       & 9.187 ± 0.13 \\
\multicolumn{1}{c|}{}                       & \multicolumn{1}{c|}{}                              & \multicolumn{1}{c|}{Nemotron-4B}      & 6156.83 ± 5.16    & 671.07 ± 6.06   & 376.61 ± 4.36  & 26.98 ± 0.65       & 14.03 ± 0.02 \\
\multicolumn{1}{c|}{}                       & \multicolumn{1}{c|}{}                              & \multicolumn{1}{c|}{MiniCPM3-4B}      & 2985.06 ± 49.51   & 332.56±1.63     & 252.20 ± 4.95  & 17.08 ± 1.01       & 9.21 ± 0.01  \\
\multicolumn{1}{c|}{}                       & \multicolumn{1}{c|}{}                              & \multicolumn{1}{c|}{MiniCPM-2.7B}      & 5291.98 ± 10.21   & 614.94 ± 3.27   & 445.00 ± 1.16  & 24.56 ± 0.18       & 14.88 ± 0.13 \\
\multicolumn{1}{c|}{}                       & \multicolumn{1}{c|}{}                              & \multicolumn{1}{c|}{GLM-Edge-1.5B} & 8929.26 ± 102.41  & 1062.21 ± 0.48  & 826.42 ± 0.57  & 61.93 ± 1.83       & 23.61 ± 0.02 \\
\multicolumn{1}{c|}{}                       & \multicolumn{1}{c|}{}                              & \multicolumn{1}{c|}{\textbf{PLM-1.8B}}           & \textbf{8266.53 ± 209.05}  & \textbf{910.61 ± 6.02}   & \textbf{680.78 ± 0.33}  & \textbf{48.05 ± 7.07}       & \textbf{23.57 ± 0.18} \\
\multicolumn{1}{c|}{}                       & \multicolumn{1}{c|}{}                              & \multicolumn{1}{c|}{Fox-1.6B}           & 10483.37 ± 69.05  & 1284.18 ±16.86  & 789.35 ± 6.44  & 58.19 ± 0.27       & 31.42 ± 0.32 \\
\multicolumn{1}{c|}{}                       & \multicolumn{1}{c|}{}                              & \multicolumn{1}{c|}{Index-1.9B}         & 7443.99 ± 46.19   & 858.28 ± 4.56   & 532.56 ± 81.38 & 40.06 ± 1.41       & 19.61 ± 0.13 \\ \cmidrule(l){2-8} 
\multicolumn{1}{c|}{}                       & \multicolumn{1}{c|}{\multirow{8}{*}{Generate 128}} & \multicolumn{1}{c|}{GLM-Edge-4B}   & 118.08 ± 0.25     & 16.63 ± 0.05    & 31.12 ± 0.08   & 7.40 ± 1.07        & 1.98 ± 0.01  \\
\multicolumn{1}{c|}{}                       & \multicolumn{1}{c|}{}                              & \multicolumn{1}{c|}{Nemotron-4B}      & 145.60 ± 0.55     & 18.34 ± 0.15    & 34.61 ± 1.04   & 10.54 ± 0.13       & 4.54 ± 0.01  \\
\multicolumn{1}{c|}{}                       & \multicolumn{1}{c|}{}                              & \multicolumn{1}{c|}{MiniCPM3-4B}      & 85.33 ± 0.33      & 12.00 ± 0.07    & 26.67 ± 0.15   & 5.46 ± 2.58        & 3.92 ± 0.00  \\
\multicolumn{1}{c|}{}                       & \multicolumn{1}{c|}{}                              & \multicolumn{1}{c|}{MiniCPM-2.7B}      & 145.41 ± 0.37     & 19.37 ± 0.13    & 40.84 ± 0.06   & 11.38 ± 0.16       & 5.61 ± 0.02  \\
\multicolumn{1}{c|}{}                       & \multicolumn{1}{c|}{}                              & \multicolumn{1}{c|}{GLM-Edge-1.5B} & 255.52 ± 0.94     & 37.88 ± 0.36    & 77.98 ± 0.11   & 24.96 ± 2.08       & 4.54 ± 0.01  \\
\multicolumn{1}{c|}{}                       & \multicolumn{1}{c|}{}                              & \multicolumn{1}{c|}{\textbf{PLM-1.8B}}           & \textbf{197.48 ± 1.08}     & \textbf{27.04 ± 0.12}    & \textbf{59.44 ± 0.15}   & \textbf{18.74 ± 1.26}       & \textbf{8.02 ± 0.03}  \\
\multicolumn{1}{c|}{}                       & \multicolumn{1}{c|}{}                              & \multicolumn{1}{c|}{Fox-1.6B}           & 226.85 ± 1.00     & 30.20 ± 0.09    & 63.66 ± 1.09   & 17.67 ± 0.93       & 8.77 ± 0.02  \\
\multicolumn{1}{c|}{}                       & \multicolumn{1}{c|}{}                              & \multicolumn{1}{c|}{Index-1.9B}         & 201.02 ± 0.75     & 26.93 ± 0.17    & 56.05 ± 0.74   & 14.66 ± 0.62       & 7.65 ± 0.03  \\ \bottomrule
\end{tabular}
}
\end{table}

% \newpage
\section{Analysis: Insights into the Design of \thellm{}}
% Analysis: Understanding \thellm{} Design
\label{obsall}

In this section, we will delve into a detailed observation of \thellm{}, aiming to uncover its capabilities, potential, and optimal integration into diverse environments. By addressing two pivotal questions—how to select suitable peripheral language models and the untapped potential of \thellm{} in ubiquitous computing—we provide insights into leveraging \thellm{} for innovative solutions and maximizing its utility across computing paradigms.

% \dc{Please add observation and comprehensive study of \thellm{} here.}

\subsection{Learned the Attention in \thellm{}} 
% \dc{From Prof wang, can we have some numerical result/calculation? related to the formulas}

Deploying models on edge devices involves challenges such as limited computational resources, memory capacity, and stringent latency requirements. While Multi-Head Latent Attention (MLA) incurs higher computational overhead than traditional multi-head attention (MHA) and Global Query Attention (GQA), its optimized cache utilization and inference efficiency make it particularly suited for edge deployment.

As shown in~\autoref{gqacache}, the number of KV heads directly impacts the size of the KV cache:
\begin{equation}
\label{gqacache}
    \text{Cache}_{\text{gqa}} = n_{\text{layers}} \times 2 \times n_{\text{kv\_heads}} \times d_{\text{h}} \times \left( \frac{\text{bit width}}{8} \right) \times N,
\end{equation}
where $n_{\text{layers}}$ is the number of layers, $n_{\text{kv\_heads}}$ is the number of KV heads, and $d_{\text{h}}$ is the dimensionality of each attention head. The factor of 2 accounts for storing both keys and values and $N$ is the number of token.

However, MLA reduces memory consumption by minimizing KV cache usage, as shown in~\autoref{mlacache}, making it more suitable for memory-constrained edge devices. While MLA incurs higher computational overhead in the prefill phase, this compute-intensive phase benefits from hardware optimizations commonly available on edge devices.

\begin{equation}
\label{mlacache}
    \text{Cache}_{\text{mla}} = n_{\text{layers}} \times (d_{\text{rope}} + d_{\text{c}}) \times \left( \frac{\text{bit width}}{8} \right) \times N ,
\end{equation}
where $d_{\text{rope}}$ and $d_{\text{c}}$ represent the dimensionalities of the rope key and compressed KV.
% , refer to~\autoref{analysis} for detailed calculation costs and cache. 
% \ly{Maybe we can calculate the cache difference and computation difference, and show when IO is more time-consuming}

% \ly{While MLA exhibits higher computational demands during the prefill phase, this compute-intensive stage can be effectively optimized through hardware acceleration features commonly available on modern edge processors. The differential decoding complexity between GQA and MLA demonstrates sequence-length dependence of $\mathcal{O}((0.5+2d_c)dN)$, with the operational difference expressed as $\mathcal{O}(2n_{\text{kv\_heads}}d_h - d_c - d_{\text{rope}})$. Notably, when I/O latency dominates computation time, MLA demonstrates superior performance characteristics compared to GQA. For comprehensive analysis of computational costs and cache requirements, see～\autoref{analysis}.} \ly{check}

While MLA exhibits increased computational demands during prefill, hardware acceleration available on modern edge processors effectively mitigates this cost. The differential decoding complexity between GQA and MLA is sequence-length dependent, scaling as $\mathcal{O}((0.5+2d_c)dN)$, with the cache difference quantified as $\mathcal{O}((2n_{\text{kv\_heads}}d_h - d_c - d_{\text{rope}})N)$. Critically, when I/O latency becomes the dominant factor, MLA outperforms GQA. For a detailed analysis of computational costs and cache requirements, refer to~\autoref{analysis}.

The primary advantage of MLA is evident during the decode phase, which is typically I/O-intensive. MLA reduces memory I/O transfer time by minimizing cache requirements, leading to faster generation despite a slight increase in computational cost. This reduction in I/O time is critical for edge devices, which demand low-latency inference.

In edge inference, quantization further enhances MLA's benefits. After quantization, the reduction in cache usage is linear, while the nonlinear nature of multiplication computations means that additional computational overhead is minimal, further improving memory I/O efficiency. This allows MLA to meet real-time inference demands on edge devices.

Furthermore, MLA minimizes memory consumption by sharing latent spaces across attention heads, reducing cache requirements per head. This is crucial for edge devices, where limited memory impacts system stability and responsiveness. By reducing memory usage, MLA improves performance, particularly in real-time or multi-task inference scenarios.

\begin{figure}[h]
  \centering
  \subfigure[Effect of token length on tokens per second (TPS) when model prefill.]{
      \label{latency_longctx1}
      \includegraphics[width=0.47\linewidth]{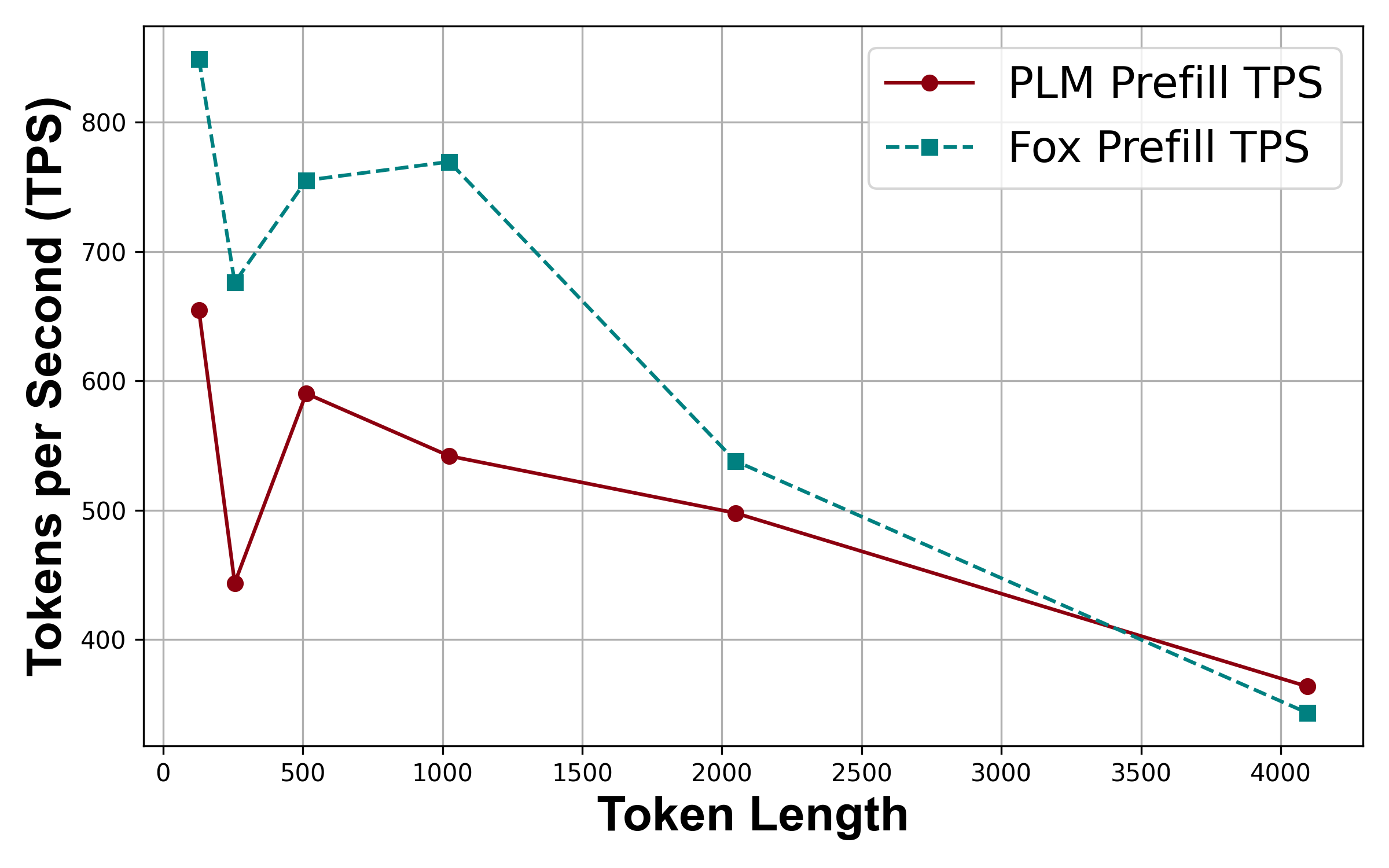}
  }
  \subfigure[Effect of token length on tokens per second (TPS) when model decode.]{
      \label{latency_longctx2}
      \includegraphics[width=0.47\linewidth]{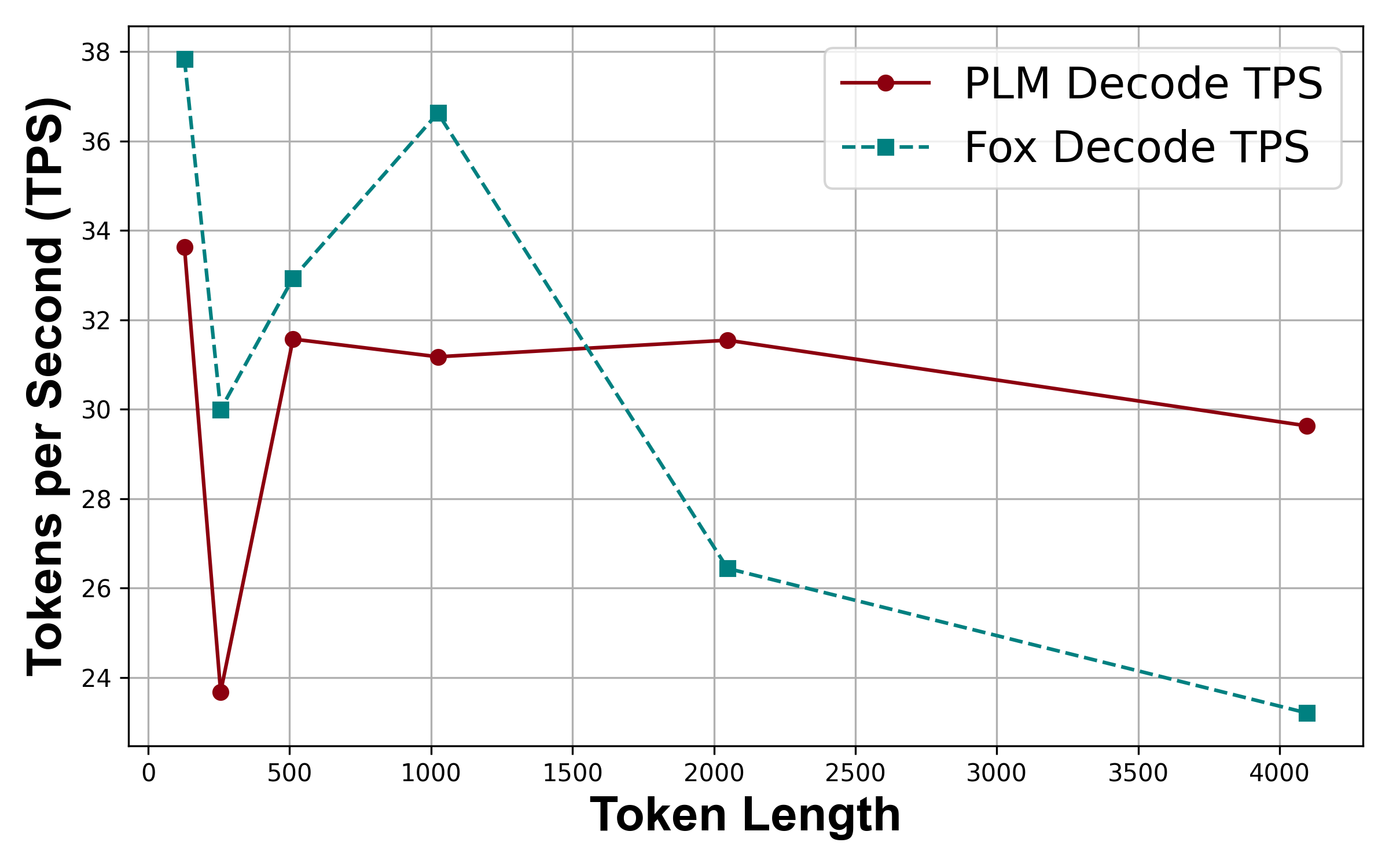}
  }
  \caption{Effect of token length (\thellm{} vs Fox).}
  \label{latency_longctx}
\end{figure}

We can anticipate that, in practical exploration, our model will exhibit advantages in certain specialized scenarios due to variations in I/O performance and computational power across different hardware devices. Based on previous experiments, it is evident that Snapdragon demonstrates superior adaptability, outperforming the GQA model Fox of the same depth to some extent, as shown in~\autoref{tab:syslatency}. Additionally, we evaluated our model's performance on slightly longer text sequences~\autoref{latency_longctx} and identified the advantages of MLA. Compared to DeepSeek’s memory-efficient approach on cloud servers, which allows for larger tensor parallelism (TP), we found that MLA maintains its advantages even in edge computing environments where TP is less critical.

% \usepackage{booktabs}
\begin{table}[h]
\centering
\caption{Latency comparison using 512-tokens prompt to generate 128 tokens and only load each layer iteratively on NVIDIA A10.}
\label{tab:iterload}
\begin{tabular}{@{}c|cc@{}}
\toprule
 Models   & Prefill(token/s)$\uparrow$ & Decode(token/s) $\uparrow$\\ \midrule
\thellm{}-Q8 & 8237.30  & \textbf{86.19}  \\
Fox-Q8 & 9673.17 & 67.82  \\
\thellm{}-Q4 & 7592.33 & \textbf{114.12} \\
Fox-Q4 & 8888.73 & 84.21  \\ \bottomrule
\end{tabular}
\end{table}

Thus, despite the increased computational load, MLA's optimization of cache utilization, inference efficiency, and memory consumption enables significant advantages for edge devices, offering efficient, low-latency performance in practical applications.

\subsection{Learned the Sparsity in \thellm{}}

Compared to MLA, the sparsification design of \thellm{} demonstrates more universally effective performance in edge-side models. From a system perspective, or more specifically, from the perspective of computational unit implementation within operating systems, computations involving zeros are highly optimized. Consequently, sparsification significantly benefits the deployment of models on edge devices.

As previously discussed, the introduction of MLA leads to a substantial increase in computational workload. However, by discarding low-rank compression of queries to reduce computational complexity during inference and incorporating sparse activation functions, our model avoids the efficiency drawbacks observed in MiniCPM3~\cite{hu2024minicpm}, whether in the prefill or decode stages.

\begin{figure}[h]
  \centering
  \subfigure[Effect of GPU offload layer number on tokens per second (TPS) when model prefill.]{
      \label{latency_iter1}
      \includegraphics[width=0.47\linewidth]{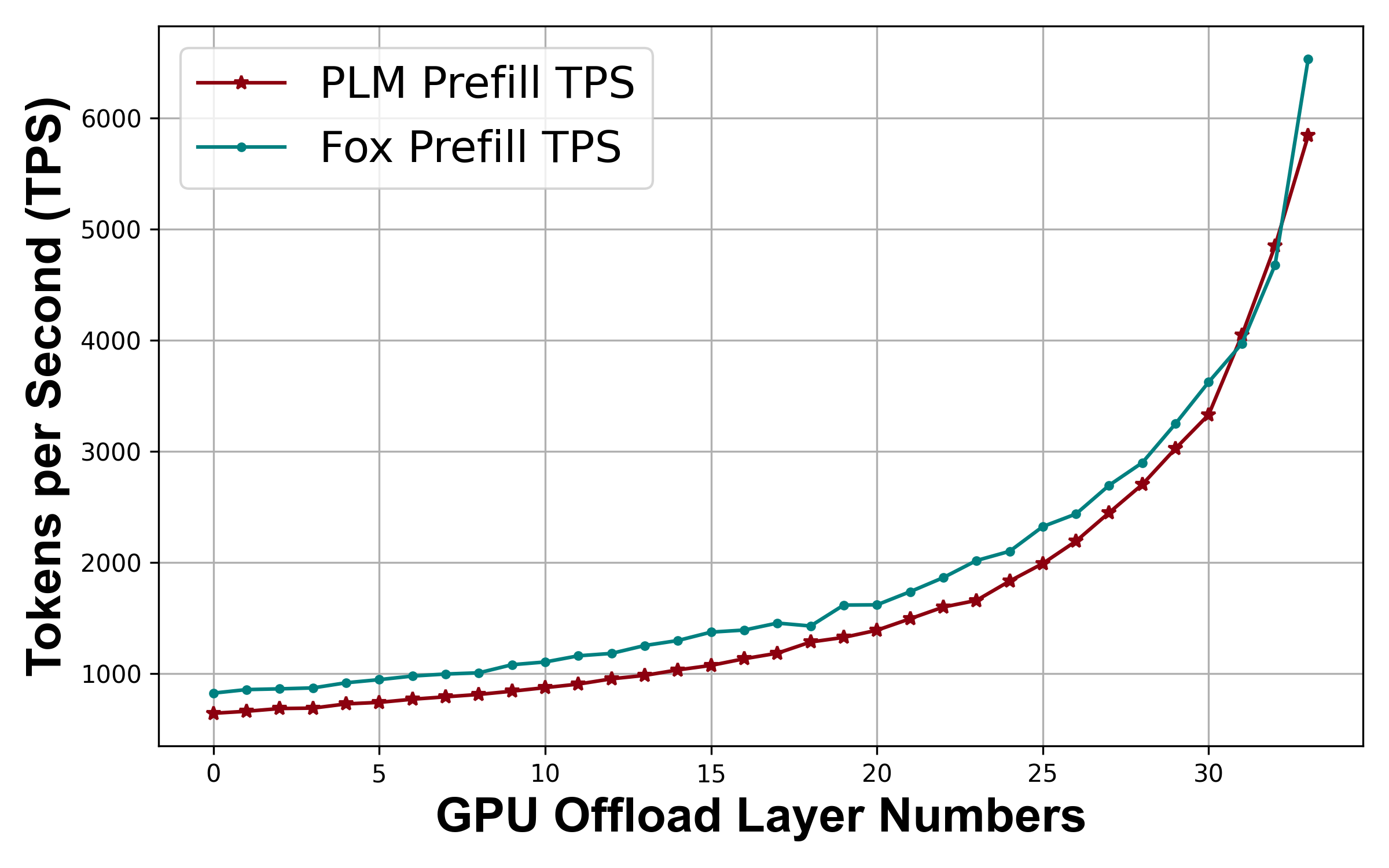}
  }
  \subfigure[Effect of GPU offload layer number on tokens per second (TPS) when model decode.]{
      \label{latency_iter2}
      \includegraphics[width=0.47\linewidth]{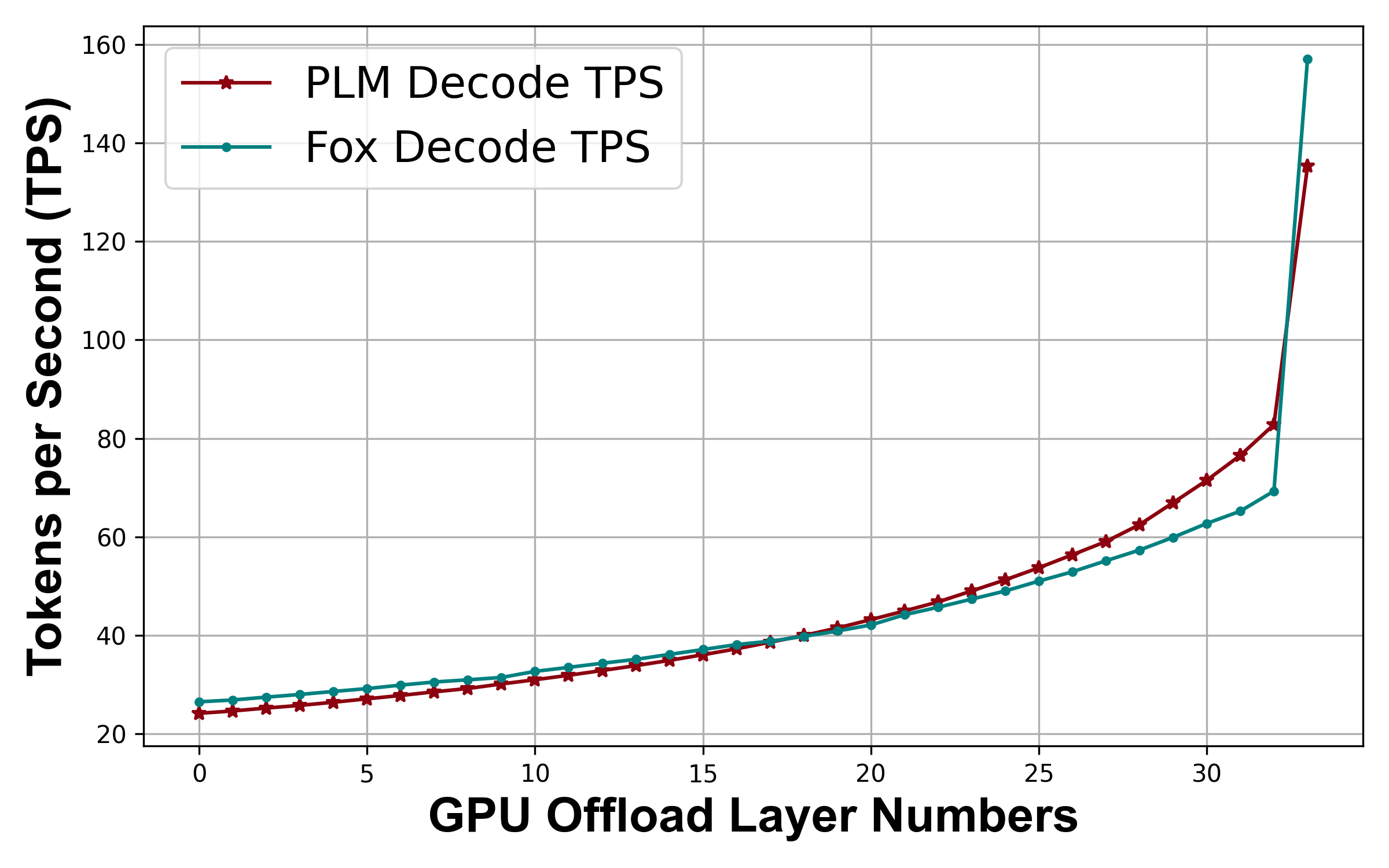}
  }
  \caption{Effect of GPU offload layer number (\thellm{} vs Fox).}
  \label{latency_iter}
\end{figure}

In many edge computing scenarios, models often cannot be fully loaded into GPU memory or system RAM, necessitating offloading~\cite{alizadeh2023llm}. Under these conditions, it is crucial to minimize the computational workload for each portion of the model being loaded. Given that deep learning models rely on stacked hidden layers for search and inference, the computational burden at each layer is particularly critical. \thellm{} mitigates this issue by leveraging KV cache storage and sparse activations to reduce computational overhead. Empirically, we set up an experiment to support the opinion shown in~\autoref{tab:iterload} and~\autoref{latency_iter}.

Experimental results validate this advantage. Specifically, when configuring \textbf{llama.cpp} to load and infer only one layer at a time, we observed clear benefits from \thellm{}'s design, highlighting its efficiency in resource-constrained environments.

\begin{figure}[h]
    \centering
    \includegraphics[width=0.6\linewidth]{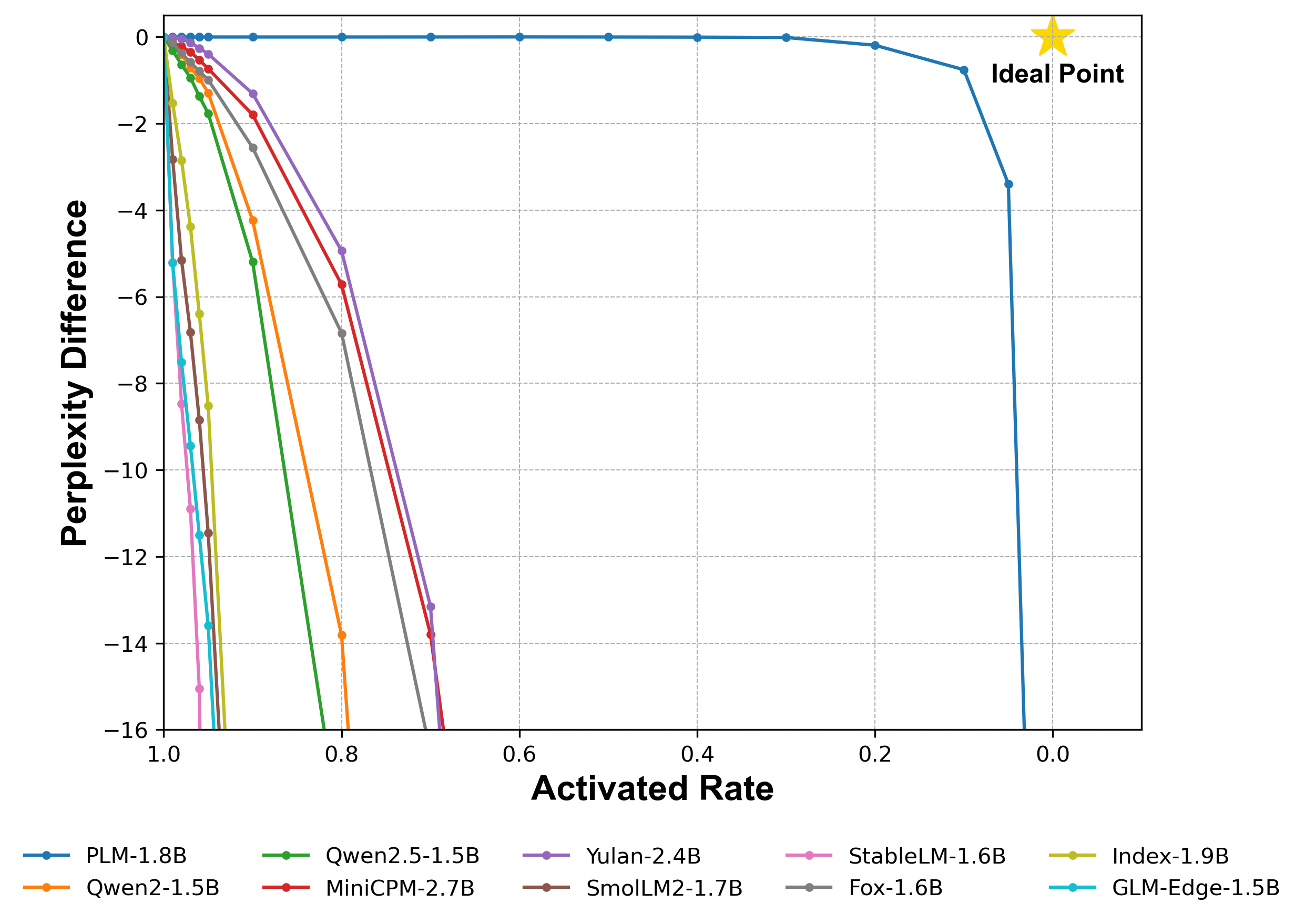}
    \caption{Perplexity increment along with pushing the model to sparse by masking the activated parameters.}
    \label{fig:sparse_eval}
\end{figure}
% \dc{Following Professor Wang's feedback, we've revised the illustration and caption for clarity. Jun and I have decided to label the x-axis "Masked Parameters."}
Moreover, we study the effective of pushing the model to sparsity. Refer to~\cite{song2024turbo}, we design an algorithm to study the sparsity in different baselines. As shown in the~\autoref{sparseexp}, the pusedo code of the experiments is~\autoref{algosparse}, we calculate and determine the activation sparsity rate in an MLP layer of an LLM, using NVIDIA A10. It first computes the activation output—optionally incorporating a gating mechanism—and establishes a baseline perplexity from a benchmark dataset. The algorithm then iteratively masks out the smallest $r\%$ of activation values (based on their absolute magnitude) and recalculates the model's perplexity by taking the exponential of the negative log likelihood. If masking at a particular sparsity level results in a perplexity reduction of at least \emph{1}, that $r\%$ is identified as the activation sparsity rate. Otherwise, it concludes that no tested sparsity rate meets the desired condition. 

% \ly{To clearly demonstrate the relationship between model perplexity and sparsity rates, we plotted the difference in perplexity (between sparse and dense models) on the y-axis against the sparsity rate (the proportion of deactivated neurons, ranging from 0 to 1) on the x-axis. As shown in Figure X, the curves for various models exhibit distinct trends. Notably, the ideal point (unachievable in practice) represents the theoretical optimum where the MLP layer is fully sparse yet maintains perfect performance—a condition where all neurons could remain inactive without degrading model quality.}
More specifically,~\autoref{fig:sparse_eval} presents the impact of sparsity on model perplexity. We plotted the difference in perplexity between sparse and dense models against the sparsity rate (ranging from 0 to 1). This allows us to observe the performance cost associated with increasing neural deactivation. The distinct curves for various models reveal the sensitivity of each architecture to sparsity. Notably, the theoretical ``ideal point'', that full sparsity with no perplexity increase, represents an unattainable goal, where the MLP layer could be completely deactivated without any loss in model quality.

Thus, we can determine the number of parameters actively involved in computation within a model after passing through the activation layer of the MLP. If the computed value is zero, considering it from a hardware perspective, multiplication by zero can be optimized for acceleration. Consequently, from this standpoint, \thellm{} exhibits a significant advantage. As illustrated in the statistics, the sparsity rate of \thellm{} reaches 90.9\% in MLP, and 74.3\% in total.

On the other hand, we can also analyze this from another perspective, by evaluating the sparsity-induced activation in the MLP, we can quantify the reduction in computational load. As shown in the ~\autoref{tab:sparse_analysis}, \thellm{} more clearly demonstrates the impact of sparse activation on the overall parameter utilization in this scenario.

\begin{table}[h]
\centering
\caption{The actual amount of parameters are calculated during the model inference.}
\label{tab:sparse_analysis}
\resizebox{0.9\textwidth}{!}{%
\begin{threeparttable}
\begin{tabular}{@{}ccccccc@{}}
\toprule
\textbf{Models} & \textbf{Raw Size} & \textbf{Score $\uparrow$} & \textbf{\begin{tabular}[c]{@{}c@{}}Activation\\ Sparsity Rate\tnote{1} \end{tabular} $\uparrow$} & \textbf{\begin{tabular}[c]{@{}c@{}}Masked \\ Parameters\end{tabular} $\uparrow$} & \textbf{\begin{tabular}[c]{@{}c@{}}Excuted\\ Parameters\end{tabular} $\downarrow$} & \textbf{\begin{tabular}[c]{@{}c@{}}Sparsity\\ Rate\end{tabular} $\downarrow$} \\ \midrule

\textbf{MiniCPM-2.6B}    & 2.74 & 51.13 & 0.0700      & 0.0743        & 2.5357& 0.9715    \\
\textbf{Yulan-Mini-2.4B} & 2.42 & 59.25 & 0.0900      & 0.0929        & 2.3271& 0.9616    \\
\textbf{Index-1.9B}      & 1.91 & 48.02 & 0.0010      & 0.0009        & 1.9091& 0.9995    \\
\textbf{SmolLM2-1.7B}    & 1.71 & 49.95 & 0.0010      & 0.0008        & 1.7092& 0.9995    \\
\textbf{StableLM2-1.6B}  & 1.64 & 38.29 & 0.0010      & 0.0005        & 1.6395& 0.9997    \\
\textbf{Fox-1-1.6B}      & 1.61 & 42.25 & 0.0600      & 0.0322        & 1.6378& 0.9807    \\
\textbf{GLM-Edge-1.5B}   & 1.59 & 38.92 & 0.0060      & 0.0042        & 1.5858& 0.9973    \\
\textbf{Qwen2.5-1.5B}    & 1.54 & 59.84 & 0.0400      & 0.0308        & 1.5092& 0.9800    \\
\textbf{Qwen2-1.5B}      & 1.54 & 54.48 & 0.0500      & 0.0385        & 1.5015& 0.9750    \\
\textbf{PLM-1.8B-Base}   & 1.83 & 57.29    & \textbf{0.9090}   & \textbf{0.4832}    & \textbf{1.3968}     & \textbf{0.7430}  \\ \bottomrule
\end{tabular}
\begin{tablenotes}
\footnotesize
    \item[1] The Activation Sparsity Rate calculation algorithm is illustrated in~\autoref{sparseexp}
\end{tablenotes}
\end{threeparttable}
}
\end{table}

% \dc{Add sparse experiments}

% \subsection{Deployment on peripheral scenario}

% We implement several deployment methods, excluding vLLM. Specifically, we introduce \thellm{} to SgLang, llama.cpp, and PowerInfer, incorporating engineering-level optimizations, so as to put the \thellm{} in to mobile phone, Jetson Orin, and Raspberry Pi.

% \paragraph{Mobile Phone}

% \paragraph{Jetson Orin NX 16GB}

% \paragraph{Raspberry Pi}

% \subsection{Adaptation and Beyond}

% During the early trial in SFT training, we found a potentially unstablility if we train the \thellm{} via Deepspeed, \dc{I think we can write some facts during the SFT training, to raise the discussion of MLA sft and SFT with Ds, \url{https://basicv8vc.github.io/posts/zero/}}
\section{Discussion: Future Works and Direction}

Throughout the entire process of designing and training \thellm{}, we draw lessons from investigating the needs of peripheral language models, empirically analyze the characteristics of LLMs on the edge, and explore the potential development of peripheral machine learning. Our findings highlight the crucial importance of treating the training of peripheral LLMs as a co-design field that integrates both modeling and system development.

\subsection{Future Works}

Based on the empirical cases in \thellm{} design, training and deployment, we discuss the future works:

\paragraph{Peripheral LLM Adaptation} To enable personalized language models on resource-constrained edge devices, we will develop efficient on-device adaptation techniques. This includes sparse update mechanisms for selective parameter modification~\cite{rios2025sparsity}, quantization-aware adaptation for low-precision fine-tuning~\cite{kwon2022alphatuning}, and memory-efficient gradient accumulation with incremental learning to distribute computational load across sessions.

\paragraph{Privacy-Preserving Federated Learning} We will establish a federated learning framework for edge devices that ensures privacy through differential privacy, knowledge distillation, and secure aggregation~\cite{arazzi2025secure}. Adaptive privacy budgeting will dynamically manage protection levels~\cite{wang2024adaptive}. This enables bidirectional knowledge transfer, allowing cloud models to learn from privacy-sanitized edge updates and edge models to benefit from global improvements~\cite{arazzi2025secure}.

\paragraph{Edge-Cloud Collaborative Framework} We will develop an edge-cloud framework that optimizes task distribution based on complexity and latency~\cite{jin2024collm}. Edge models will handle simple, real-time queries, while complex tasks will be delegated to the cloud. Efficient communication channels will minimize data transfer and maintain contextual coherence. Dynamic task allocation~\cite{wang2025research} based on various factors will create an adaptive system, providing users with a seamless experience that combines immediate responses with deep reasoning capabilities.

\paragraph{Omni-MLLM and Enhanced Capabilities} We will expand \thellm{} to a multimodal model~\cite{fu2024vita,guo2025m2}, capable of processing and integrating diverse data types such as text, images, audio, and video to enable advanced applications like visual question answering, image captioning, and multimodal robotics. This enhancement will allow the model to perform tasks requiring cross-modal understanding, such as aligning video footage with textual instructions or analyzing 3D engineering data alongside textual descriptions. Model scaling will be achieved through expansion techniques~\cite{wang2023lemon}. Reasoning abilities will be improved by fine-tuning with distilled reasoning data, math and code capabilities will be enhanced with RL and rule-verifiable data~\cite{guo2025deepseek}. Furthermore, the overall capabilities of \thellm{} could be enhanced by integrating its domain-specialized counterpart~\cite{tinyr1proj} using tools like MergeKit~\cite{goddard2024arcee}. Long-context modeling skills can be further improved through RAG-like methods~\cite{xiao2025activation}. Lastly, detecting and mitigating hallucinations~\cite{zhang2023enhancing} remains a crucial aspect of refinement.

\paragraph{Small Language Model for Reasoning} We will investigate methodologies to enhance the reasoning capabilities of SLMs while preserving computational efficiency. Knowledge distillation~\cite{team2024gemma} and supervised fine-tuning~\cite{guo2025deepseek} are proven techniques to transfer knowledge from larger models, thus improving the accuracy of reasoning without significantly increasing resource demands. RL also presents a viable approach for optimizing reasoning processes; however, SLMs often face challenges with sparse reward signals when utilizing RL. Process-based reinforcement learning techniques~\cite{zhang2025lessons,wang2024openr} may offer a more effective solution to guide SLMs. During the inference stage, adaptive test-time scaling emerges as a highly promising research direction, enabling dynamic adjustment of computational pathways based on input complexity to ensure efficient inference. Additionally, generative reward models~\cite{mahan2024generative} can provide supplementary supervision to SLMs, further boosting their reasoning performance. These advancements are poised to significantly elevate the reasoning capabilities of SLMs. Additionally, introducing peripheral language models in multi-agent scenarios can leverage Multi-Agent Reinforcement Learning (MARL) to elicit meta-thinking behaviors~\cite{Wan2025ReMALT}, enabling them to engage in reflective reasoning and navigate complex decision-making more effectively.

\subsection{Direction}

Empirically, MLA is designed to reduce the memory footprint of the Key-Value (KV) cache during inference by compressing key and value representations into a latent vector. This compression can alleviate memory bottlenecks, which is beneficial for deploying LLMs on edge devices with limited resources.

However, MLA introduces additional computational overhead due to the need to decompress the latent vectors during attention computations. This decompression step increases the number of matrix multiplications, potentially impacting inference speed~\cite{flashmla2025}. For edge devices, which often have constraints on both memory and processing power, this trade-off between reduced memory usage and increased computational demand must be carefully considered.

Consequently, while MLA can help manage memory limitations on edge devices, its suitability depends on the specific balance between memory capacity and computational capabilities of the target hardware. Evaluating this trade-off is crucial to determine if MLA is appropriate for a given edge deployment scenario. 

On the other hand, from the MLSys co-design perspective, the future of sparse LLMs hinges on a deep integration of algorithmic sparsity with hardware optimizations~\cite{xu2024device}. This involves designing custom accelerators that can efficiently handle sparse operations, optimizing memory architectures to minimize data movement, and implementing on-chip mechanisms for dynamic sparsity management. Hardware-aware sparsity~\cite{yuan2025native} is crucial for realizing the potential efficiency gains of sparse LLMs, enabling faster inference and training while reducing energy consumption.

System-level optimizations are equally vital. This includes developing efficient sparse data formats, optimizing communication protocols for distributed training, and leveraging compiler optimizations to exploit sparsity in LLMs~\cite{zheng2023pit}. Memory hierarchy optimization and power efficiency are also critical considerations. By creating systems that are aware of sparsity patterns and can dynamically adjust resource allocation, we can unlock the full potential of sparse LLMs. This necessitates the creation of libraries and frameworks that can handle sparse data formats seamlessly across different hardware platforms.

Ultimately, the future direction emphasizes algorithm-hardware co-design~\cite{VIDA}. This involves developing training algorithms that produce hardware-friendly sparsity patterns~\cite{agarwalla2024enabling,frantar2023sparsegpt}, enabling dynamic sparsity adaptation during runtime~\cite{yang2025lserve,zhou2025progressivesparseattentionalgorithm}, and combining sparsity with quantization techniques to further reduce computational and memory footprints~\cite{zhang2024q}. By tightly coupling algorithmic innovations with hardware capabilities, we can create systems that can efficiently execute sparse LLMs, leading to significant advances in performance, energy efficiency, and scalability. This holistic approach will pave the way for the widespread deployment of sparse LLMs in resource-constrained environments.

The deployment of language models in peripheral scenario presents unique challenges and requirements distinct from cloud-based implementations. Our research identifies four critical directions that guide the development of effective edge-side models.

\begin{itemize}[leftmargin=1.2em]
\item Edge-side models require rapid response to meet time-sensitive application demands, as high latency significantly reduces their utility~\cite{zhang2024edgeshard}. To achieve this within fixed hardware constraints, architectural innovations must prioritize inference speed without compromising model quality~\cite{zheng2024review}.
\item Edge-side models enhance privacy by processing data locally, mitigating risks associated with remote server transmission~\cite{yu2024edge}. To maintain this privacy, models must effectively extract information from limited local data while adhering to strict data movement restrictions.
\item Edge models must be highly resource-efficient to avoid impacting primary device functions, operating within strict computational and storage limits. This requires model compression~\cite{hohman2024model}, efficient attention mechanisms~\cite{chen2025attentionengine}, and dynamic resource allocation~\cite{shanmugam2023dynamic} to ensure they act as auxiliary, not primary, resource consumers.
\item Power efficiency is paramount for edge models on battery-powered devices, as extended computation directly impacts battery life and user experience. Therefore, energy consumption must be a primary design consideration, employing techniques like sparse activation and precision reduction to minimize power draw without sacrificing performance~\cite{paramanayakam2024less}.
\end{itemize}

Rapid response, privacy, resource efficiency, and power consumption are not mere limitations, but core challenges driving the development of peripheral LLMs. By focusing on these parameters, we aim to enable seamless integration of intelligent AI capabilities into peripheral and edge devices, delivering real-time responses while prioritizing user privacy and respecting device limitations.

\section{Conclusion}
We present \thellm{}—an Edge Language Model optimized for resource-limited devices through a co-designed architecture combining Multi-head Latent Attention and squared ReLU for enhanced sparsity and reduced memory usage. \thellm{} surpasses existing small language models on standard benchmarks and runs efficiently on devices ranging from PCs to mobile phones and Raspberry Pis. Moreover, we discuss the future direction of LLM on the edge. Finally, we release \thellm{}-1.8B-Base and \thellm{}-1.8B-Instruct to foster further research in peripheral computing. Along with the design and training of \thellm{}, we analyze the needs, features, and potential development of peripheral language models, highlighting the crucial importance of viewing peripheral LLMs as a co-design field that integrates both hardware and algorithm development.

\newpage
%Bibliography
\bibliographystyle{unsrt}  
\bibliography{references}  

\newpage
\appendix
\section{Architecture Searching Experiments}
\label{arch-search}
We detail our sandbox experiments of model architecture. Through these experiments, we seek to find the most appropriate configuration making a balance between performance and computation. Therefore, the hyperparameters $n_{layer}$, $d_{head}$, and $n_{head}$ should be focused. We have conducted several architecture experiments and present the most relevant candidates here. We present the candidate configuration in table \ref{tab:arch-search-list}, and computation details in table \ref{tab:arch-can-attr}. 

\begin{table}[h]
\caption{Architectures search candidates.}
\label{tab:arch-search-list}
\centering
\resizebox{0.7\textwidth}{!}{%
\begin{tabular}{@{}cc|cccccccc@{}}
\toprule
\textbf{No.} & \textbf{\#Params}  & $n_{layer}$ & $d_{model}$ & $n_{head}$ & $d_{head}$  & $d_{ffn}$ & $kv_{rank}$ & $q_{rank}$ & $d_{rope}$ \\ \midrule
\textbf{1} & 1.54B & 28 & 2816 & 44 & 64 & 7040 & 256 & 768 & 32 \\
\textbf{2} & 1.21B & 32 & 2304 & 36 & 64 & 5760 & 256 & 768 & 32 \\
\textbf{3} & 1.47B & 32 & 2560 & 40 & 64 & 6400 & 256 & 768 & 32 \\
\textbf{4} & 1.36B & 36 & 2304 & 36 & 64 & 5760 & 256 & 768 & 32 \\
\textbf{5} & 1.51B & 40 & 2304 & 36 & 64 & 5760 & 256 & 768 & 32 \\
\textbf{6} & 1.55B & 36 & 2048 & 32 & 64 & 8192 & 256 & 768 & 32 \\
\textbf{7} & 1.54B & 32 & 2048 & 16 & 128 & 8192 & 512 & 1536 & 64 \\  \bottomrule
\end{tabular}
}
\end{table}
\vspace{-1em}
\begin{table}[h]
\caption{Hardware-aware attributes of the candidates, tested on Raspberry Pi.}
\label{tab:arch-can-attr}
\centering
\resizebox{1.0\textwidth}{!}{%
\begin{tabular}{@{}c|cccccccccc@{}}
\toprule
\multirow{3}{*}{\textbf{No.}}& \multirow{2}{*}{MACs} & \multirow{2}{*}{FLOPs} & \multirow{2}{*}{FWD+BWD} & \multirow{2}{*}{FWD+BWD} & \multicolumn{2}{c}{Factors}  & \multirow{2}{*}{TTFT} & \multirow{2}{*}{Time/Layer} & \multirow{2}{*}{Token Rate} & \multirow{2}{*}{Peak Memory} \\\cline{6-7}
&\multirow{2}{*}{(G)}&\multirow{2}{*}{(G)}&\multirow{2}{*}{(GMACs)}&\multirow{2}{*}{(GFLOPs)}&\multirow{2}{*}{MACs/Param}&\multirow{2}{*}{FLOPs/Param}&\multirow{2}{*}{(ms)}&\multirow{2}{*}{(ms)}&\multirow{2}{*}{(tok/s)}&\multirow{2}{*}{(GB)}\\
&&&&&&&&&&\\
\midrule
\textbf{1} & 206 & 412 & 618 & 1240 & 103 & 205 & 9075 & 324 & 7.8E-2 & 7.0E-5  \\
\textbf{2} & 164 & 327  & 491 & 982 & 102 & 205 & 3178 & 99 & 2.6E-1  & 7.8E-5  \\
\textbf{3} & 198 & 396 & 594 & 1190 & 104 & 208  & 3775 & 118 & 2.2E-1 & 6.6E-5  \\
\textbf{4} & 184 & 368 & 552 & 1100 & 105 & 209 & 3592 & 100 & 2.3E-1 & 6.8E-5  \\
\textbf{5} & 205  & 409 & 614 & 1230 & 107 & 214  & 4040 & 101 & 2.1E-1 & 7.1E-5  \\
\textbf{6} & 207 & 414 & 622 & 1240 & 109 & 218 & 3995 & 111 & 2.2E-1 & 7.2E-5 \\
\textbf{7} & 203 & 406 & 609 & 1220 & 108 & 216 & 3354 & 105 & 2.6E-1 & 6.7E-5 \\  \bottomrule   
\end{tabular}
}
\end{table}
\vspace{-1em}
\begin{table}[h]
\caption{Sandbox experiments hyperparameters.}
\label{tab:hyper_seach}
\centering
\resizebox{0.3\textwidth}{!}{%
\begin{tabular}{cc}
\toprule
Hyperparameter & Value  \\ \midrule
Global Batch Size & 960  \\ 
Micro Batch Size & 3  \\ 
Max Learning Rate & 5.00E-04  \\ 
Min Learning Rate & 5.00E-06  \\ 
Weight Decay & 0.1  \\ 
Adam $\beta_{1}$ & 0.9  \\ 
Adam $\beta_{2}$ & 0.95  \\ 
Clip Grad & 1.0  \\ 
Attention Dropout & 0  \\ 
Hidden Dropout & 0  \\ 
Sequence Length & 4096  \\ 
\bottomrule
\end{tabular}%
}
\end{table}
\vspace{-1em}
\begin{table}[H]
\caption{Training data distribution and allocation.}
\label{tab:data_pretrain_sandbox}
\centering
\resizebox{0.9\textwidth}{!}{%
\begin{tabular}{@{}ccccc@{}}
\toprule
\textbf{Data Source} & \textbf{Selected Size (B)} & \textbf{Training Allocation (B)} & \textbf{Normalized Ratio} & \textbf{Category} \\ \midrule
mlfoundations/dclm-baseline-1.0 & 1485          & 59                     & 0.59             & English  \\
m-a-p/MAP-CC                    & 518           & 22                     & 0.22             & Chinese  \\
BAAI/CCI3-HQ                    & 120           & 7                     & 0.07             & Chinese  \\
EleutherAI/proof-pile-2         & 50            & 2                      & 0.02             & Math     \\
GAIR/MathPile                   & 9             & 1                       & 0.01             & Math     \\
bigcode/starcoderdata           & 226           & 10                     & 0.10             & Code     \\
\bottomrule
\end{tabular}%
}
\end{table}

In each experiment, we strive to adopt our final pre-training configuration, as shown in the table~\autoref{tab:hyper_seach}, using WSD learning rate scheduler and a corpus with similar data distribution as the final pre-train corpus, shown in~\autoref{tab:data_pretrain_sandbox}.

\section{Datasets}

\subsection{Pre-Train Dataset}
\label{pretrain-dataset}
\begin{table}[h]
\caption{Training data distribution and allocation.}
\label{tab:data_pretrain_1}
\centering
\resizebox{\textwidth}{!}{%
\begin{tabular}{@{}cccccc@{}}
\toprule
\textbf{Phase} & \textbf{Data Source} & \textbf{Selected Size (B)} & \textbf{Training Allocation (B)} & \textbf{Normalized Ratio} & \textbf{Category} \\ \midrule
1  & mlfoundations/dclm-baseline-1.0 & 1485          & 966                     & 0.59             & English  \\
1  & m-a-p/MAP-CC                    & 518           & 363                     & 0.22             & Chinese  \\
1  & BAAI/CCI3-HQ                    & 120           & 120                     & 0.07             & Chinese  \\
1  & EleutherAI/proof-pile-2         & 50            & 39                      & 0.02             & Math     \\
1  & GAIR/MathPile                   & 9             & 9                       & 0.01             & Math     \\
1  & bigcode/starcoderdata           & 226           & 158                     & 0.10             & Code     \\ \midrule
2  & HuggingFaceTB/smollm-corpus         & 222           & 220                     & 0.4              & English  \\
2  & allenai/dolmino-mix-1124/data/flan  & 74            & 80                      & 0.15      & English  \\
2  & opencsg/chinese-fineweb-edu-v2      & 221           & 120                     & 0.22      & Chinese  \\
2  & allenai/dolmino-mix-1124/data/pes2o & 60            & 60                      & 0.1      & Paper    \\
2  & OpenCoder-LLM/opc-annealing-corpus  & 64            & 62                      & 0.11      & Code     \\
2  & bigcode/the-stack-v2                & 8             & 8                       & 0.01      & Code     \\\midrule
3  & HuggingFaceTB/smollm-corpus         & 220           & 100                     &     0.36         & English  \\
3  & opencsg/chinese-fineweb-edu-v2      & 221           & 40                     &   0.14  & Chinese \\
3  & yulan-team/YuLan-Mini-Datasets      & 96            &  90                   &  0.32  & Code,QA \\
3  & allenai/dolmino-mix-1124/data/math      & 11           & 10                     &   0.04  & Math \\
% 5  & allenai/dolmino-mix-1124/data/pes2o      & 60           & 60                     &     & Paper \\
3  & plm/sft-data      &      46     &            40          &  0.14   &  QA,MATH,Code \\
\bottomrule
\end{tabular}%
}
\end{table}

\subsection{Internal Pre-Train Dataset}
\label{internal-pretrain-dataset}

We utilized public available supervised fine-tuning datasets to re-structure into pre-train data. The collected datasets are listed in~\autoref{tab:sft-data}.

% \newpage

\subsection{Post-Train Data}

\paragraph{Supervised fine-tuning data}
Based on the collected SFT-like data, we gather and form the two phases SFT datasets, shown as~\autoref{tab:sft2datasets}.
\label{sft-data}

\begin{table}[h]
\centering
\caption{Datasets used in phase 1 and phase 2 of SFT.}
\resizebox{0.7\textwidth}{!}{%
\begin{tabular}{>{\raggedright\arraybackslash}p{0.45\textwidth} >{\raggedright\arraybackslash}p{0.45\textwidth}}
\toprule
\textbf{Phase 1 Datasets} & \textbf{Phase 2 Datasets} \\
\midrule
\href{https://huggingface.co/datasets/allenai/tulu-3-sft-mixture}{\texttt{allenai/tulu-3-sft-mixture}} & \href{https://huggingface.co/datasets/ibm/ProvoQ}{\texttt{ibm/ProvoQ}} \\
\href{https://huggingface.co/datasets/teknium/GPTeacher-General-Instruct}{\texttt{teknium/GPTeacher-General-Instruct}} & \href{https://huggingface.co/datasets/ibm/AttaQ}{\texttt{ibm/AttaQ}} \\
\href{https://huggingface.co/datasets/O1-OPEN/OpenO1-SFT}{\texttt{O1-OPEN/OpenO1-SFT}} & \href{https://huggingface.co/datasets/allenai/wildguardmix}{\texttt{allenai/wildguardmix}} \\
\href{https://huggingface.co/datasets/O1-OPEN/OpenO1-SFT-Ultra}{\texttt{O1-OPEN/OpenO1-SFT-Ultra}} & \href{https://huggingface.co/datasets/allenai/tulu-3-sft-mixture}{\texttt{allenai/tulu-3-sft-mixture}} \\
\href{https://huggingface.co/datasets/O1-OPEN/OpenO1-SFT-Pro}{\texttt{O1-OPEN/OpenO1-SFT-Pro}} & \href{https://huggingface.co/datasets/hkust-nlp/dart-math-hard}{\texttt{hkust-nlp/dart-math-hard}} \\
\href{https://huggingface.co/datasets/causal-lm/ultrachat}{\texttt{causal-lm/ultrachat}} & \href{https://huggingface.co/datasets/hkust-nlp/dart-math-uniform}{\texttt{hkust-nlp/dart-math-uniform}} \\
\href{https://huggingface.co/datasets/Open-Orca/SlimOrca}{\texttt{Open-Orca/SlimOrca}} & \href{https://huggingface.co/datasets/meta-math/MetaMathQA_GSM8K_zh}{\texttt{meta-math/MetaMathQA\_GSM8K\_zh}} \\
\href{https://huggingface.co/datasets/GAIR/lima}{\texttt{GAIR/lima}} & \href{https://huggingface.co/datasets/meta-math/MetaMathQA}{\texttt{meta-math/MetaMathQA}} \\
\href{https://huggingface.co/datasets/BelleGroup/train_2M_CN}{\texttt{BelleGroup/train\_2M\_CN}} & \href{https://huggingface.co/datasets/HuggingFaceTB/math_tasks}{\texttt{HuggingFaceTB/math\_tasks}} \\
\href{https://huggingface.co/datasets/suolyer/webqa}{\texttt{suolyer/webqa}} & \href{https://huggingface.co/datasets/allenai/tulu-3-sft-personas-math}{\texttt{allenai/tulu-3-sft-personas-math}} \\
\href{https://huggingface.co/datasets/YeungNLP/firefly-train-1.1M}{\texttt{YeungNLP/firefly-train-1.1M}} & \href{https://huggingface.co/datasets/nvidia/OpenMathInstruct-2}{\texttt{nvidia/OpenMathInstruct-2}} \\
\href{https://huggingface.co/datasets/deepctrl/deepctrl-sft-data}{\texttt{deepctrl/deepctrl-sft-data}} & \href{https://huggingface.co/datasets/BAAI/Infinity-Instruct}{\texttt{BAAI/Infinity-Instruct}} \\
\href{https://huggingface.co/datasets/shibing624/sharegpt_gpt4}{\texttt{shibing624/sharegpt\_gpt4}} & \href{https://huggingface.co/datasets/BAAI/IndustryInstruction}{\texttt{BAAI/IndustryInstruction}} \\
& \href{https://huggingface.co/datasets/BAAI/InfinityMATH}{\texttt{BAAI/InfinityMATH}} \\
& \href{https://huggingface.co/datasets/BAAI/AquilaEdu-Instruct}{\texttt{BAAI/AquilaEdu-Instruct}} \\
& \href{https://huggingface.co/datasets/BAAI/AquilaMed-Instruct}{\texttt{BAAI/AquilaMed-Instruct}} \\
& \href{https://huggingface.co/datasets/amztheory/codegen-instruct-python-1k}{\texttt{amztheory/codegen-instruct-python-1k}} \\
& \href{https://huggingface.co/datasets/amztheory/codegen-instruct-CPlus-1k}{\texttt{amztheory/codegen-instruct-CPlus-1k}} \\
& \href{https://huggingface.co/datasets/AI-MO/NuminaMath-CoT}{\texttt{AI-MO/NuminaMath-CoT}} \\
& \href{https://huggingface.co/datasets/nvidia/AceMath-Instruct-Training-Data}{\texttt{nvidia/AceMath-Instruct-Training-Data}} \\
& \href{https://huggingface.co/datasets/glaiveai/glaive-code-assistant-v2}{\texttt{glaiveai/glaive-code-assistant-v2}} \\
& \href{https://huggingface.co/datasets/sahil2801/CodeAlpaca-20k}{\texttt{sahil2801/CodeAlpaca-20k}} \\
& \href{https://huggingface.co/datasets/iamtarun/python_code_instructions_18k_alpaca}{\texttt{iamtarun/python\_code\_instructions\_18k\_alpaca}} \\
& \href{https://huggingface.co/datasets/likaixin/InstructCoder}{\texttt{likaixin/InstructCoder}} \\
& \href{https://huggingface.co/datasets/Codegen/codegen-instruct}{\texttt{Codegen/codegen-instruct}} \\
& \href{https://huggingface.co/datasets/hfl/ruozhiba_gpt4_turbo}{\texttt{hfl/ruozhiba\_gpt4\_turbo}} \\
& \href{https://huggingface.co/datasets/wiki_qa}{\texttt{wiki\_qa}} \\
& \href{https://huggingface.co/datasets/ultrachat_200k}{\texttt{ultrachat\_200k}} \\
\bottomrule
\end{tabular}
}
\label{tab:sft2datasets}
\end{table}

\paragraph{Reinforcement learning with human feedback data.}

For the reinforcement learning with human feedback, we adopt the data from~\autoref{tab:rlhf-data}.

\begin{table}[h]
\caption{Reinforcement learning with human feedback dataset.}
\label{tab:rlhf-data}
\centering
\resizebox{0.6\textwidth}{!}{%
\begin{tabular}{@{}ccccc@{}}
\toprule
\textbf{No} & \textbf{Dataset} & \textbf{Sample}  & \textbf{Class}   \\ \midrule
1 & allenai/llama-3.1-tulu-3-70b-preference-mixture & 334302 & General Dialogue \\
2 & HuggingFaceH4/ultrafeedback\_binarized & 5120 & General Dialogue \\
3 & coseal/CodeUltraFeedback & 9433 & Code \\
\bottomrule
\end{tabular}
}
\end{table}

% \usepackage{booktabs}
% \vspace{-5pt}
\begin{table}[h]
\caption{Supervised fine-tuning dataset collection.}
\label{tab:sft-data}
\centering
\resizebox{0.61\textwidth}{!}{%
\begin{tabular}{@{}ccccc@{}}
\toprule
\textbf{No} & \textbf{Dataset}                           & \textbf{Sample}    & \textbf{Class}   \\ \midrule
1  & ELM-Internal-Instruct                               & 123569     & Knowledge        \\
2  & amztheory/codegen-instruct-python-1k                & 1234       & Code             \\
3  & amztheory/codegen-instruct-CPlus-1k                 & 813        & Code             \\
4  & PowerInfer/LONGCOT-Refine-500K                      & 521921     & Thought          \\
5  & PowerInfer/QWQ-LONGCOT-500K                         & 498082     & Thought          \\
6  & causal-lm/ultrachat                                 & 5084540    & General          \\
7  & allenai/wildjailbreak                               & 261538     & Safety           \\
8  & ibm/ProvoQ                                          & 2705       & Safety           \\
9  & ibm/AttaQ                                           & 1402       & Safety           \\
10 & allenai/wildguardmix                                & 37976      & Safety           \\
11 & allenai/tulu-3-sft-mixture                          & 939343     & General          \\
12 & nvidia/ChatQA-Training-Data                         & 442261     & General          \\
13 & HuggingFaceTB/smoltalk                              & 1098865    & General          \\
14 & HuggingFaceH4/ultrachat\_200k                       & 515311     & General Dialogue \\
15 & WizardLMTeam/WizardLM\_evol\_instruct\_V2\_196k     & 143000     & General          \\
16 & anon8231489123/ShareGPT\_Vicuna\_unfiltered         & 94145      & General          \\
17 & kanhatakeyama/wizardlm8x22b-logical-math-coding-sft & 284303     & logit            \\
18 & kanhatakeyama/logical-wizardlm-7b                   & 21529967   & logit            \\
19 & Open-Orca/SlimOrca-Dedup                            & 363491     & General          \\
20 & openchat/openchat\_sharegpt4\_dataset               & 105003     & General          \\
21 & LDJnr/Capybara                                      & 16006      & General Dialogue \\
22 & hkust-nlp/deita-10k-v0                              & 10000      & General          \\
23 & hkust-nlp/vrt-baseline                              & 590601     & Math             \\
24 & hkust-nlp/dart-math-hard                            & 585392     & Math             \\
25 & hkust-nlp/dart-math-uniform                         & 590705     & Math             \\
26 & hkust-nlp/dart-math-pool-gsm8k                      & 2738984    & Math             \\
27 & hkust-nlp/dart-math-pool-math                       & 1615233    & Math             \\
28 & meta-math/MetaMathQA\_GSM8K\_zh                     & 463370     & Math             \\
29 & hkust-nlp/gsm8k-fix                                 & 7473       & Math             \\
30 & meta-math/MetaMathQA-40K                            & 40000      & Math             \\
31 & meta-math/GSM8K\_zh                                 & 17582      & Math             \\
32 & meta-math/GSM8K\_Backward                           & 2540       & Math             \\
33 & meta-math/MetaMathQA                                & 395000     & Math             \\
34 & HuggingFaceTB/math\_tasks                           & 21292      & Math             \\
35 & pvduy/synth\_code\_preference\_4k                   & 4052       & Code             \\
36 & allenai/RLVR-GSM                                    & 7473       & Math             \\
37 & allenai/tulu-3-sft-personas-math                    & 149960     & Math             \\
38 & RUC-AIBOX/long\_form\_thought\_data\_5k             & 4922       & Thought          \\
39 & nvidia/ChatQA2-Long-SFT-data                        & 128000     & General          \\
40 & nvidia/OpenMathInstruct-2                           & 21972789   & Math             \\
41 & mengfn/MathOdyssey                                  & 389        & Math             \\
42 & mengfn/longthoughts-synth                           & 14973      & Thought          \\
43 & mengfn/MATH-clean                                   & 7500       & Math             \\
44 & O1-OPEN/OpenO1-SFT                                  & 77685      & Thought          \\
45 & BAAI/Infinity-Instruct                              & 1456927    & General          \\
46 & BAAI/IndustryInstruction                            & 2706143    & General          \\
47 & BAAI/InfinityMATH                                   & 101380     & MATH             \\
48 & BAAI/AquilaEdu-Instruct                             & 269917     & General          \\
49 & BAAI/AquilaMed-Instruct                             & 318357     & General          \\
50 & BAAI/COIG                                           & 275985     & Code             \\
51 & Codegen (LINK)                                      & 4535       & Code             \\
52 & sahil2801/CodeAlpaca-20k                            & 20022      & Code             \\
53 & iamtarun/python\_code\_instructions\_18k\_alpaca    & 18612      & Code             \\
54 & likaixin/InstructCoder                              & 108391     & Code             \\
56 & O1-OPEN/OpenO1-SFT-Pro                              & 125894     & Thought          \\
\multicolumn{2}{c}{\textbf{Total}}                       & \textbf{60533509} &           \\ \bottomrule
\end{tabular}
}
\end{table}

\section{Training Details}

\autoref{tab:argument_pretrain} presents the training hyperparameters used during the three phases of \thellm{} pre-training, highlighting variations in the learning rate schedule, weight decay, batch sizes, and optimization parameters to facilitate effective model convergence across different training stages.

\subsection{Pre-training}
\label{pre_args}

\begin{table}[H]
\caption{Training arguments for pre-training.}
\label{tab:argument_pretrain}
\centering
\resizebox{0.78\textwidth}{!}{%
\begin{tabular}{@{}cccc@{}}
\toprule
\textbf{} & \textbf{\thellm{} Per-train Phase 1} & \textbf{\thellm{} Per-train Phase 2} & \textbf{\thellm{} Per-train Phase 3} \\ \midrule
Global Batch Size & 3744 & 3744 & 3744 \\ 
Micro Batch Size & 3 & 3 & 3 \\ 
Learning Rate & 3.00E-04 & 3.00E-04 & 3.00E-05 \\ 
Min Learning Rate & 3.00E-04 & 5.00E-06 & 0.00E+00 \\ 
Lr Scheduler Type & Warm-up and Stable & Decay & Constant \\ 
Weight Decay & 0.1 & 0.1 & 0.1 \\ 
Adam Beta1 & 0.9 & 0.9 & 0.9 \\ 
Adam Beta2 & 0.95 & 0.95 & 0.95 \\ 
Clip Grad & 1.0 & 1.0 & 1.0 \\ 
Init Method Std & 0.008 & Null & Null \\ 
Attention Dropout & 0 & 0 & 0 \\ 
Hidden Dropout & 0 & 0 & 0 \\ 
Seq Length & 4096 & 4096 & 4096 \\ 
Epoch & 1 & 1 & 1 \\
\bottomrule
\end{tabular}%
}
\end{table}

\begin{figure}[H]
    \centering
    \includegraphics[width=0.5\linewidth]{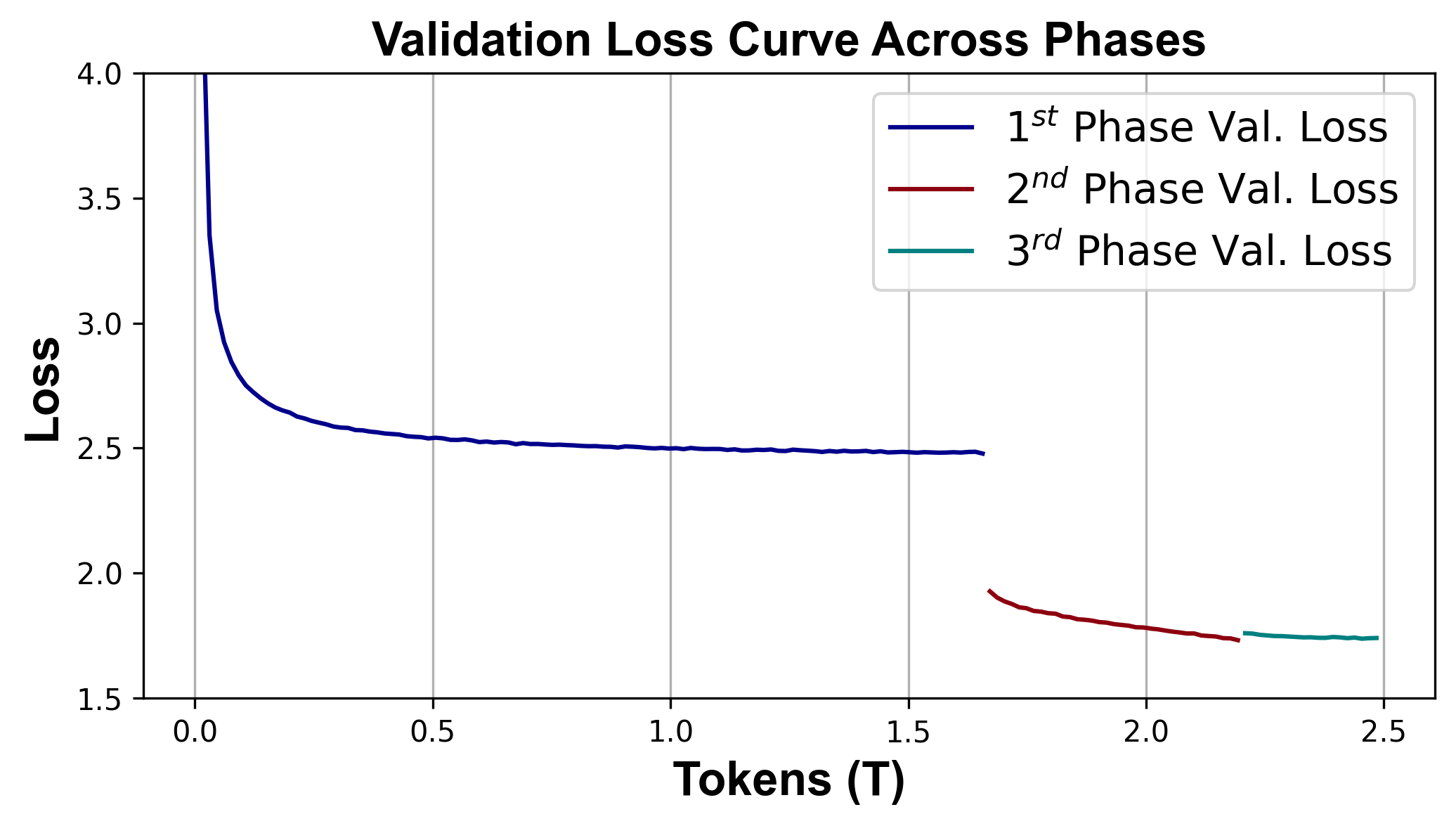}
    \caption{Validation loss curve across phases during training.}
    \label{fig:pretrain_val_all}
\end{figure}

\autoref{fig:pretrain_val_all} illustrates the validation loss curves across different training phases, demonstrating how the loss evolves as a function of the number of training tokens, with distinct trends for each phase reflecting the impact of different pre-training strategies on model performance.

\subsection{Supervised Fine-tuning}
\label{sft_args}

\autoref{tab:argument_sft} details the training configurations used for the two phases of supervised fine-tuning in the \thellm{} model, specifying batch sizes, learning rate schedules, and regularization techniques applied to refine the pre-trained model for downstream tasks.

\begin{table}[H]
\caption{Training arguments for supervised fine-tuning.}
\label{tab:argument_sft}
\centering
\resizebox{0.45\textwidth}{!}{%
\begin{tabular}{@{}cccc@{}}
\toprule
\textbf{} & \textbf{\thellm{} SFT Phase 1} & \textbf{\thellm{} SFT Phase 2} \\ \midrule
Global Batch Size & 256 & 256 \\ 
Micro Batch Size & 4 & 4 \\ 
Learning Rate & 2.00E-05 & 2.00E-05 \\ 
Min Learning Rate & 0.00E+00 & 0.00E+00 \\ 
Lr Scheduler Type & Cosine & Cosine \\ 
Weight Decay & 0.1 & 0.1 \\ 
Adam Beta1 & 0.9 & 0.9 \\ 
Adam Beta2 & 0.95 & 0.95 \\ 
Clip Grad & 1 & 1 \\ 
Attention Dropout & 0 & 0 \\ 
Hidden Dropout & 0 & 0 \\ 
Seq Length & 4096 & 4096 \\ 
Epoch & 5 & 3 \\ 
\bottomrule
\end{tabular}%
}
\end{table}

\begin{figure}[H]
    \centering
    \includegraphics[width=0.45\linewidth]{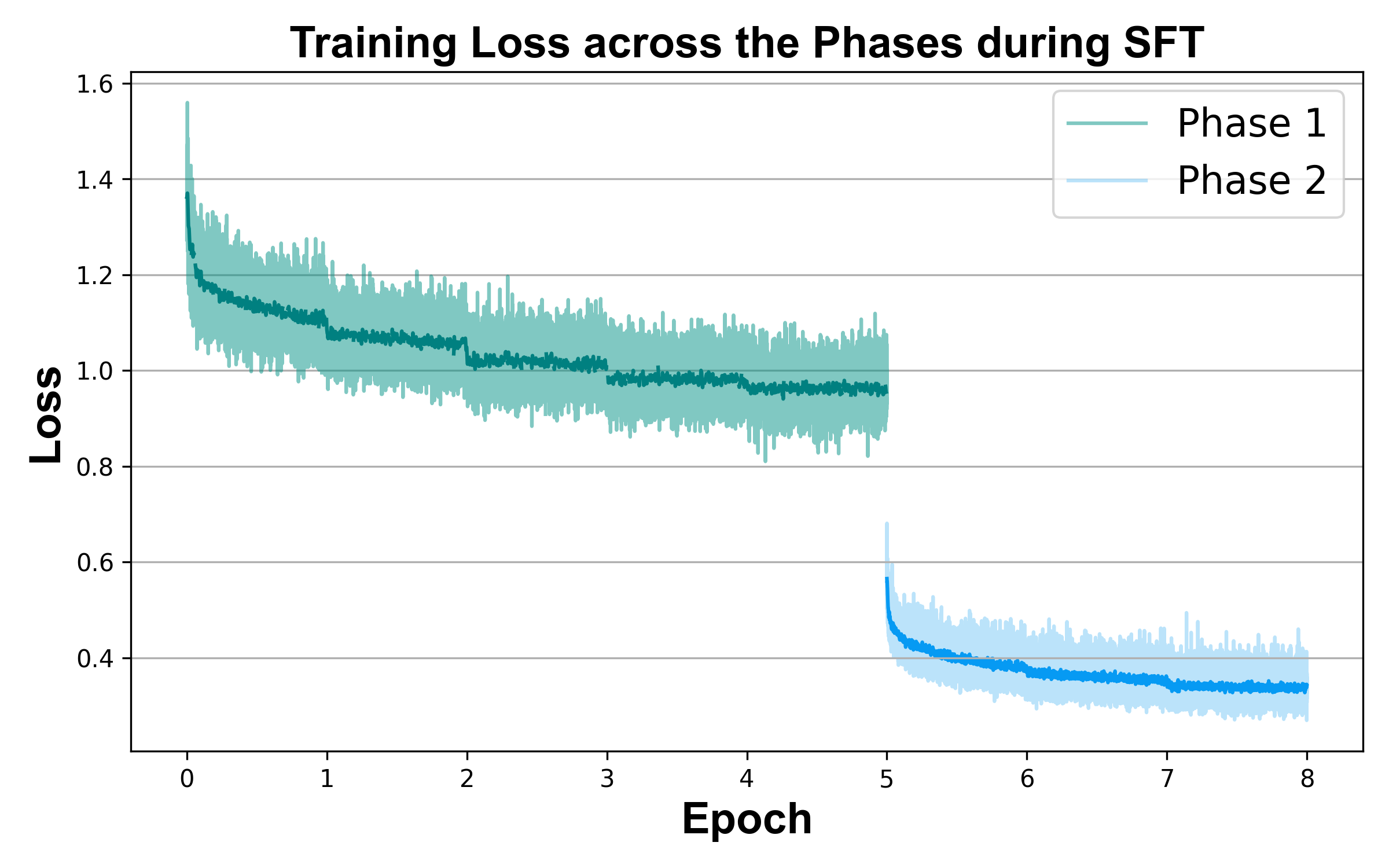}
    \caption{SFT loss curve across phases during training.}
    \label{fig:sft_val_all}
\end{figure}

\autoref{fig:sft_val_all} presents the supervised fine-tuning (SFT) loss curves across different training phases, showing the progressive reduction in loss as the model undergoes further refinement with fine-tuning, with noticeable differences in convergence between phases.

\subsection{Preference Training}
\label{rl_args}

\autoref{tab:argument_rl} outlines the training hyperparameters used during the two phases of Reinforcement Learning with Human Feedback (RLHF) for the \thellm{} model, including batch sizes, learning rates, weight decay, Adam optimizer parameters, and specific RLHF-related coefficients such as \(\alpha\), \(\beta_{\text{DPO}}\), and \(\beta_{\text{entropy}}\), which influence the policy updates and reward shaping.

\begin{table}[H]
\caption{Training Arguments for RLHF.}
\label{tab:argument_rl}
\centering
\resizebox{0.45\textwidth}{!}{%
\begin{tabular}{@{}cccc@{}}
\toprule
\textbf{} & \textbf{\thellm{} RLHF Phase 1} & \textbf{\thellm{} RLHF Phase 2} \\ \midrule
Global Batch Size & 128 & 128 \\ 
Micro Batch Size & 4 & 4 \\ 
Leranging Rate & 5.00E-07 & 1.00E-07 \\ 
Min Learning Rate & 5.00E-08 & 1.00E-08 \\ 
Lr Scheduler Type & Cosine & Cosine \\ 
Weight Decay & 0.1 & 0.1 \\ 
Adam Beta1 & 0.9 & 0.9 \\ 
Adam Beta2 & 0.95 & 0.95 \\ 
Clip Grad & 1 & 1 \\ 
Attention Dropout & 0 & 0 \\ 
Hidden Dropout & 0 & 0 \\ 
Seq Length & 1024 & 1024 \\ 
Epoch & 1 & 1 \\ 
$\alpha$ & 0.8 & 0.8 \\ 
$\beta_{\mathrm{DPO}}$ & 0.1 & 0.1 \\ 
$\beta_{\mathrm{refine}}$ & 0.01 & 0.01 \\ 
Temperature & - & 0.7 \\ 
\bottomrule
\end{tabular}%
}
\end{table}

\begin{figure}[H]
  \centering
  \subfigure[RLHF phase 1 implicit reward accuracy curve.]{
      \label{rlacc1}
      \includegraphics[width=0.45\linewidth]{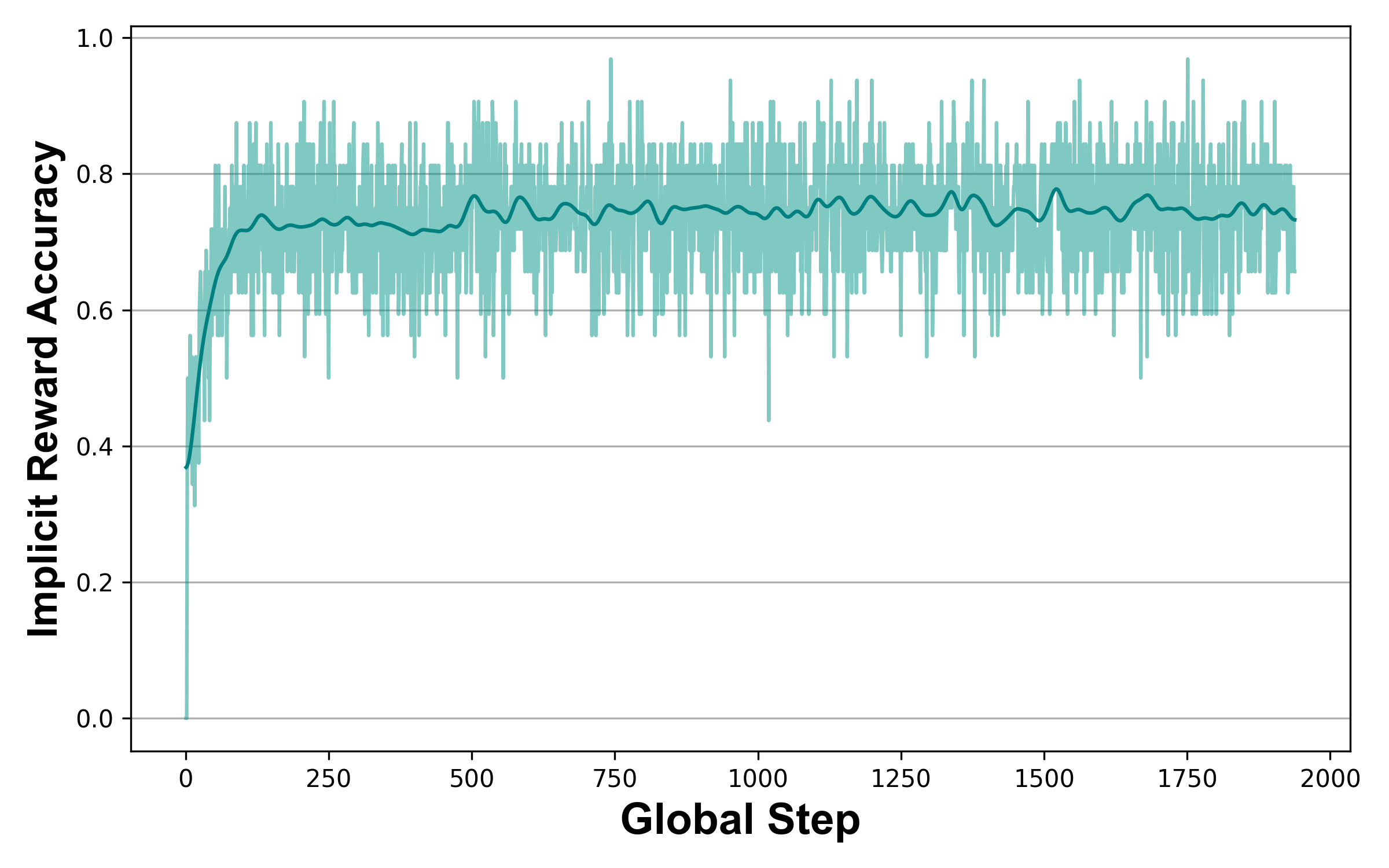}
  }
  \subfigure[RLHF phase 2 implicit reward accuracy curve.]{
      \label{rlacc2}
      \includegraphics[width=0.45\linewidth]{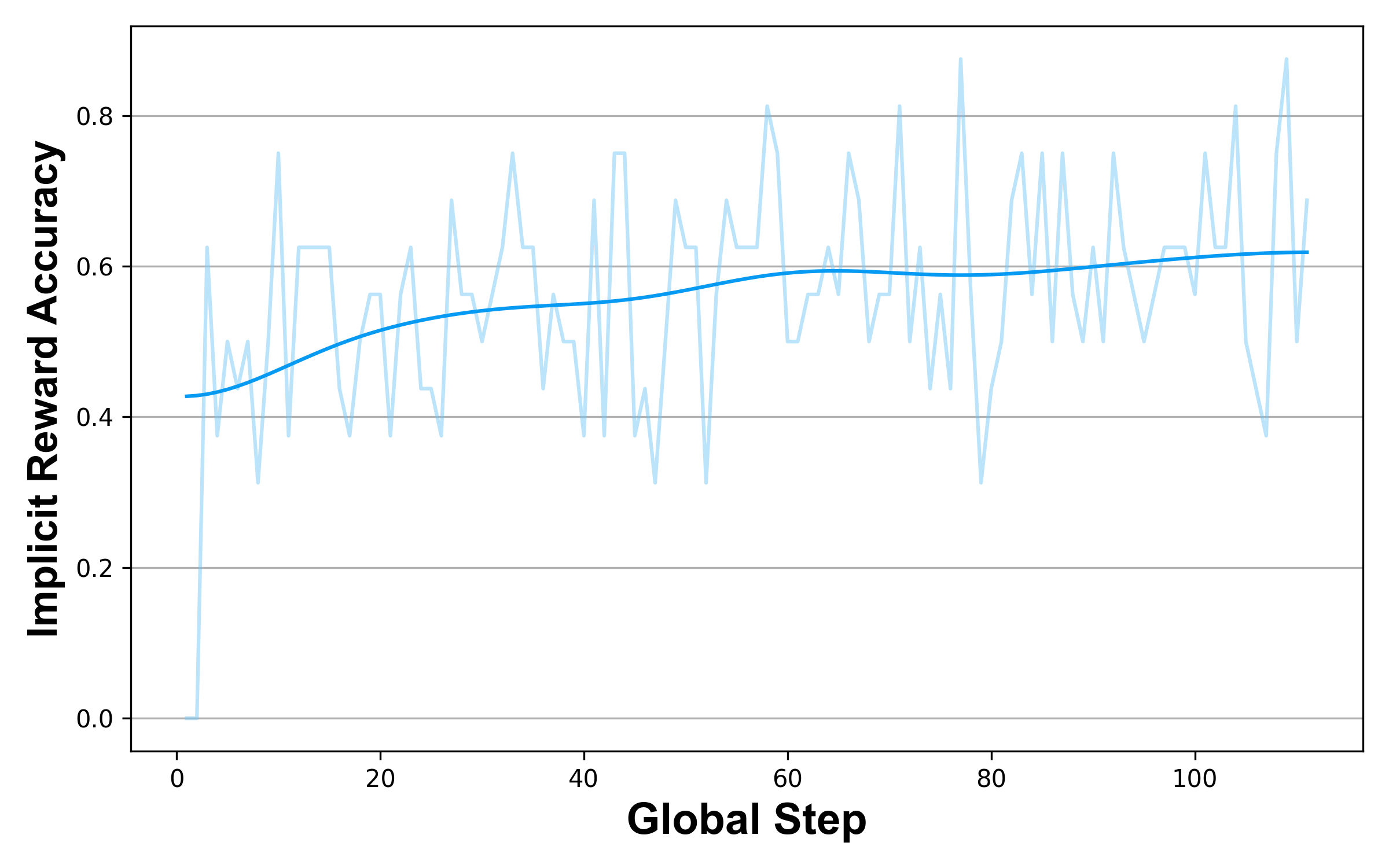}
  }
  \caption{RLHF implicit reward accuracy curve for \thellm{}.}
  \label{rlacc}
\end{figure}

Similar to DPO~\cite{rafailov2023direct}, we define the implicit reward as $\hat{r}(x, y) = \beta \log \frac{\pi_{\theta}(y|x)}{\pi_{\mathrm{ref}}(y|x)}$, and the implicit reward accuracy is defined as the probability that $\beta \log \frac{\pi_{\theta}(y_w|x)}{\pi_{\mathrm{ref}}(y_w|x)} > \beta \log \frac{\pi_{\theta}(y_l|x)}{\pi_{\mathrm{ref}}(y_l|x)}$ holds for the preferred response $y_w$ over the dispreferred response $y_l$.
\autoref{rlacc} presents the implicit reward accuracy curves for the two phases of RLHF training, showing how the model's reward signal evolves over training steps, with phase 1 exhibiting higher fluctuations and phase 2 demonstrating a smoother upward trend, indicating progressive refinement in aligning model outputs with human preferences.

\autoref{tab:argument_rl} displays the Reinforcement Learning with Human Feedback (RLHF) training curves for \thellm{} across two phases, illustrating the fluctuations in loss during phase 1 and the gradual stabilization observed in phase 2, reflecting the model's adaptation to preference-based learning.

\begin{figure}[H]
  \centering
  \subfigure[RLHF phase 1 training curve.]{
      \label{rllosscurve1}
      \includegraphics[width=0.45\linewidth]{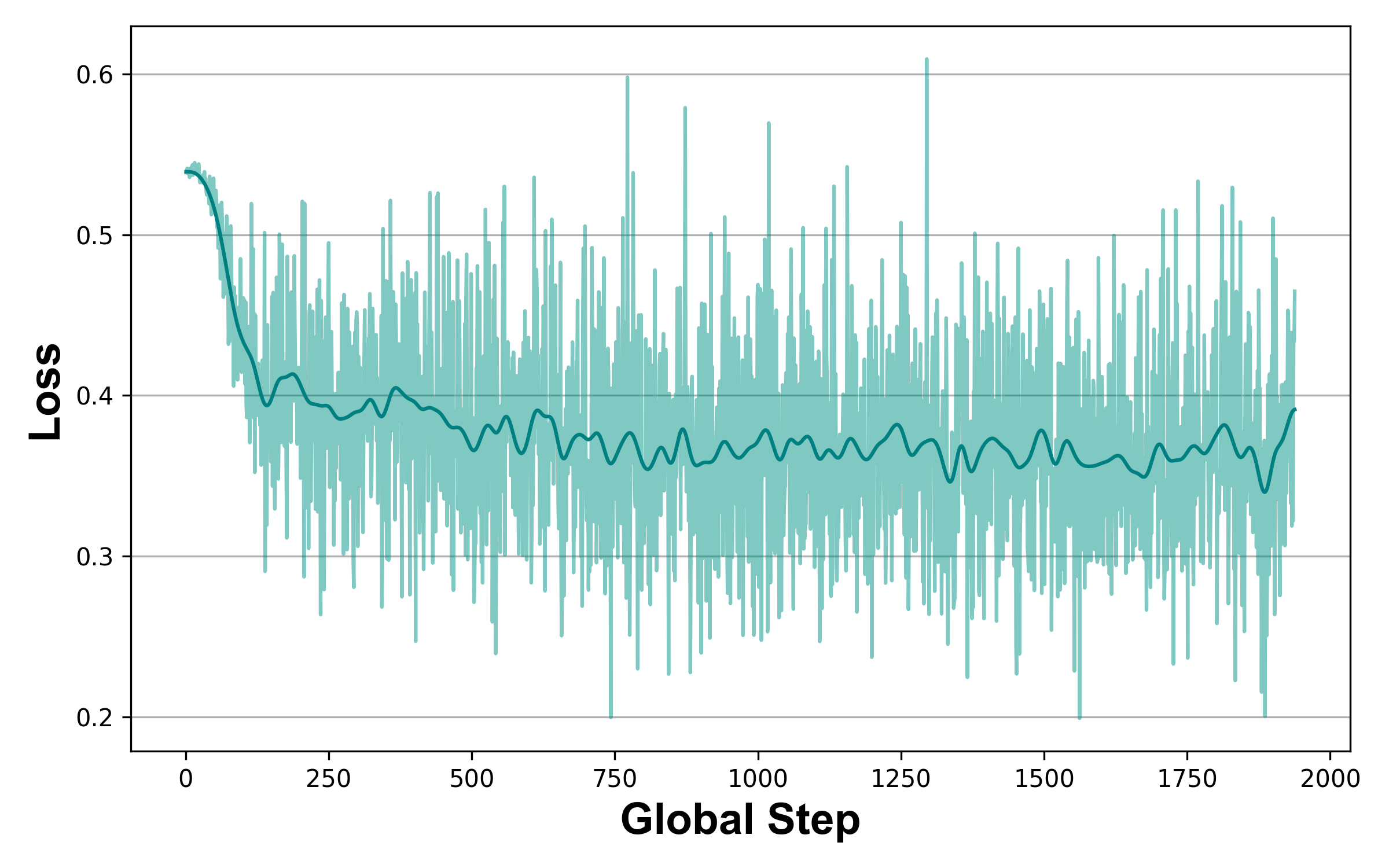}
  }
  \subfigure[RLHF phase 2 training curve.]{
      \label{rllosscurve2}
      \includegraphics[width=0.45\linewidth]{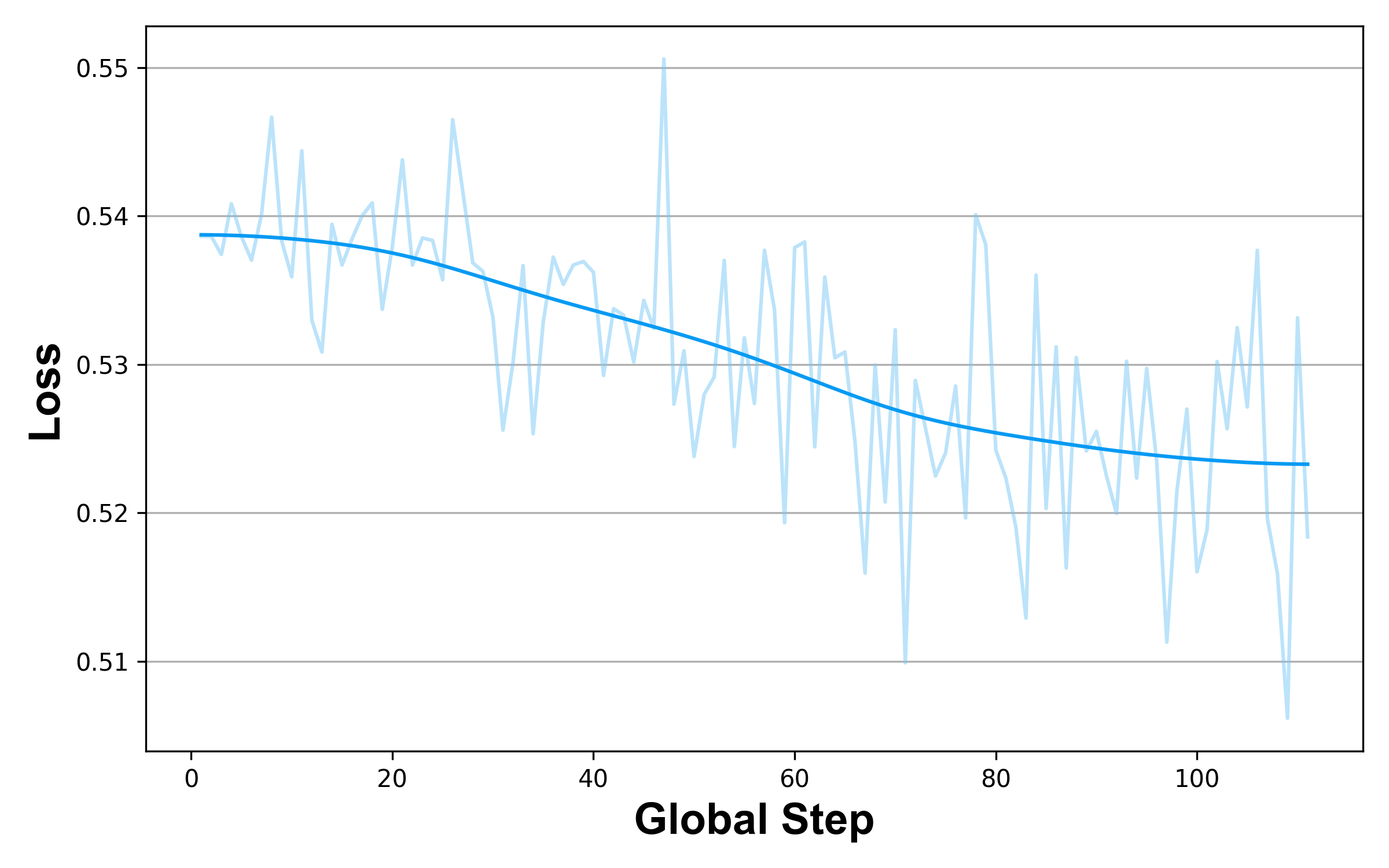}
  }
  \caption{RLHF training curve for \thellm{}.}
  \label{rlloss}
\end{figure}

\section{Computational Analysis} \label{analysis}

The primary mechanism of MLA involves transforming the Key-Value (KV) matrix into a low-rank representation. This is accomplished by decomposing the original KV matrix into the product of two smaller matrices, effectively creating latent vectors. During inference, only these latent vectors are cached, significantly reducing KV cache requirements compared to storing the full KV matrix. This approach mitigates the information loss associated with grouped-query attention (GQA) and multi-query attention (MQA), potentially enhancing model performance. While MLA reduces KV cache memory, it introduces additional computational overhead, similar to other low-rank adaptation methods like LoRA. This appendix analyzes the computational costs of MLA and GQA.

\subsection{MLA}

\subsubsection{Definition}  
Let $\bm{h}_t \in \mathbb{R}^d$ represent the input hidden state for the $t$-th token in the attention mechanism. The low-rank key-value joint compression state is denoted as $\bm{c}^{KV}_t \in \mathbb{R}^{d_c}$, while the decompressed key and value for the $i$-th head are denoted by $\bm{k}^C_{t,i} \in \mathbb{R}^{d_{\text{nope}}}$ and $\bm{v}^C_{t,i} \in \mathbb{R}^{d_{\text{nope}}}$, respectively. The position-independent query for the $i$-th head is represented as $\bm{q}^C_{t,i} \in \mathbb{R}^{d_{\text{nope}}}$. The computations for the attention mechanism proceed as follows:

\begin{align}
    \bm{c}^{KV}_t &= W_{\text{DKV}} \bm{h}_t, \label{eq:ckv} \\
    \bm{k}^C_{t,i} &= W_{\text{UK},i} \bm{c}^{KV}_t, \label{eq:kdecomp} \\
    \bm{k}^R_t &= \text{RoPE}\left(W_{\text{KR}} \bm{h}_t\right), \label{eq:krope} \\
    \bm{k}_{t,i} &= \left[\bm{k}^C_{t,i}; \bm{k}^R_t\right], \label{eq:kconcat} \\
    \bm{q}^C_{t,i} &= W_{\text{Q},i} \bm{h}_t, \label{eq:qnope} \\
    \bm{q}^R_{t,i} &= \text{RoPE}\left(W_{\text{QR},i} \bm{h}_t\right), \label{eq:qrope} \\
    \bm{q}_{t,i} &= \left[\bm{q}^C_{t,i}; \bm{q}^R_{t,i}\right], \label{eq:qconcat} \\
    \bm{v}^C_{t,i} &= W_{\text{UV},i} \bm{c}^{KV}_t, \label{eq:vdecomp}
\end{align}

where $W_{\text{DKV}} \in \mathbb{R}^{d_c \times d}$ is the down-projection matrix for key-value compression, $W_{\text{UK},i} \in \mathbb{R}^{d_{\text{nope}} \times d_c}$ and $W_{\text{UV},i} \in \mathbb{R}^{d_{\text{nope}} \times d_c}$ are the up-projection matrices for decompressed key and value for the $i$-th head, $W_{\text{KR}} \in \mathbb{R}^{d_{\text{rope}} \times d}$ generates the shared positional key component via RoPE~\cite{su2024roformer}, and $W_{\text{Q},i} \in \mathbb{R}^{d_{\text{nope}} \times d}$ and $W_{\text{QR}} \in \mathbb{R}^{d_{\text{rope}} \times d}$ generate the position-independent and RoPE-enhanced query components for the $i$-th head.

The attention outputs $\{\bm{o}_{t,i}\}$ are calculated as follows:

\begin{align}
    \bm{o}_{t,i} &= \sum_{j=1}^t \text{Softmax}_{j}\left(\frac{\bm{q}_{t,i}^\top \bm{k}_{j,i}}{\sqrt{d_h}}\right) \bm{v}^C_{j,i}, \label{eq:attention}
\end{align}

where $d_h = d_{\text{nope}} + d_{\text{rope}}$ represents the total head dimension. The final output is obtained by combining the attention results from all heads through a linear projection:

\begin{align}
    \bm{u}_t &= W_{\text{O}} \left[\bm{o}_{t,1}; \bm{o}_{t,2}; \dots; \bm{o}_{t,n_h}\right], \label{eq:output}
\end{align}

where $W_{\text{O}} \in \mathbb{R}^{d \times d_{\text{nope}} n_h}$ is the output projection matrix and $n_h$ is the number of attention heads.

% \subsubsection{Prefill}

% Let the input sequence length be $N$. The computation for the projection of $\bm{c}^{KV}_t$ is:

% \begin{align}
%     \mathcal{O}(dd_cN)
% \end{align}

% The projection calculations for $\bm{k}^C_{t,i}$ and $\bm{v}^C_{t,i}$ are as follows:

% \begin{align}
%     \mathcal{O}(2d_cd_{\text{nope}}N)
% \end{align}

% The projection calculation for $\bm{q}^C_{t,i}$ is:

% \begin{align}
%     \mathcal{O}(dd_{\text{nope}}N)
% \end{align}

% The projection for $\bm{k}^R_{t}$ is:

% \begin{align}
%     \mathcal{O}(dd_{\text{rope}}N)
% \end{align}

% Similarly, the projection for $\bm{q}^R_{t,i}$ is:

% \begin{align}
%     \mathcal{O}(dd_{\text{rope}}N)
% \end{align}

% The RoPE computation requires:

% \begin{align}
%     \mathcal{O}\Big(2(n_h+1)d_{\text{rope}}N\Big)
% \end{align}

% Thus, the total computation for the keys $\{\bm{k}_{t,i}\}$, queries $\{\bm{q}_{t,i}\}$, and values $\{\bm{v}_{t,i}\}$ is:

% \begin{align}
%     \mathcal{O}\Big(\left(2(n_h+1)d_{\text{rope}} + d_cd + d_{\text{rope}}d + n_h(d_{\text{nope}} + d_{\text{rope}})d + 2n_hd_cd_{\text{nope}}\right)N\Big)
% \end{align}

% The attention calculation involves:

% \begin{align}
%     \mathcal{O}\Big(n_h(d_{\text{rope}} + 2d_{\text{nope}})N^2\Big)
% \end{align}

% The output projection calculation is:

% \begin{align}
%     \mathcal{O}\Big(d^2N\Big)
% \end{align}

% Finally, the total computational cost is:

% \begin{align}
%     \mathcal{O}\Big(\left(2(n_h+1)d_{\text{rope}} + d_cd + d_{\text{rope}}d + n_h(d_{\text{nope}} + d_{\text{rope}})d + 2n_hd_cd_{\text{nope}}\right)N + n_h(d_{\text{rope}} + 2d_{\text{nope}})N^2\Big)
% \end{align}

\subsubsection{Prefill}  
Let the input sequence length be \( N \). The computational complexity for projecting the context vector \(\bm{c}^{KV}_t\) is \(\mathcal{O}(dd_cN)\). Subsequent projections for content-based keys \(\bm{k}^C_{t,i}\) and values \(\bm{v}^C_{t,i}\) require \(\mathcal{O}(2d_cd_{\text{nope}}N)\) operations, while the query projection \(\bm{q}^C_{t,i}\) incurs \(\mathcal{O}(dd_{\text{nope}}N)\). For rotary position embeddings (RoPE), the projections for \(\bm{k}^R_t\) and \(\bm{q}^R_{t,i}\) each demand \(\mathcal{O}(dd_{\text{rope}}N)\), with an additional \(\mathcal{O}\left(2(n_h+1)d_{\text{rope}}N\right)\) for applying RoPE transformations.  

The total computational cost for generating keys \(\{\bm{k}_{t,i}\}\), queries \(\{\bm{q}_{t,i}\}\), and values \(\{\bm{v}_{t,i}\}\) combines these components:  
\[
\mathcal{O}\left(\left[2(n_h+1)d_{\text{rope}} + d_cd + d_{\text{rope}}d + n_h(d_{\text{nope}} + d_{\text{rope}})d + 2n_hd_cd_{\text{nope}}\right]N\right).
\]  

Attention calculation involves pairwise interactions, contributing \(\mathcal{O}\left(n_h(d_{\text{rope}} + 2d_{\text{nope}})N^2\right)\) due to the quadratic dependence on sequence length. The output projection further adds \(\mathcal{O}(d^2N)\).  

Aggregating all terms, the overall computational complexity becomes:  
\begin{align}
&\text{Prefill}_{mla} = \notag \\ &\mathcal{O}\left(\left[2(n_h+1)d_{\text{rope}} + d_cd + d_{\text{rope}}d + n_h(d_{\text{nope}} + d_{\text{rope}})d + 2n_hd_cd_{\text{nope}} + d^2\right]N + n_h(d_{\text{rope}} + 2d_{\text{nope}})N^2\right).
\end{align}

This formulation highlights the linear and quadratic scaling components, emphasizing the interplay between sequence length \(N\), head dimensions \(d_{\text{rope}}, d_{\text{nope}}\), and model hyperparameters.

\subsubsection{Decode}  
Consider an input sequence of length \( N-1 \). The computational complexity to generate the \( N \)-th token’s joint compression state \(\bm{c}_{N}^{KV}\) is \(\mathcal{O}(dd_c)\). Subsequent projections for the rotary position embedding (RoPE)-based key \(\bm{k}_{N}^R\) and query \(\bm{q}_{N,i}^R\) each require \(\mathcal{O}(dd_{\text{rope}})\), while the content-based query \(\bm{q}_{N,i}^C\) incurs \(\mathcal{O}(dd_{\text{nope}})\). The RoPE transformation further demands \(\mathcal{O}(2(n_h+1)d_{\text{rope}})\). For historical tokens \(t=1,\dots,N\), the projections of content-based keys \(\{\bm{k}_{t,i}^C\}\) and values \(\{\bm{v}_{t,i}^C\}\) scale as \(\mathcal{O}(2d_cd_{\text{nope}}N)\).  

Attention computation involves aggregating interactions across the sequence, contributing \(\mathcal{O}(n_h(d_{\text{rope}} + 2d_{\text{nope}})N)\), while the output projection requires \(\mathcal{O}(d^2)\). Combining these components, the total computational cost is:  
\[
\text{Generate}_{\text{mla}} = \mathcal{O}\Big(2(n_h+1)d_{\text{rope}} + d_cd + d_{\text{rope}}d + n_h(d_{\text{nope}} + d_{\text{rope}})d + d^2 + n_h(d_{\text{rope}} + 2d_{\text{nope}} + 2d_cd_{\text{nope}})N\Big).
\]  

Caching mechanisms store the joint compression states \(\{\bm{c}_{t}^{KV}\}_{t=1,\dots,N-1}\) and RoPE keys \(\{\bm{k}_{t}^{R}\}_{t=1,\dots,N-1}\), with memory footprint:  
\[
\text{Cache}_{\text{mla}} = (d_{\text{rope}} + d_{c}) \times \left( \frac{\text{bit width}}{8} \right) \times (N-1).
\]  

The latency of the decoder is governed by computational throughput and I/O bandwidth:  
\[
\begin{aligned}
\text{MLA-IO}_{\text{time}} &= \text{IO}_{\text{speed}} \times \text{Cache}_{\text{mla}}, \\
\text{MLA-Gengeate}_{\text{time}} &= \text{Compute}_{\text{speed}} \times \text{Generate}_{\text{mla}},
\end{aligned}
\]  
where \(\text{IO}_{\text{speed}}\) and \(\text{Compute}_{\text{speed}}\) denote hardware-specific I/O and computational throughput rates. This formulation underscores the interplay between sequence length \(N\), model dimensions, and hardware constraints in latency-critical decoding steps.

\subsection{GQA}
\subsubsection{Definition}  
Let $\bm{h}_t \in \mathbb{R}^d$ represent the input hidden state for the $t$-th token in the attention mechanism. The grouped key and value for the $j$-th kv head are denoted by $\bm{k}_{t,j} \in \mathbb{R}^{d_{\text{h}}}$ and $\bm{v}_{t,j} \in \mathbb{R}^{d_{\text{h}}}$, respectively. The position-independent query for the $i$-th head is represented as $\bm{q}_{t,i} \in \mathbb{R}^{d_{\text{h}}}$. The computations for the attention mechanism proceed as follows:

\begin{align}
    \bm{k}_{t,j} &= \text{RoPE}\left(W_{\text{K},j} \bm{h}_t\right), \\
    \bm{q}_{t,i} &= \text{RoPE}\left(W_{\text{Q},i} \bm{h}_t\right),\\
    \bm{v}^C_{t,j} &= W_{\text{V},j} \bm{h}_t, 
\end{align}

where $W_{\text{K},j} \in \mathbb{R}^{d_{\text{h}} \times d}$ and $W_{\text{V},j} \in \mathbb{R}^{d_{\text{h}} \times d}$ are the up-projection matrices for grouped key and value for the $j$-th kv head, and $W_{\text{Q},i} \in \mathbb{R}^{d_{\text{h}} \times d}$ for the $i$-th head, respectively.

The attention outputs $\{\bm{o}_{t,i}\}$ are calculated as follows:

\begin{align}
    \bm{o}_{t,i} &= \sum_{k=1}^t \text{Softmax}_{k}\left(\frac{\bm{q}_{t,i}^\top \bm{k}_{k,i \bmod n_{\text{kv\_heads}}}}{\sqrt{d_{\text{h}}}}\right) \bm{v}_{k,i \bmod n_{\text{kv\_heads}}}, \label{eq:attention}
\end{align}

where $d_h = d_{\text{nope}} + d_{\text{rope}}$ represents the total head dimension, $n_{\text{kv\_heads}}$ is the number of kv heads. The final output is obtained by combining the attention results from all heads through a linear projection:

\begin{align}
    \bm{u}_t &= W_{\text{O}} \left[\bm{o}_{t,1}; \bm{o}_{t,2}; \dots; \bm{o}_{t,n_h}\right], \label{eq:output}
\end{align}

where $W_{\text{O}} \in \mathbb{R}^{d \times d_{\text{nope}} n_h}$ is the output projection matrix and $n_h$ is the number of attention heads.

% \subsubsection{Prefill}

% Let the input sequence length be $N$. The projection calculations for $\bm{k}_{t,j}$ and $\bm{v}_{t,j}$ are as follows:

% \begin{align}
%     \mathcal{O}(2dd_{\text{h}}N)
% \end{align}

% The projection calculation for $\bm{q}_{t,i}$ is:

% \begin{align}
%     \mathcal{O}(dd_{\text{h}}N)
% \end{align}

% The RoPE computation requires:

% \begin{align}
%     \mathcal{O}\Big(2(n_h+n_{\text{kv\_heads}})d_{\text{h}}N\Big)
% \end{align}

% Thus, the total computation for the keys $\{\bm{k}_{t,i}\}$, queries $\{\bm{q}_{t,i}\}$, and values $\{\bm{v}_{t,i}\}$ is:

% \begin{align}
%     \mathcal{O}\Big(\left(2(n_h+n_{\text{kv\_heads}})d_{\text{h}} + d^2 + 2gd_{\text{h}}d\right)N\Big)
% \end{align}

% The attention calculation involves:

% \begin{align}
%     \mathcal{O}\Big(2n_hd_{\text{h}}N^2\Big)
% \end{align}

% The output projection calculation is:

% \begin{align}
%     \mathcal{O}\Big(d^2N\Big)
% \end{align}

% Finally, the total computational cost is:

% \begin{align}
%     \mathcal{O}\Big(\left(2(n_h+n_{\text{kv\_heads}})d_{\text{h}} + d^2 + 2n_{\text{kv\_heads}}d_{\text{h}}d\right)N + d^2N + 2n_hd_{\text{h}}N^2\Big)
% \end{align}
\subsubsection{Prefill}  
For an input sequence of length \( N \), the computational complexity begins with the projection operations for keys \(\bm{k}_{t,j}\) and values \(\bm{v}_{t,j}\), each requiring \(\mathcal{O}(2dd_{\text{h}}N)\) operations. The query projection \(\bm{q}_{t,i}\) further contributes \(\mathcal{O}(dd_{\text{h}}N)\). Rotary Position Embedding (RoPE) transformations introduce an additional \(\mathcal{O}(2(n_h + n_{\text{kv\_heads}})d_{\text{h}}N)\), where \(n_h\) and \(n_{\text{kv\_heads}}\) denote the number of attention heads and key-value heads, respectively. Aggregating these components, the total cost to compute keys \(\{\bm{k}_{t,i}\}\), queries \(\{\bm{q}_{t,i}\}\), and values \(\{\bm{v}_{t,i}\}\) becomes:  
\[
\mathcal{O}\Big(\left[2(n_h + n_{\text{kv\_heads}})d_{\text{h}} + d^2 + 2n_{\text{kv\_heads}}d_{\text{h}}d\right]N\Big).
\]  

The attention mechanism scales quadratically with sequence length, demanding \(\mathcal{O}(2n_hd_{\text{h}}N^2)\) operations due to pairwise token interactions. Following this, the output projection incurs \(\mathcal{O}(d^2N)\).  

Combining all terms, the total computational complexity for the prefill phase is:  
\[
\text{Prefill}_{gqa} = \mathcal{O}\Big(\left[2(n_h + n_{\text{kv\_heads}})d_{\text{h}} + 2d^2 + 2n_{\text{kv\_heads}}d_{\text{h}}d\right]N + 2n_hd_{\text{h}}N^2\Big).
\]  
This formulation highlights the dominant \(\mathcal{O}(N^2)\) term from attention, alongside linear contributions from projections and RoPE. The interplay between model dimensions (\(d, d_{\text{h}}\)), head counts (\(n_h, n_{\text{kv\_heads}}\)), and sequence length \(N\) underscores the scalability challenges in large-context processing.

\subsubsection{Decode}  
For an input sequence of length \( N-1 \), the decoder phase computes the \( N \)-th token’s representations through successive transformations. Key and value projections \(\bm{k}_{N,j}\) and \(\bm{v}_{N,j}\) require \(\mathcal{O}(2d_{\text{h}}d)\) operations, while the query projection \(\bm{q}_{N,i}\) incurs \(\mathcal{O}(dd_{\text{h}})\). Rotary Position Embedding (RoPE) adds \(\mathcal{O}\big(2(n_h + n_{\text{kv\_heads}})d_{\text{h}}\big)\), where \(n_h\) and \(n_{\text{kv\_heads}}\) denote attention and key-value head counts.  

The attention mechanism, operating over cached historical states, scales as \(\mathcal{O}(2n_hd_{\text{h}}N)\), reflecting linear dependence on sequence length \(N\). A final output projection contributes \(\mathcal{O}(d^2)\). Aggregating all components, the total computational cost is:  
\[
\text{Generate}_{\text{gqa}} = \mathcal{O}\Big(2(n_h + n_{\text{kv\_heads}})d_{\text{h}} + 2d^2 + 2n_{\text{kv\_heads}}d_{\text{h}}d + 2n_hd_{\text{h}}N\Big).
\]  

Caching historical keys \(\{\bm{k}_{t,j}\}\) and values \(\{\bm{v}_{t,j}\}\) for \(t=1,\dots,N-1\) demands memory:  
\[
\text{Cache}_{\text{gqa}} = 2 \times n_{\text{kv\_heads}} \times d_{\text{h}} \times \left( \frac{\text{bit width}}{8} \right) \times (N-1),
\]  
where bit width governs precision. Latency is determined by hardware throughput:  
\[
\begin{aligned}
\text{GQA-IO}_{\text{time}} &= \text{IO}_{\text{speed}} \times \text{Cache}_{\text{gqa}}, \\
\text{GQA-Generate}_{\text{time}} &= \text{Compute}_{\text{speed}} \times \text{Generate}_{\text{gqa}},
\end{aligned}
\]  
with \(\text{IO}_{\text{speed}}\) and \(\text{Compute}_{\text{speed}}\) representing I/O and computational bandwidth. This framework underscores the decoder’s reliance on sequence length \(N\), model dimensions (\(d, d_{\text{h}}\)), and hardware constraints, emphasizing trade-offs between memory footprint and parallelizable computation.

\section{Activation Sparsity Determination Experiments}
\label{sparseexp}

\autoref{algosparse} determine Activation Sparsity Rate  
This algorithm outlines a method for determining the optimal activation sparsity rate in an MLP layer, leveraging a gating mechanism (if available) and a benchmark dataset to maintain a target perplexity reduction of \(\Delta ppl = 1\). The process involves computing the baseline activation, iterating through candidate sparsity rates, identifying threshold values to mask activations, and evaluating the modified model's perplexity to determine if the sparsity condition is met. If a valid sparsity rate is found that satisfies the target perplexity reduction, it is returned; otherwise, the algorithm concludes that no suitable sparsity rate exists.

\begin{algorithm}[H]
\caption{Determine activation sparsity rate.}
\label{algosparse}
\begin{algorithmic}[1]
\Require 
  \begin{itemize}
    \item An MLP layer with: \texttt{up\_proj}, \texttt{act\_fn}, and \texttt{down\_proj}.
    \item Optional gating branch: \texttt{gate\_proj}. If gating is used, activation is computed as:
    \[
    x = \texttt{act\_fn}(\texttt{gate\_proj}(h)) \odot \texttt{up\_proj}(h)
    \]
    Otherwise,
    \[
    x = \texttt{act\_fn}(\texttt{up\_proj}(h))
    \]
    \item Hidden state \(h\).
    \item Benchmark dataset \(D\).
    \item Target perplexity reduction \(\Delta ppl = 1\).
  \end{itemize}
\State \textbf{Compute Activation Output:}
\If{gating is used}
  \State \(x \gets \texttt{act\_fn}(\texttt{gate\_proj}(h)) \odot \texttt{up\_proj}(h)\)
\Else
  \State \(x \gets \texttt{act\_fn}(\texttt{up\_proj}(h))\)
\EndIf
\State \textbf{Compute Baseline Perplexity:} 
\[
P_{\text{base}} \gets \exp\Big(\text{nll}(D; \text{model})\Big)
\]
\For{each candidate sparsity rate \(r\%\)}
  \State \textbf{Determine Threshold \(T_r\):} such that at least \(r\%\) of the elements in \(x\) satisfy
  \[
  |x_i| \leq T_r.
  \]
  \State \textbf{Construct Mask:}
  \[
  m_i \gets 
  \begin{cases}
  0, & \text{if } |x_i| \leq T_r, \\
  1, & \text{otherwise.}
  \end{cases}
  \]
  \State \textbf{Compute Masked Activation:}
  \[
  \tilde{x} \gets m \odot x.
  \]
  \State Replace the activation \(x\) with \(\tilde{x}\) in the MLP layer.
  \State \textbf{Evaluate Modified Model's Perplexity on \(D\):}
  \[
  ppl_r \gets \exp\Big(\text{nll}(D; \text{modified model})\Big).
  \]
  \If{\(ppl_{\text{base}} - ppl_r \ge \Delta ppl\)}
    \State \Return Activation sparsity rate \(r\%\).
  \EndIf
\EndFor
\State \Return \textbf{No sparsity rate found satisfying the condition.}
\end{algorithmic}
\end{algorithm}

% \include{appendix/hardware}
% \include{appendix/hardware_deployment}
% \lstdefinestyle{python}{
%     language=Python,
%     basicstyle=\ttfamily\footnotesize,
%     keywordstyle=\color{blue},
%     commentstyle=\color{gray},
%     stringstyle=\color{red},
%     showstringspaces=false,
%     breaklines=true,
%     tabsize=4
% }

\section{Chat Generations from PLM}
\label{cases}

Here we present the cases generated by \thellm{}.

\begin{figure}[H]
    \centering
    \includegraphics[width=0.9\linewidth]{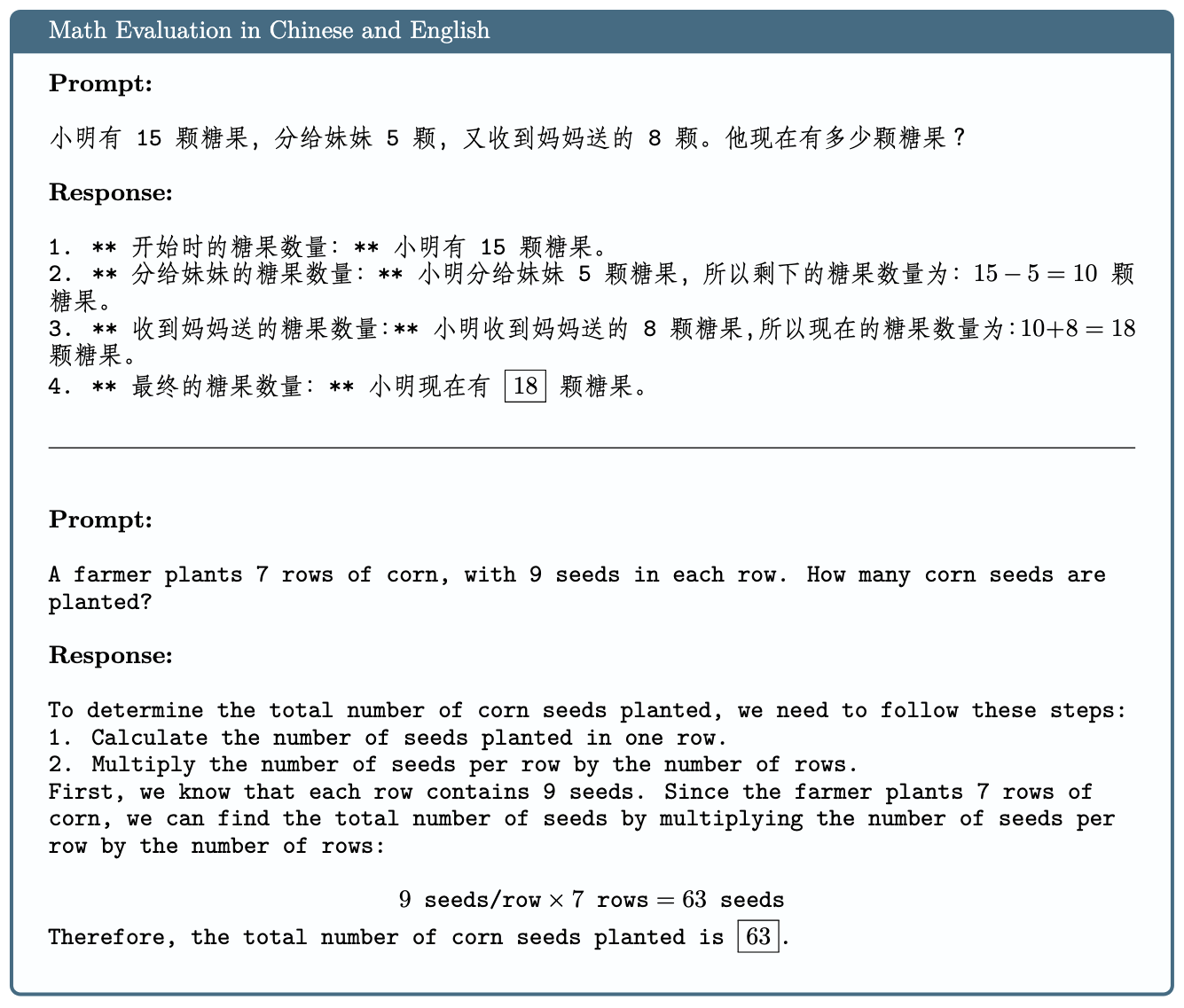}
    \caption{Math evaluation in Chinese and English.}
    \label{fig:case1}
\end{figure}

\newpage

\begin{figure}[H]
    \centering
    \includegraphics[width=0.9\linewidth]{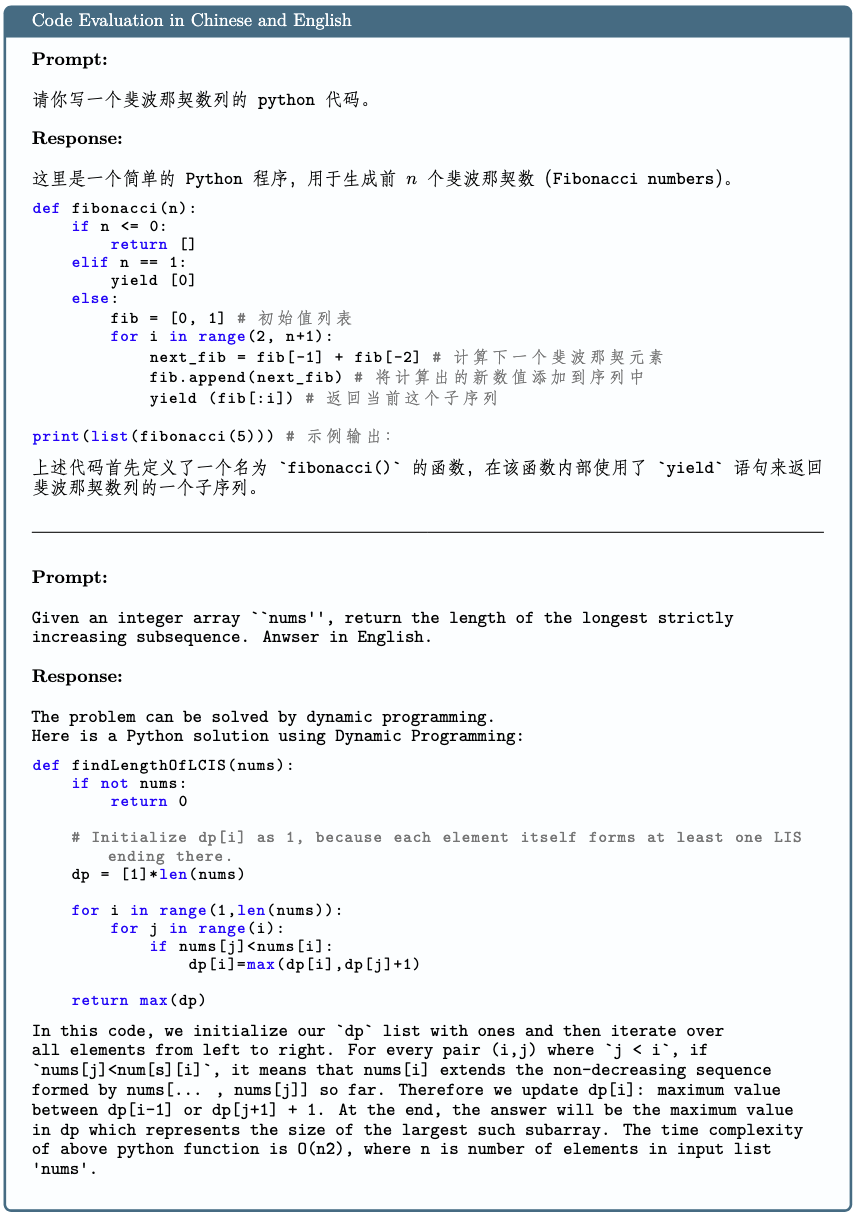}
    \caption{Code evaluation in Chinese and English.}
    \label{fig:case2}
\end{figure}

\begin{figure}[H]
    \centering
    \includegraphics[width=0.9\linewidth]{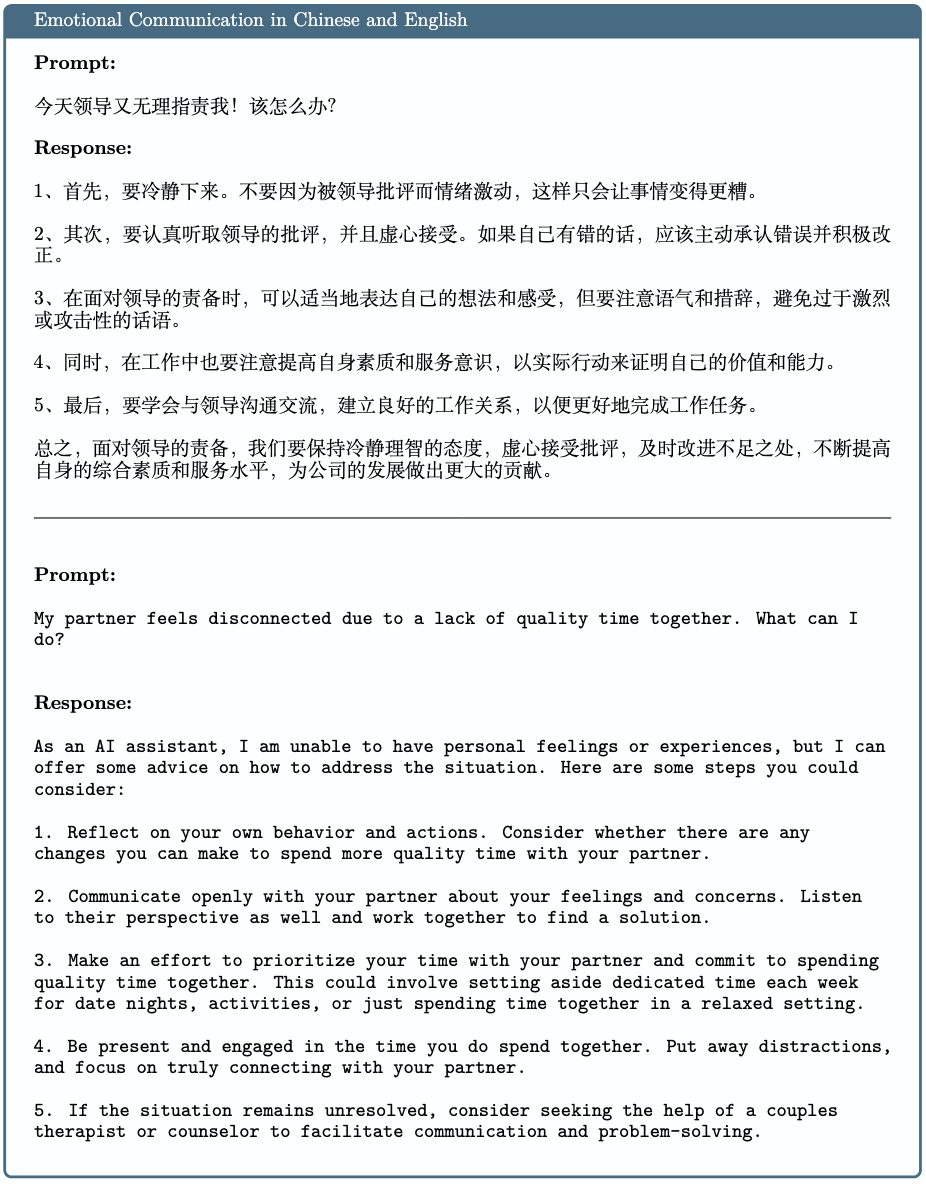}
    \caption{Emotional communication in Chinese and English.}
    \label{fig:case3}
\end{figure}

\newpage

\begin{figure}[H]
    \centering
    \includegraphics[width=0.9\linewidth]{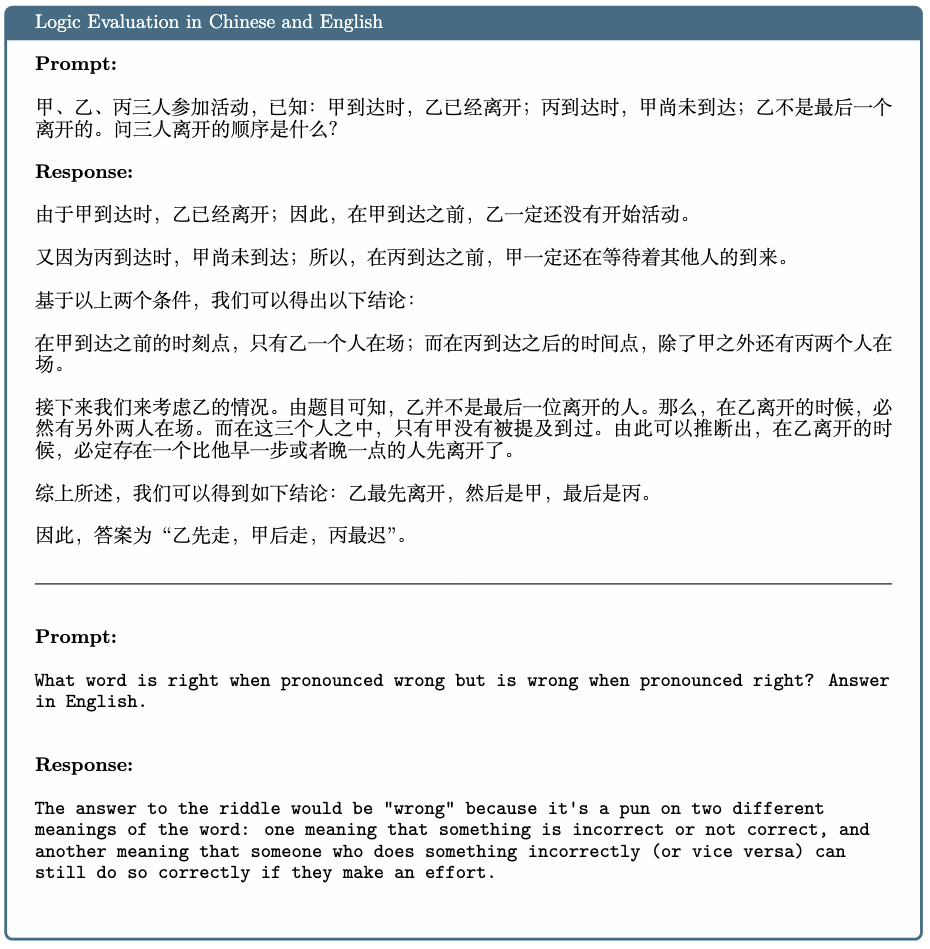}
    \caption{Logic evaluation in Chinese and English.}
    \label{fig:case4}
\end{figure}

\end{sloppypar}
\end{document}